\documentclass[letterpaper,twocolumn,10pt]{article}
\usepackage{usenix2020}

\usepackage{tikz}
\usepackage{amsmath}
\usepackage{caption}
\usepackage{subcaption}

\usepackage{amssymb}

\usepackage{amsfonts}
\usepackage{enumitem}
\usepackage{algorithm,algpseudocode,multirow,multicol,xspace}

\usepackage[capitalise]{cleveref}

\newtheorem{definition}{Definition}

\newif\ifcomm
\commtrue
\ifcomm
\newcommand{\crupdate}[1]{{#1}}
\newcommand{\lam}[1]{\textbf{\color{blue}Lam: #1}}
\newcommand{\minghao}[1]{\textbf{\color{red}Minghao: #1}}
\newcommand{\shay}[1]{\textbf{\color{green}Shay: #1}}
\newcommand{\ran}[1]{\textbf{\color{brown}Ran: #1}}
\newcommand{\minlan}[1]{\textbf{\color{purple}Minlan: #1}}
\newcommand{\mitz}[1]{\textbf{\color{purple}Mitz: #1}}
\newcommand{\kevin}[1]{\textbf{\color{orange}Kevin: #1}}
\newcommand{\sherry}[1]{\textbf{\color{yellow}Sherry: #1}}
\else

\newcommand{\lam}[1]{}
\newcommand{\minghao}[1]{}
\newcommand{\shay}[1]{}
\newcommand{\ran}[1]{}
\newcommand{\minlan}[1]{}
\newcommand{\mitz}[1]{}
\newcommand{\kevin}[1]{}
\newcommand{\sherry}[1]{}
\fi
\newcommand{\sysname}{THC\xspace}

\newcommand{\set}[1]{\ensuremath{\left\{#1\right\}}}
\newcommand{\norm}[1]{\left\lVert#1\right\rVert}

\newcommand{\ceil}[1]{\left\lceil#1\right\rceil}
\newcommand{\parentheses}[1]{\left(#1\right)}
\newcommand{\angles}[1]{\left\langle#1\right\rangle}

\newcommand{\secref}[1]{\S\ref{#1}}
\newcommand{\codesm}[1]{\texttt{\small #1}}
\DeclareMathOperator*{\minimize}{minimize}

\usepackage{tikz}
\usepackage{amsmath}
\usepackage{authblk}

\begin{document}
%-------------------------------------------------------------------------------

\title{THC: Accelerating Distributed Deep Learning Using Tensor Homomorphic Compression}

% % make title bold and 14 pt font (Latex default is non-bold, 16 pt)
% \title{\Large \bf THC: Accelerating Distributed Deep Learning Using Homomorphic Compression \ran{We need to change the format to match SIGCOMM's}}
%\titlenote{Produces the permission block, and copyright information}
%\subtitle{Extended Abstract}

%don't want date printed
\date{}

\author{Minghao Li$^{\dagger \dagger}$, Ran Ben Basat$^{\dagger}$, Shay Vargaftik$^{\mathsection}$, ChonLam Lao$^{\dagger \dagger}$, \\Kevin Xu$^{\dagger \dagger}$, Michael Mitzenmacher$^{\dagger \dagger}$, Minlan Yu$^{\dagger \dagger}$}
\affil{
      Harvard University~$^{\dagger \dagger}$, University College London~$^{\dagger}$, VMware Research~$^\mathsection$

}

% \author{Paper \# 821, 12 pages}
% \author{Minghao Li}
% % \authornote{Note}
% % \orcid{1234-5678-9012}
% \affiliation{%
%   \institution{Harvard University}
%   % \streetaddress{Address}
%   % \city{Cambridge} 
%   % \state{State} 
%   % \postcode{Zipcode}
% }
% % \email{minghaoli@g.harvard.edu}
% \author{Ran Ben Basat}
% \affiliation{
%   \institution{University College London}
% }
% \author{Shay Vargaftik}
% \affiliation{
%   \institution{VMware Research}
% }

\maketitle

% %-------------------------------------------------------------------------------
\begin{abstract}

Deep neural networks (DNNs) are the de facto standard for essential use cases, such as image classification, computer vision, and natural language processing.
As DNNs and datasets get larger, they require distributed training on increasingly larger clusters. 
A main bottleneck is the resulting communication overhead where workers exchange model updates (i.e., gradients) on a per-round basis.
To address this bottleneck and accelerate training, a widely-deployed approach is compression. However, previous deployments often apply bi-directional compression schemes by simply using a uni-directional gradient compression scheme in each direction. This results in significant computational overheads at the parameter server and increased compression error, leading to longer training and lower accuracy.  

We introduce Tensor Homomorphic Compression (THC), a novel bi-directional compression framework that enables the direct aggregation of compressed values and thus eliminating the aforementioned computational overheads.
%while optimizing the bandwidth to accuracy tradeoff
Moreover, THC is compatible with in-network aggregation (INA), which allows for further acceleration. Our evaluation shows that \crupdate{training representative vision and language models with} THC reaches target accuracy by \crupdate{$1.40\times$ to $1.47\times$} faster using INA and \crupdate{$1.28\times$ to $1.33\times$} faster using a software PS compared with state-of-the-art systems.
\end{abstract}

%Finally, we demonstrate that THC is scalable and tolerant for acceptable packet-loss rates. 

%and require distributed training, developing efficient training frameworks that incur low synchronization costs has drawn significant research interest. 
%
%A widely-deployed approach to accelerate training is gradient compression. However, previous deployments often apply simple compression schemes bi-directionally. Such an approach might lead to surprisingly poor training performance due to lossy compression and compression-related computational overhead. In this paper, we introduce Tensor Homomorphic Compression (THC), which is a novel theoretical framework for proposing and adopting algorithms that support the direct aggregation of compressed values. We also demonstrate enabling in-network aggregation in the THC framework with a system prototype. Testbed experiments show that the system prototype improves time-to-accuracy by up to 1.51x. We also present simulation results to highlight THC's scalability and packet-loss tolerance.  
% % %-------------------------------------------------------------------------------

%-------------------------------------------------------------------------------
\vspace*{-2mm}
\section{Introduction}
In the past decade, the scale of machine learning training and data volume has increased dramatically due to the growing demand for various ML applications~\cite{yolo_2016_CVPR, LaMDA, gpt3, vgg, resnet, wang2018machine, wang2016database, wu2016google}. 
%For example, Nvidia's Megatron-LM~\cite{megatron-lm-in-scale} used 3072 GPUs across 384 machines to train. \shay{Maybe update examples to more recent architectures? } \lam{I consider this might still SOTA but I can search for examples with new numbers after we finish others} 
%while Google's PaLM~\cite{palm} was trained using 6144 TPUs. 
Alibaba's general-purpose ML platforms also reported a rapid increase in ML training data, from hundreds of gigabytes to tens or even hundreds of terabytes, at an internet scale, within a few years~\cite{weng2022mlaas}. This trend is expected to continue in the future~\cite{ParamScaling} with the rapid advancements of giant models~\cite{LaMDA, gpt3, dlrm, moe, pmlr-v162-rajbhandari22a}. 
To support these large-scale models, we need large-scale distributed training ~\cite{gholami2018integrated, cassini, sipml, TopoOpt}.

However, distributed training incurs high communication overhead. 
Recent research \cite{ByteComp} has shown that the synchronization cost of GPT2~\cite{GPT2} and BERT-base~\cite{bert} in a 8-worker setting 
can be as high as 42\% and 49\% of the total time during training, even with state-of-the-art frameworks.
As the number of workers increases, the communication overhead rises substantially~\cite{ParamScaling}. 
Meanwhile, computing devices are pushing more data into the network with specialized ML accelerators~\cite{brainwave, vitis_ai, htfpga, finn-resnet} and more advanced GPU/TPU hardware, which {further increases the communication overhead~\cite{network-bottleneck1, network-bottleneck2}.}

\begin{figure*}[t!]
    \centering
    \includegraphics[clip, trim=0cm 4.7cm 0cm 2.7cm, width=0.95\linewidth]{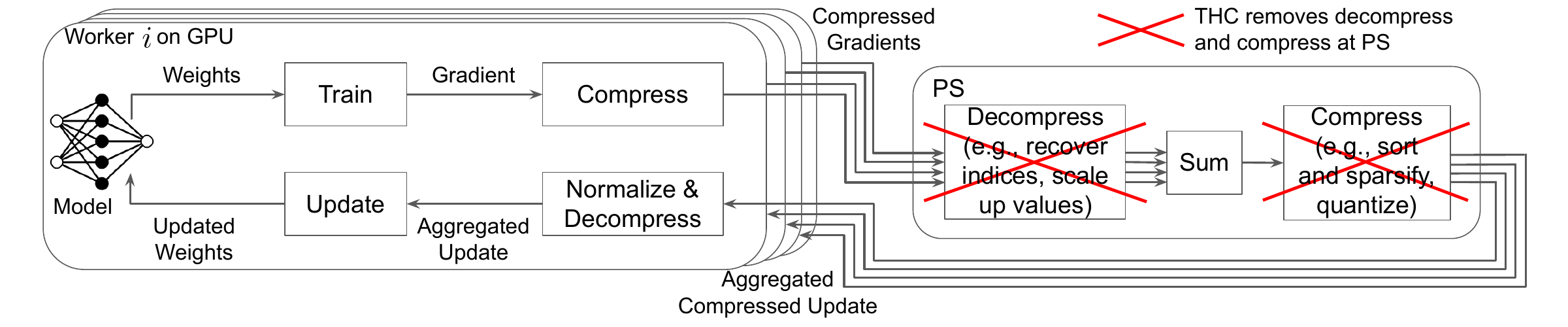}
    \vspace{-4mm}
	\caption{\crupdate{End-to-end workflow when using typical compression algorithms.}
        }
    \label{fig:e2e_normal_comrpession_example}
    \vspace{-4mm}
\end{figure*}

To reduce the communication overhead, many compression schemes have been conceived~\cite{HiPress, ByteComp, bernstein2018signsgd, TopK, TernGrad,dorfman2023docofl}. %\minlan{cite signSGD etc theory papers}
One common problem of these solutions is that they apply bi-directional compression. For example, in the Parameter Server (PS) architecture, the PS nodes first decompress all gradients, aggregate them, and then compress the aggregated gradients again. Such compression and decompression operations result in significant computational overheads and affect the training convergence time and attainable accuracy (see Section~\secref{background_compression}). 

To address these problems, we introduce Tensor Homomorphic Compression (THC), a novel bi-directional compression framework enabling the direct aggregation of compressed tensors (e.g., gradients) without first decompressing them, eliminating much of the aforementioned computational overhead. Additionally, direct aggregation can also run on programmable switches for further acceleration.

% Our framework further allows us to optimize the bandwidth-accuracy tradeoff.

This paper focuses on developing a THC framework to reduce the communication overhead of data parallelism for the parameter server architecture. Importantly, PS is effective in GPU/CPU hybrid clusters~\cite{BytePS, ByteComp, bytescheduler, janus}, which are common in clouds~\cite{microsoftsparecpu}. Furthermore, when we colocate a PS with each worker, it essentially functions as an AllReduce~\cite{BytePS}.
% If we colocate a PS with each worker, then it is essentially all-reduce~\cite{BytePS} for more efficient communications.

%the unified distributed training systems such as BytePS~\cite{BytePS}, which include both parameter servers~\cite{PS} and all-reduce~\cite{AllReduce} communications as special cases. Such PS architecture

%\minlan{Can theory people check/rewrite the following paragraphs?}

\crupdate{
From an algorithmic standpoint, the technical challenge is designing an algorithm that enables workers to accurately compress their gradients in a way that allows aggregating their results without decompressing each worker's message. 
We propose \emph{Homomorphic Compression} -- a property that enables this and develops efficient schemes that satisfy it while optimizing the accuracy.}
%
%To that end, we present the Homomorphic Compression property which means that the average of decompressing each compressed gradient is mathematically equivalent to decompressing (once) the average compressed gradient.
%while achieving compression scheme that does not slow overall training time and providing an aggregated result that yields overall training accuracy that competes with or exceeds the results when compared against baselines and state-of-the-art existing solutions. 

%From an algorithmic standpoint, the technical challenge is designing an algorithm that allows workers to compress their gradients so that their results can be aggregated without the need for decompressing each individual worker's message, while achieving compression scheme that does not slow overall training time and providing an aggregated result that yields overall training accuracy that competes with or exceeds the results when compared against baselines and state-of-the-art existing solutions. 
%\minlan{do we need to mention the proper explicitly?}
%\minlan{then say sth on how we retain the property and how to improve accuracy}

A key idea behind \sysname's ability to satisfy this property is an initial communication stage with minimal information exchange between the workers at the beginning of each round that allows them to \emph{coordinate} and ensure that their compressed gradients are directly aggregable. 
%
%satisfy this property by having all workers use a global `scale', and ensure accuracy using a lookup table that we optimize for the distribution of the worker vectors after applying the Randomized Hadamard Transform. Namely, a worker communicates each transformed coordinate by sending a table index, 
%
%
%to combine pre- and post-processing techniques\minlan{This sounds like sec 5. Can we summarize sec 4 here as well?} with minimal information exchange between the workers at the beginning of each communication stage to satisfy the Homomorphic Compression by \emph{coordinating} their compression, which allows for accurate quantization and aggregation in compressed form. 
%
To ensure high accuracy, \sysname does not directly encode the gradients but rather pre- and post-processes them with the GPU-friendly Randomized Hadamard Transform (RHT) to transform the gradients to a different representation that is amenable to accurate quantization. Since RHT preserves the tensor sizes (i.e., norms), by merely exchanging these norms during a preliminary light communication stage (a single float per client), the clients can align their quantization values such that they can be averaged without decompression. Furthermore, this communication step overlaps with applying the RHT transform and thus does not increase the compression time.
\sysname also employs an advanced non-uniform quantization technique we developed and an error-feedback mechanism to further improve the bandwidth to accuracy tradeoff.  %\shay{New text - end. Feel free to edit}

 We built \sysname on top of the BytePS \cite{BytePS} PyTorch extension. Our PS can run on either software or programmable switches. 
 %We also incorporate it into in-network aggregation services~\cite{SwitchML, ATP, PANAMA}, which employ {programmable switches to perform gradient aggregation. }
We perform extensive evaluation over seven representative DNN models on a local testbed and AWS EC2.
%a single rack testbed with four workers, a programmable switch, and 100 Gbps network. 
\crupdate{Testbed results show that} THC 
achieves the target accuracy \crupdate{$1.42\times$ to $1.47\times$} faster with aggregation on a programmable switch and \crupdate{$1.28\times$ to $1.33\times$} with a software PS, compared to the state-of-the-art distributed training framework (Horovod RDMA). THC with the programmable switch also improves the training throughput by up to 54\% over Horovod RDMA. \footnote{THC is available at \url{https://github.com/SophiaLi06/BytePS_THC}.
}
\begin{figure}[t]
    \centering
         \begin{subfigure}[t]{0.49\textwidth}
    \begin{minipage}[t]{0.48\textwidth}
        \centering
        \includegraphics[width=\textwidth]{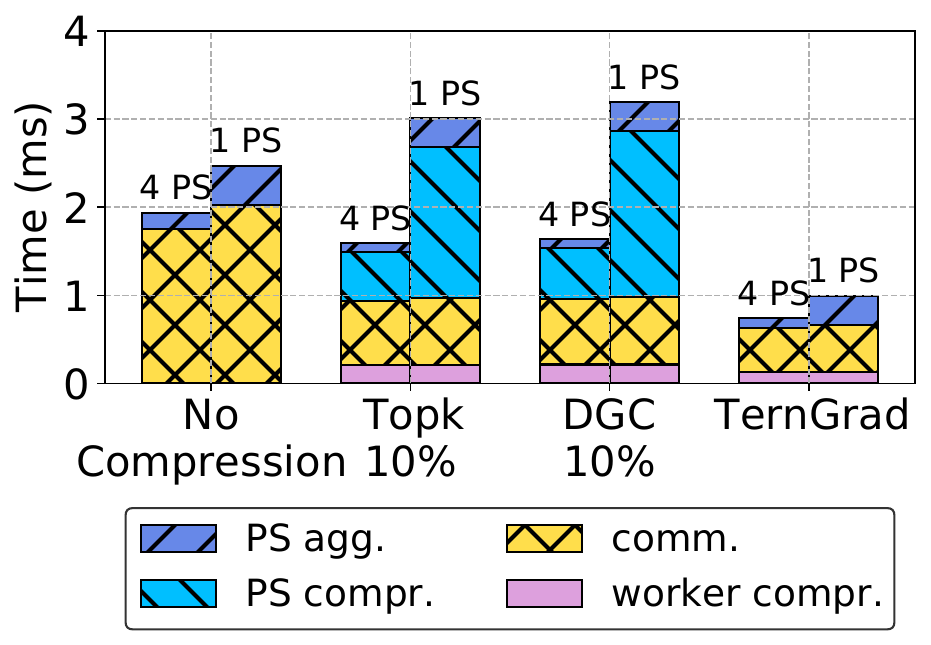} % first figure itself
        \caption{\crupdate{Communication round time of one 4MB partition.}}
        \label{fig:round_time_ratio}
    \end{minipage}\hfill
    \begin{minipage}[t]{0.48\textwidth}
        \centering
         \includegraphics[width=\textwidth, height=2.83cm]{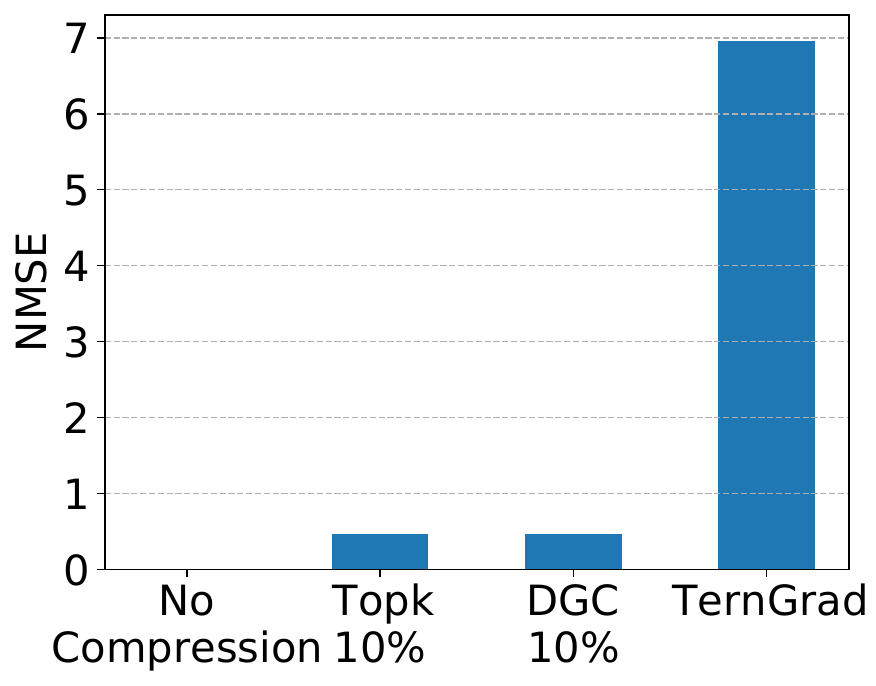}
         \caption{\crupdate{NMSE of various compression schemes with four workers.}}
         \label{fig:nmse_round_time}
    \end{minipage}
         \end{subfigure}
         \vspace*{1mm}
         \caption{Communication time and error.}
        \label{fig:compression_background}
        \vspace*{-5mm}
\end{figure}

\vspace*{-1mm}

\section{Background and Motivation}\vspace*{-1mm}

\begin{figure*}[t!]
    \centering 
    \includegraphics[width=0.71248\textwidth]{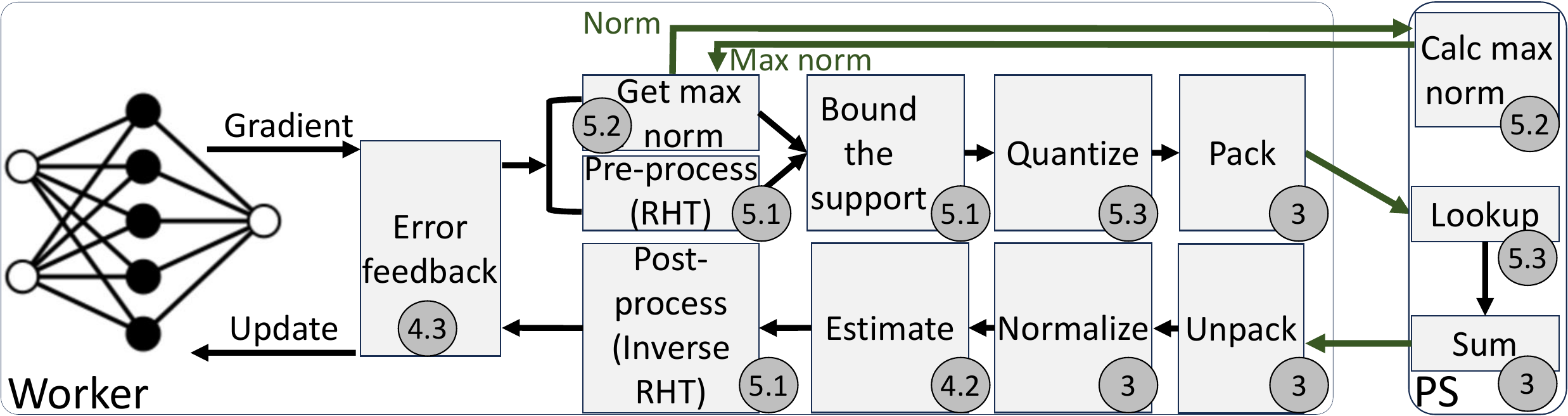} 
    \vspace*{-2mm}
    \caption{An illustration of THC, with the section numbers of each component. %\mitz{Maybe add in the caption why Tables are 8 bits -- e.g., summing 4 bit values and want to avoid overflow}\minlan{Is 32, 4, 8 specific in your algorithm or configurable? Make it clear in figure caption}
    % \ran{We need to add a division by $n$ after the decompression to the figure}
    % \ran{Make figure more vertical space efficient by putting the table indices to the side of the arrows and moving the `aggregated compressed update' text to the left of the vector}
    % \minghao{also on worker change into "Normalize and Decompress"}
    }
    \vspace*{-5mm}
    \label{fig:AlgorithmOverview}
\end{figure*}

% \minghao{
% The benefits of compression
% \begin{enumerate}
%     \item Reduce the traffic in both worker-to-PS and PS-to-worker direction
% \end{enumerate}
% The limitations of compression
% \begin{enumerate}
%     \item Not all training tasks can converge with aggressive compression (this is a major reason why sometimes we can't use high compression ratios and should choose in-network aggregation)
%     \item high computational overhead, especially when run on CPU
%     \item compression algorithms usually don't support direct aggregation and hence require additional operations 
% \end{enumerate}
% }
\crupdate{In this section, we give the background on compression and in-network aggregation and motivate the need for direct aggregation on compressed data. Gradient compression, a well-studied approach to 
lower the network communication of distributed training, reduces the volume of gradients transmitted in synchronization steps at the expense of convergence speed and accuracy.} There are two key techniques: sparsification and quantization. Sparsification may filter out coordinates of small magnitude in the gradient tensor. TopK~\cite{TopK} is a straightforward sparsification algorithm that only sends the top $k$ percent (by magnitude) of coordinates and their indices. \crupdate{Sparsification becomes lossy when many coordinates are nonzero.}
Quantization reduces the size of each element by reducing the precision of gradients. For example, TernGrad\cite{TernGrad} reduces the bit length of the gradient coordinates to two bits by converting each float into a value $x\in \{-1,0,1\}$.
Different compression techniques offer various accuracy, bandwidth, and time complexity tradeoffs.

%Such compression is often lossy (as we quantize real-valued coordinates into a small set of quantization values),

\vspace*{-1mm}
\subsection{\crupdate{Compression Cost and Tradeoffs}}
\label{background_compression}

Figure~\ref{fig:e2e_normal_comrpession_example} shows the bi-directional compression process in existing PS systems~\cite{HiPress, ByteComp}. Workers send compressed gradients to the PS. The PS decompresses and aggregates gradients. Then, to reduce the traffic back to workers, the PS compresses the aggregated results again before transmission. In such a design, while compression reduces the communication cost, decompressing and compressing data on PS nodes introduce high compute overhead and additional compression errors. 

% instead of communicating the gradient of each layer one by one. 
% We can view such a partition as the unit of data during synchronization. 

To quantize the computational overhead and estimation error of compression schemes, we run a microbenchmark transmitting a single 4MB partition \crupdate{(the recommended partition size that balances pipelining efficiency and system overheads as specified in BytePS repository\cite{BytePSrepo})} on our local testbed.
Training frameworks usually batch gradients and chunk them into same size partitions before communication~\cite{bytescheduler, li2020pytorch}. 
Since communication time grows linearly with the number of partitions, we simply measure the times for a single partition in our microbechmark. We measure the worker-side \textit{compression and decompression time ("worker compr.")}, the PS side \textit{compression and decompression time ("PS compr.")}, the PS \textit{aggregation time ("PS agg.")} and the worker-PS \textit{communication time ("comm.")} as shown in Figure~\ref{fig:round_time_ratio}. 

With one PS and four workers, \crupdate{two sparsification schemes TopK 10\% and DGC 10\%\cite{DGC} that communicate the top 10\% coordinates by magnitude slow down the end-to-end time by $19.3\%$ and $27.1\%$ of the no-compression round time.} This is due to the high PS computational overhead of compression and decompression that contributes \crupdate{up to $56.9\%$} of the round time even when the communication time is reduced. Note that the computational overhead of TopK \crupdate{ and DGC grow} as the PS aggregates more coordinates, making it a poor choice when the worker-to-parameter server ratio is imbalanced.
%This suggests that compression provides little to no benefit 
%when the worker-to-parameter server ratio is imbalanced, and each parameter server serves multiple machines.

When we use colocated PS (i.e., have four PS in total), \crupdate{TopK 10\% reduces the communication time by $60.4\%$ of that of no compression but takes an extra $0.54$ ms ($34.0\%$ of the round time)} to run compression/decompression on the PS. The end-to-end round time reduction is therefore diluted to \crupdate{$20.6\%$} of that of no compression.

% We measure the compression and decompression time in both worker-side and PS-side,  

%\minlan{Are we adding signSGD, and removing one of TopK?} \minghao{ah sorry I didn't add signSGD into the actual system prototype orz}
One can use other compression schemes (e.g., TernGrad) that take less time for decompression/compression at the PS side. However, these schemes have larger quantization errors. Figure~\ref{fig:nmse_round_time} shows the NMSE (Normalized Mean Squared Error, \crupdate{$NMSE(x,\hat{x})=\frac{||x-\hat{x}||_2^2}{||x||^2_2}$}), which quantifies the difference between the actual vector $x$ and the vector restored after compression $\hat{x}$. 
%\minlan{as we have more clients, would NMSE be fine for terngrad?}\lam{good question... any thoughts?@others who know TernGrad}\shay{TernGrad is biased... similartly to topK, its improvement will halt at some point}
TernGrad results in an NMSE that is by an order of magnitude larger than that of TopK 10\% (i.e., $6.95$ vs. $0.46$ with four workers). This large gap in NMSE means that TernGrad requires more iterations to reach the target accuracy or might fail to reach the target accuracy at all. In fact, provable convergence rates for distributed SGD have a linear dependence on NMSE (e.g., \cite{ErrorFeedback}), rendering quantization schemes with large NMSE less appealing for distributed training.

To address these limitations, 
\sysname allows direct aggregation of compressed gradients, which eliminates decompression and compression operations at PSes while ensuring high accuracy. A detailed comparison between THC and other compression schemes under different bit budgets is in Section \secref{sec:simulation}.

\vspace{-3mm}
\subsection{In-network Aggregation}\label{background_switch}
\vspace{-1mm}
% \minghao{start with saying that we are going over the important use cases of the THC framework}
% \minghao{need to mention that compression helps to increase the sliding window size on the switch.}
% \mitz{Say what aggregation is-- not clear to the reader yet!}% \lam{fixed}

In-network aggregation~\cite{NVIDIASHARP} is another option to reduce communication overhead. 
Recent research~\cite{SwitchML, ATP} has demonstrated that programmable switches can aggregate gradients from multiple workers, reducing the switch-to-PS traffic and resulting in a substantial training performance improvement.
% Recent studies~\cite{ATP, SwitchML} have shown that a programmable switch can aggregate gradients from many workers and thus reduce the traffic from the switch to the PS. SwitchML~\cite{SwitchML} demonstrates up to 5.5 times training performance speed up for real world ML models.
%These programmable network switches, which occupy a strategic central position among workers, have the capability to handle packet-level operations while maintaining high-speed transmission of network traffic. 
%\minlan{by how much? and how much improvement can they get?}. 
% However, the traffic from the PS back to the workers are of equal amount and cannot be reduced by switches.
\crupdate{However, using switches does not reduce the traffic volume generated by the workers as gradient compression does.}
% \lam{the traffic output from the worker also cannot be reduced right?}. 
% \lam{add hardware acceleration arguments}

\crupdate{
Most existing compression solutions are incompatible with in-network aggregation solutions because switches can not easily decompress and compress the data due to their programmability and resource limitations~\cite{ATP, whenshouldhotos19}.
Since THC supports direct aggregation over compressed data, it only requires the PS to do summation, which programmable switches can readily perform with their ALUs.} \crupdate{This} also offers new opportunities for incorporating compression with in-network {aggregation to further improve training performance.}

\renewcommand{\algorithmiccomment}[1]{\hfill$\triangleright$\ \ \textcolor{gray}{\footnotesize #1}}
\vspace{-3mm}
\section{THC Overview}\label{sec:overview}
\vspace{-1mm}

% Traditional bi-directional compression approaches require the PS to perform $n$ decompression operations in order to aggregate the gradients before recompressing them and introducing an additional error. Instead, 
We propose the \emph{Tensor Homomorphic Compression (THC)} framework, which allows the PS to merely aggregate the incoming compressed gradients and transmit the (still compressed!) aggregated values back to the workers. THC hence enables us to avoid the computational overhead of compression and decompression at PS while still having accurate estimation of the gradients' average.
We first introduce the homomorphic compression property to model such system constraints. 
Consider $n$ workers and let $\nabla_i$ be the \crupdate{$i$-th worker's gradient}. We define the Uniform Homomorphic Compression (UHC) property as follows:
{
\begin{definition}[Uniform Homomorphic Compression]\label{def:hc}
\end{definition}
\vspace*{-4mm}
{\small
\begin{align*}\mbox{\vspace*{-94mm}}
    \widehat\nabla_{\mathit{avg}} &= \frac{1}{n} \cdot\sum_{i} \texttt{Decompress}(\texttt{Compress}(\nabla_i)) \\&=
    \texttt{Decompress}\parentheses{\frac{1}{n} \cdot\sum_{i} \texttt{Compress}(\nabla_i)}.
\end{align*}
}
}
That is, with the Uniform Homomorphic Compression property, the average of the decompressed gradients is mathematically equivalent to decompressing the average compressed gradient.
Leveraging this property, the PS simply sums the compressed gradients and sends the result back (still in compressed form). 
Finally, each worker averages the result and applies the decompression \mbox{to derive the update $\widehat\nabla_{avg}$.} 

The key challenge for THC is to design compression algorithms that retain the UHC property while ensuring high accuracy.
%argue that these solutions are common ones
% this one doesn't follow UHC but is accurate
Non-homomorphic compression techniques require the PS to decompress the gradients before aggregation because they rely only on worker-local information. When workers use different quantization ranges (e.g., \cite{QSGD,nuqsgd}), the PS must decompress each gradient separately, sum them up, and compress again, increasing processing delay at the PS side and the errors caused by compression. 
We know of one previous compression scheme, SignSGD~\cite{bernstein2018signsgd},
that is homomorphic as it simply counts, for each coordinate, the number of workers which had a positive value for it.  However, this scheme is biased, and thus its error does not decrease with the number of workers, making it yield large errors in practice.  
% another is follows UHC but not accurate
%Or we can rely on biased quantization (e.g., \cite{bernstein2018signsgd}) that :, but these approaches introduce even larger errors in compression. 
In THC, we retain an accuracy that is similar to that of the uncompressed baseline and achieve the UHC property.
% .\minlan{Say a bit more about THC's key theory innovation: build on unbiased quantization by aligning ranges, but allows rotation to imporve accuracy, etc...} 

%**** Instead of this, give high-level ideas
% [stochastic quantization, choice of quantization values, bounding the support]
% Intuitively, this is because THC's quantization error adds variance to the estimate of the average gradient that is smaller than that of the inherent variance in distributed SGD.\minlan{I don't get where's THC error yet} Namely, we optimize the bandwidth-accuracy tradeoff \minlan{why? bandwidth means bit budget. I don't get how THC work and where's the bit budget yet} and pick an operating point in which the estimation accuracy is sufficient. 

We present the THC framework, illustrated in Figure~\ref{fig:AlgorithmOverview}. 
% \minlan{Is there a way we can talk about the design without numbers? The specific numbers are introducing questions on what about other \# bits. Say, call them gradient values and index bits etc. and then talk about the bit numbers in the end}
% \minghao{avoid binding to actual number of bits here. First describe high-level workflow and then get into number of bits}
% This process is visualized in \cref{fig:thc_ps_zoom_in}. 
The THC workflow starts by having workers compress their gradients, which commonly consist of $32$-bit floats. Each coordinate is quantized and then encoded into a \emph{table index} \crupdate{(formally introduced in Section \secref{sec:uhc})} that requires a small number of bits. 
The table indices are then packed and sent to the PS. The PS looks the table indices up in a lookup table to restore the corresponding \emph{table values} and sums up the looked-up table values coordinate-wise. After summing values from all workers, PS packs the result and broadcasts it. Finally, each worker decompresses and normalizes the result to obtain the average gradient's estimate.

The table here serves two purposes: first, it allows an efficient expansion of the indices to wider values that allow summation without overflows. Second, we can use the table to minimize the quantization error, as we later show, by picking the table values in correspondence to the underlying data distribution. \crupdate{Since the table is small (of size $2^b$, where $b$ is a small constant) and hardcoded (does not depend on the gradients or number of workers), and lookups do not require arithmetic operations, we \mbox{consider it as part of the direct aggregation.}}

\begin{figure}[t]
    \centering 
    \includegraphics[clip, trim=0cm 1.9cm 0cm 2.7cm, width=1\linewidth]{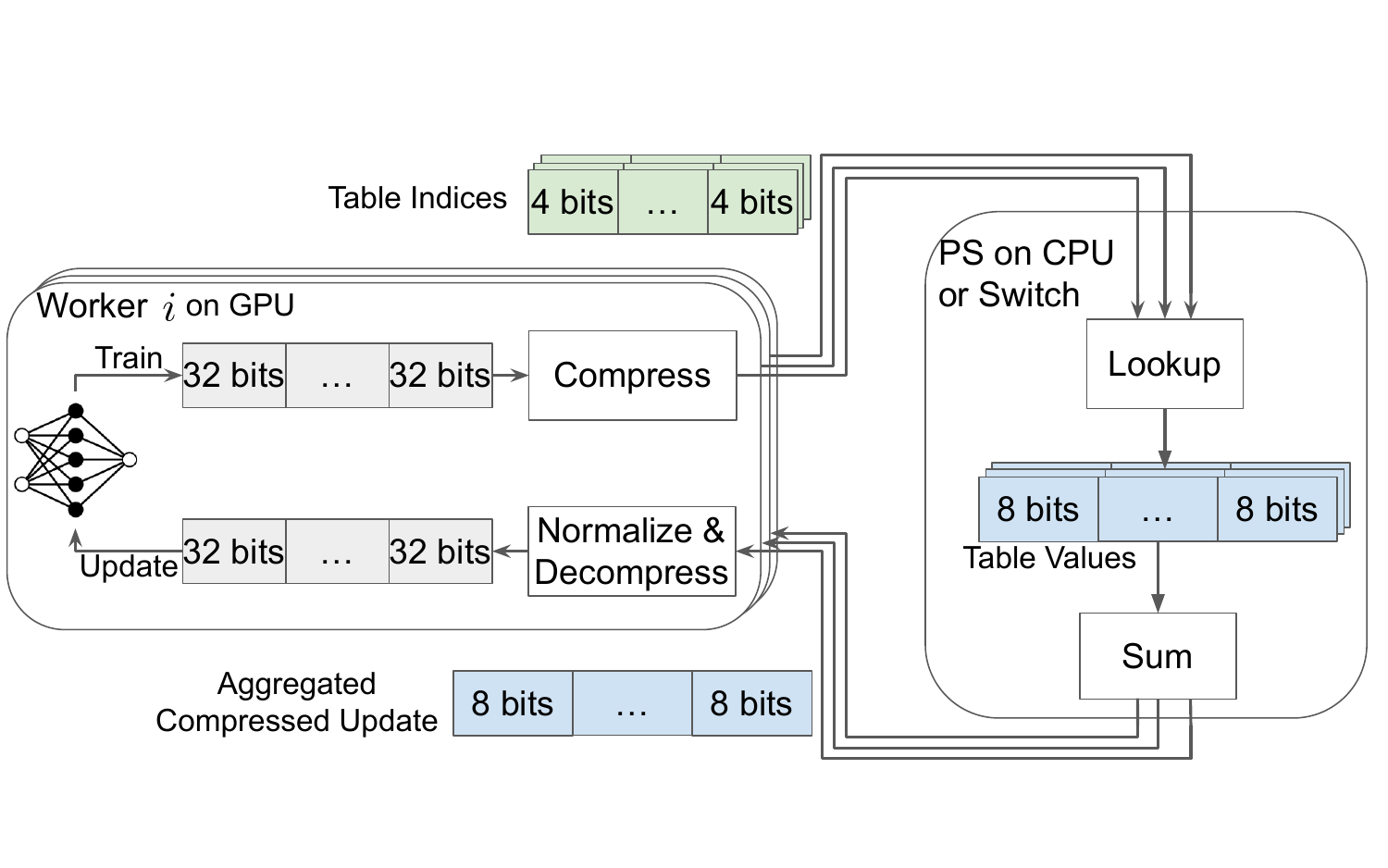} 
    \vspace{-6mm}
    \caption{THC's workflow illustration. %\mitz{Maybe add in the caption why Tables are 8 bits -- e.g., summing 4 bit values and want to avoid overflow}\minlan{Is 32, 4, 8 specific in your algorithm or configurable? Make it clear in figure caption}
    % \ran{We need to add a division by $n$ after the decompression to the figure}
    % \ran{Make figure more vertical space efficient by putting the table indices to the side of the arrows and moving the `aggregated compressed update' text to the left of the vector}
    % \minghao{also on worker change into "Normalize and Decompress"}
    }
    \vspace{-5mm}
    \label{fig:thc_ps_zoom_in}
\end{figure}

Figure~\ref{fig:thc_ps_zoom_in} visualizes the THC implementation we adopt in our system prototype. Namely, each $32$-bit coordinate is encoded into a $4$-bit \emph{table index}, which then gets converted into a $8$-bit \emph{table value} on the PS. The broadcast summation result also uses $8$ bits per coordinate. Therefore, our system prototype provides a $\times 8$ bandwidth reduction from workers to the PS and a $\times 4$ bandwidth reduction in the other direction.

The THC algorithm is described in detail in Section \secref{THC}; in Section~\secref{sec:choiceofparams}, we introduce further optimizations for THC and the end-to-end training procedure; then, Section~\secref{sec:others} shows how THC can be seamlessly used in conjunction with in-network aggregation to further accelerate the training.

\vspace{-2mm}
\section{\! \resizebox{.921\linewidth}{!}{Tensor Homomorphic Compression}}
\label{THC}
\vspace{-2mm}

In this section, we explain how to achieve the UHC property effectively.
We start by giving background on stochastic quantization, a core building block of our THC approach.
We then show (Section \secref{sec:uhc}) that stochastic quantization with uniform intervals, a technique that has been used previously in compression, has the UHC property.  However, its performance in terms of the accuracy per bit is relatively poor, because it is unoptimized. This poor compression performance can lead to worse training time and/or model accuracy than an uncompressed baseline. We, therefore, introduce further optimizations that improve  the compression performance. Our main conceptual advance is to use non-uniform quantization intervals while maintaining the UHC property;  that is, we show how to optimize the choice of quantization values in Section~\secref{sec:thc}.  Finally, Section~\secref{sec:choiceofparams} provides important technical optimizations for speed and accuracy.    

\vspace{-3mm}
\subsection{Background on Stochastic Quantization}
\vspace{-1mm}

A main building block that is used for gradient compression is \emph{quantization}, a technique that allows representing gradient entries (e.g., 32-bit floats) using a small (e.g., 4) number of bits while bounding the error.
At a high level, our Tensor Homomorphic Compression (THC) framework leverages Stochastic Quantization (SQ) that rounds a given real-value $a$ to one of two quantization values $q_0,q_1$ such that $q_0\le a\le q_1$. SQ quantizes $a$ to $q_0$ with probability $(q_1-a)/(q_1-q_0)$ and to $q_1$ otherwise. An appealing property of SQ is that the expected value of the quantization is exactly $a$, i.e., it is unbiased. This is especially useful in distributed scenarios where using SQ results in a decrease in the error of estimating the average as the number of workers increases \cite{vargaftik2021drive}. Our focus in what follows is minimizing the error introduced by quantization while maintaining the UHC property.

\vspace{-3mm}
\subsection{The Uniform THC Algorithm}\label{sec:uhc}
\vspace{-1mm}

As we now show, it turns out that using a variation of SQ with uniform intervals \emph{where all workers use the same set of intervals} already yields a solution that is both unbiased and homomorphic. 
However, the accuracy per-bit of this solution leaves much to be desired, which is where we focus our efforts in the following sections.
 
The most popular form of SQ is \emph{Uniform} SQ (USQ), in which the quantization values are uniformly spaced. For example, given the range $[m,M]$ and using $2^b$ quantization values, their locations are $q_0=m, q_1 = m + (M-m)/(2^b-1),\ldots, q_{2^b}=M$. USQ quantizes a value $a\in[m,M]$ to one of its nearest quantization values.  We first show that USQ, when all workers globally use the same range $[m,M]$ and $b$, satisfies the UHC property. Note that for implementation this requires all workers to first obtain the global minimum and maximum to perform quantization, which is different than the standard use of USQ (where each worker quantizes based on their own local minimum and maximum value). \crupdate{For convenience, we henceforth denote $\angles i=\set{0,\ldots,i-1}$ for any $i\in\mathbb N$.}
%
%Accordingly, we make the following definition:
%
\begin{definition}[Homomorphic USQ]
    Let $\nabla_0,\nabla_1,\ldots,\nabla_{n-1}\in \mathbb R^d$ be the input gradients, and let $m_i=\min \set{\nabla_i}$ and $M_i = \max \set{\nabla_i}$ be the $i$'th gradient's minimum and maximum. Let $m=\min_i \set{m_i}$ and $M=\max_i \set{M_i}$ and consider a set of uniformly spaced quantization values $\set{m+k\cdot\frac{M-m}{2^b-1}\mid k\in\angles{2^{b}}}$. The workers perform stochastic quantization using these quantization values on all input gradients.
\end{definition}%}%(Definition~\ref{def:homomorphic_compression}), since
\vspace{-1mm}
This approach is homomorphic (\cref{def:hc}) since ($C$ and $D$ stand for \texttt{Compress} and \texttt{Decompress}):
%\resizebox{1.0009095\columnwidth}{!}{
%\centering
%%\noindent
%\hspace*{-7.0mm}
%$
%\begin{array}{l}
%\widehat\nabla_{avg} = \frac{1}{n} \cdot\sum_{i} \texttt{D}(\texttt{C}(\nabla_i))= \\ \frac{1}{n} \cdot\sum_{i} \parentheses{m + \texttt{C}(\nabla_i) \cdot \frac{M-m}{2^b-1}} = m + \parentheses{\frac{1}{n} \cdot\sum_{i\in\angles{n}} \texttt{C}(\nabla_i)} \cdot \frac{M-m}{2^b-1}\\
%    = D\parentheses{\frac{1}{n} \cdot\sum_{i\in\angles{n}} \texttt{C}(\nabla_i)}=\texttt{D}\parentheses{\frac{1}{n} \cdot\sum_{i} \texttt{C}(\nabla_i)}.
%\end{array}
%$
%}
%
%{\small
%\begin{multline*}
%    \widehat\nabla_{avg} = \frac{1}{n} \cdot\sum_{i} \texttt{Decompress}(\texttt{Compress}(\nabla_i)) \\= \frac{1}{n} \cdot\sum_{i} \parentheses{m + \texttt{Compress}(\nabla_i) \cdot \frac{M-m}{2^b-1}} = m + \parentheses{\frac{1}{n} \cdot\sum_{i\in\angles{n}} \texttt{Compress}(\nabla_i)} \cdot \frac{M-m}{2^b-1}\\
%    = D\parentheses{\frac{1}{n} \cdot\sum_{i\in\angles{n}} \texttt{Compress}(\nabla_i)}=\texttt{Decompress}\parentheses{\frac{1}{n} \cdot\sum_{i} \texttt{Compress}(\nabla_i)}.
%\end{multline*}
%}%(Definition~\ref{def:homomorphic_compression}), since
\vspace{-2mm}
{\small
\begin{multline*}
\frac{1}{n} \cdot\sum_{i\in\angles{n}} D(C(\nabla_i)) = 
\frac{1}{n} \cdot\sum_{i\in\angles{n}} \parentheses{m + C(\nabla_i) \cdot \frac{M-m}{2^b-1}}\\
= m + \parentheses{\frac{1}{n} \cdot\sum_{i\in\angles{n}} C(\nabla_i)} \cdot \frac{M-m}{2^b-1}
= D\parentheses{\frac{1}{n} \cdot\sum_{i\in\angles{n}} C(\nabla_i)}.
\end{multline*}
}
\vspace{-2mm}

%Note that the choice of using the range $[m,M]$ is optimal since extending it only increases the quantization error, and narrowing it down does not allow for unbiasedness. In practice, computing $m,M$ accurately requires a preliminary communication round. 

%{\bfhttps://www.overleaf.com/project/62efdc85b5832d851ee138d0 MM:  Make clearer here or earlier that uniform refers to uniform intervals, which make stochastic quantization easy, we generalize to non-uniform intervals in what follows.}

With this primitive at hand, we introduce a simplified (uniform) variant of the THC framework, which we generalize to a non-uniform setting to obtain better performance.
%Namely, Uniform THC supports a variety of pre- and post-processing that can work in conjunction with our homomorphic uniform stochastic quantization.
%A vital question is then what families of solutions are compatible with THC, and we identify the class of pre- and post-processing solutions in which all workers use \emph{the same linear transformation} on the input.
%For example, using the randomized Hadamard transform~\cite{hadamardSQ} or Kashin's representation~\cite{kashin1,kashin2} meet this criterion.
% \minlan{What's the key challenge/idea here?}

The pseudo-code of Uniform THC is given by Algorithm \ref{alg:thc_initial}. 
It begins with a preliminary communication round where each worker %preprocesses its vector and then 
computes and sends the smallest and largest gradient entry to the PS (lines \ref{line:miMi}-\ref{line:pre_send}). In turn, the PS computes the extreme global values and distributes them back to the workers (lines \ref{line:comp_mM}-\ref{line:mM}). This communication round is light and requires each worker to transmit and receive only two floats. Next, each worker quantizes its gradient using the global extremes (i.e., perform Homomorphic USQ) and sends the result to the PS (lines \ref{line:comp_usq}-\ref{line:send_usq}). Then, the PS sums all the quantized vectors and sends their sum to the worker (lines \ref{line:compSumXi}-\ref{line:sendSumXi}). Finally, each worker divides the sum by the number of workers to obtain the estimate of the average \mbox{(line \ref{line:comp_avg}).}

%\minlan{Can I view preprocessing and postprocessing as compression and decompression? If not, please explain the differences. In general, it is unclear what you mean by pre-processing and postprocessing? what example functions do you run there?}\ran{We think of them as inversible transform that replaces the input vector with something that is more amenable to low-error compression. We will try to change this if there's time before the deadline. }

As detailed in \cref{sec:choiceofparams}, we can reduce the quantization error by pre-processing the input gradient prior to its quantization and \mbox{post-processing the average's estimate at the end.}
%Some choices of preprocessing allow for an optimization where the preprocessing of the vector (\cref{line:preprocess}) can be done in parallel to the preliminary communication round (lines \ref{line:miMi}-\ref{line:mM}).

%In Homomorphic Stochastic Quantization, the workers first exchange their $\set{m_i,M_i}$ values to calculate the global minimum and maximum .\
%Then, given a (uniform or non-uniform) set of 

% Save the original lengths
\newlength{\oldtextfloatsep}\setlength{\oldtextfloatsep}{\textfloatsep}
\newlength{\oldintextsep}\setlength{\oldintextsep}{\intextsep}

% Set new lengths
\setlength{\textfloatsep}{4pt}
\setlength{\intextsep}{4pt}

\begin{algorithm}[t]
\small
\caption{Uniform THC}
\label{code:alg1}
%\begin{multicols}{2}
\begin{algorithmic}[1]%\vspace*{-8mm}
  \Statex\hspace*{-5mm} \textit{\textbf{Input:}} bit budget $b$\vspace{-1mm}
  \Statex\hspace*{-5mm}\hrulefill\vspace{-1mm}
  \Statex\hspace*{-5mm} \qquad\textit{Preliminary stage}\vspace{-2mm}
  \Statex\hspace*{-5mm}\hrulefill
  \Statex 
  \hspace*{-4mm}\textbf{Worker $i$:}
  %\State $x_i'= \text{Preprocess}(x_i)$\label{line:preprocess}
  \State Compute $m_i=\min \set{\nabla_i}$ and $M_i = \max \set{\nabla_i}$\label{line:miMi}
  \State Send $(m_i,M_i)$ to the PS\label{line:pre_send}
  \Statex 
  \hspace*{-4mm}\textbf{PS:}
  \State Compute $m=\min_i \set{m_i}$ and $M=\max_i \set{M_i}$\label{line:comp_mM}
  \State Send $(m,M)$ to workers\label{line:mM}
  \vspace{-2mm}
  \Statex\hspace*{-5mm}\hrulefill
  \Statex\hspace*{-5mm} \qquad\textit{Main stage}\vspace{-2mm}
  \Statex\hspace*{-5mm}\hrulefill
  \Statex 
  \hspace*{-4mm}\textbf{Worker $i$:}  
  \State Compute $X_i = \mathit{USQ}(\nabla_i,m,M,b)$\label{line:comp_usq}
  \State Send $X_i$ to PS\label{line:send_usq}
  \Statex 
  \hspace*{-4mm}\textbf{PS:}
  \State Compute $X = \sum_{i\in\angles{n}} X_i$\label{line:compSumXi}
  \State Send $X$ to workers\label{line:sendSumXi}
  \Statex 
  \hspace*{-4mm}\textbf{Worker $i$:}  
  \State Estimate $\widehat\nabla_{\mathit{avg}}  = m + \frac{1}{n} \cdot X \cdot \frac{M-m}{2^b-1}$\label{line:comp_avg}
  %\State $\widehat x_{\mathit{avg}} = \text{Postprocess}(\widehat x_{\mathit{avg}}')$\label{line:postprocess}
\end{algorithmic}
\label{alg:thc_initial}
\end{algorithm}

% Restore the original lengths
%\setlength{\textfloatsep}{\oldtextfloatsep}
%\setlength{\intextsep}{\oldintextsep}

\vspace{0mm}
\subsection{The Non-uniform THC Algorithm}\label{sec:thc}
\vspace{-2mm}
%\minlan{start with a transition from uniform to non-uniform}
In many cases, it is possible to choose the quantization values in a non-uniform manner to optimize the accuracy-bandwidth tradeoff. %\minlan{explain more}
However, it is unclear how to leverage existing non-uniform quantization methods (e.g.,~\cite{nuqsgd,hadamardSQ,eden,basat2022quick,ben2024optimal}) to improve our homomorphic compression.
Namely, in non-uniform methods, the sender transmits a \emph{table index} $z$ that is then converted to the \emph{table value} $T[z]$ by the receiver. Here, $T$ is a \emph{lookup table} that converts indices to values that may not be uniformly spaced, i.e., $T[z+1]-T[z]$ may not equal $T[z'+1]-T[z']$ for different indices $z,z'$.

For example, consider quantizing values in the range $[-1,1]$ using four quantization values. 
Using USQ, the lookup table is \crupdate{$T_0[0]=-1, T_0[1]=-1/3,T_0[2]=1/3,T_0[3]=1$.}
%\minlan{We need a figure for this}
%
%Using USQ, the table indices are $\angles{4}=\set{0,1,2,3}$ and 
%the values are $\set{T[z]\mid z\in\angles{4}}=\set{-1,-1/3,1/3,1}$. 
%
In this case, any sum of table indices corresponds to a single sum of quantization values. 
For example, suppose two senders send indices $z,z'$ and consider two cases: (1) $z=0, z'=3$ and (2) $z=1,z'=2$. In both cases, \crupdate{$T_0[z]+T_0[z']=0$}. That is, $z+z'=3$ implies that \crupdate{$T_0[z]+T_0[z']=0$}. Intuitively, the receiver can sum the indices and deduce the sum of sent values instead of performing two table lookups. 

%\ran{We are here.}
In contrast, consider non-uniform quantization that uses the table \crupdate{$T_1[0] = -1, T_1[1]=-1/2, T_1[2]=1/2, T_1[3]=1$}.
%There are multiple sums of two table values that yield the same sum of indices, e.g., 
Consider the two cases in which $z+z'=4$: (1) $z=1,z'=3$ and (2) $z=z'=2$. In the first case, \crupdate{$T_1[z]+T_1[z']=1/2$} while in the second \crupdate{$T_1[z]+T_1[z']=1$}. That is, the sum of table indices does not determine the sum of table values.

Our insight is that we can overcome this by using a subset of the uniformly spaced quantization values.  
% Our insight is that we can overcome this by mapping each message to a different representation that is homomorphically aggregable. \minlan{say a bit more highlevel on what this representation is. This part needs another figure}
To that end, for a bit-budget of $b$ bits per coordinate, we define a hyperparameter $g\ge 2^b-1$ called \emph{granularity}. 
%This parameter defines the uniform quantization values $\angles{g+1}$ from which we are allowed to pick a subset of $2^b$ non-uniform quantization values.
We then use a table $T:\angles{2^b}\to\angles{g+1}$ that maps table indices to values that are integers in the larger range. Intuitively, one can think about this as running USQ with $g+1$ quantization values, but where all senders are only allowed the same $2^b$ indices.
A table value $T[z]$ then corresponds to the quantization value \crupdate{$m + T[z]\cdot \frac{M-m}{g+1}$}, same as in USQ.
For instance, in the above example \crupdate{of quantizing values in the range $[-1,1]$}, the PS can use a table \crupdate{$T_2[0]=0, T_2[1]=1, T_2[2]=3, T_2[3]=4$} 
to map each table index into (the larger) range $\angles{5}$. The benefit is that the table values are now directly aggregable; 
for example, consider three senders with the two cases (1) $z=z'=z''=1$ and thus \crupdate{$T_2[z] = T_2[z'] = T_2[z''] = 1$}, i.e., all quantization values are \crupdate{$-1+\frac{1-(-1)}{4}=-\frac{1}{2}$}; (2) $z=z'=0, z''=2$ and thus \crupdate{$T_2[z]=T_2[z']=0, T_2[z'']=3$}, i.e., the first two quantization values are $-1$ and the third is \crupdate{$-1+3\times\frac{1-(-1)}{4}=\frac{1}{2}$}.  Notice that the sum of table values is the same in both cases, and so is the sum of quantization \mbox{values; in contrast, the sum of table indices differ.}
%
%
%In general, our framework can use any table whose values include $0$ and $g$ to allow for unbiasedness. 
%The mapping then associates each message with a value in this set of values.
%In the above example, $g=4$, we mapped into the values $\set{0,1,3,4}\subsetneq\angles{g+1}$, where the message $0$ is mapped to $0$ and the message $3$ is mapped to $g$.

{
Intuitively, $g$ introduces a tradeoff where larger $g$ results in more fine-grained quantization values and lower error but also requires more bits to represent the summation and higher bandwidth requirement from the PS or switch to the workers. 
}

The pseudocode for the non-uniform variant of THC appears in~\cref{code:alg2} (the Preliminary stage is the same as in~\cref{code:alg1}). Notice that the lookup table $T_{b,g}$ depends on bit-budget $b$ and the granularity $g$. When clear from context, we omit the subscripts and write $T$.

Each worker $i$ then calculates the set of quantization values $Q\subset [m,M]$ given the (global) $m,M$ and the granularity $g$ (\cref{line:alg2:comp_q}).
Next, it stochastically quantizes each coordinate 
%in its pre-processed vector $x_i'$ 
to a value in $Q$, giving the result $X_i\in Q^d$ (\cref{line:alg2:comp_sq}). The worker then transforms each quantized coordinate into its uniform-HC value in $\angles{g+1}$ using the linear transformation (\cref{line:alg2:comp_Y}).
The next stage involves computing which $b$-bit table indices (in $\angles{2^{b}}$) would map to the uniformly partitioned values by applying the inverse mapping (\cref{line:alg2:comp_Z}). 
\mbox{Finally, it sends the result $Z_i$ to the PS (\cref{line:alg2:send_Z}).}

The PS then gets the set of vectors $\set{Z_i}_{i\in\angles{n}}$. It looks up each vector (coordinate-wise) and sums them up, corresponding to the sum of the $Y_i$'s (\cref{line:alg2:compsumZi}). Then, the PS sends the result, $Y$, to the workers, still in compressed form (\cref{line:alg2:sendSumXi}). Observe that the PS only performs table lookups and summation without decompressing the vectors and without additional processing and re-compression that increases the error.

The final part takes place in parallel, where % at the workers.
each worker $i$ first computes the average of $\sum_{i\in\angles{n}} Y_i$ by dividing $Y$ by $n$ (\cref{line:alg2:compsumYavg}).
It then applies the inverse transformation to \cref{line:alg2:comp_Y} to \mbox{obtain an estimate $\widehat\nabla_{\mathit{avg}}$ of the average gradient (\cref{line:alg2:comp_avg}).
}
% to update its model (\cref{line:alg2:postprocess})
%Last, each worker post-processes $\widehat x_{\mathit{avg}}$ to estimate the average gradient and can then update its model (\cref{line:alg2:postprocess}).

%a $b$-bits message per coordinate $j$: $Z_i[j]\in\set{0,\ldots,2^b-1}$.
%The PS will then map the message using a lookup table $T_g$, such that the aggregated value is $T_g[Z_i[j]]\in\set{0,\ldots,g}$. Here

%\ran{work in progress... }

%The pseudo-code is given by Algorithm~\ref{alg:thc_advanced}.
%Given a lookup-table $T:\set{0,\ldots,2^b-1}\to\set{0,\ldots,g}$, we generalizes our UHC definition.

To show that our algorithm is homomorphic, we generalize \mbox{\cref{def:hc} to account for the lookup table} ($C,D$ and  $T$ stand for \texttt{Compress}, \texttt{Decompress} and \texttt{table lookup})
\vspace{-1mm}
\begin{definition}[{ Non-uniform homomorphic compression}]\label{def:nu_homomorphic_compression}
%A non-uniform compression scheme $(C,D,T)$ is \emph{homomorphic} if for any $n,d\in\mathbb N$ and  $\nabla_0,\nabla_1,\ldots,\nabla_{n-1} \in \mathbb R^d$,  it satisfies 
\end {definition}\vspace{-1.5mm}
$$\widehat\nabla_{\mathit{avg}} =\frac{1}{n} \cdot\sum_{i\in\angles{n}} D(T(C(\nabla_i))) = D\parentheses{\frac{1}{n} \cdot\sum_{i\in\angles{n}} T(C(\nabla_i))}.$$ 
That is, a non-uniform HC (NUHC) scheme generalizes UHC by allowing the PS to apply a lookup table $T$ to the compressed gradients before aggregating them.

Notice that if $g=2^b-1$ and $T$ is the identity mapping, NUHC is identical to UHC (in that case, the lookup table is redundant).
As shown above, for $T$ that is the identity mapping, the compression is uniform homomorphic.  We now show that if $0=T(0) < T(1)<\ldots< T(2^b-1)=g$ then Algorithm~\ref{alg:thc_advanced} is homomorphic.
\footnote{In fact, it is sufficient that $T$ is injective and satisfies $0,g\in Im(T)$; the above definition is without loss of generality.
} This is because\vspace{-2mm}
{\small
\begin{multline*}
\frac{1}{n} \cdot\sum_{i\in\angles{n}} D(T(C(\nabla_i))) = 
\frac{1}{n} \cdot\sum_{i\in\angles{n}} D(T(Z_i)) = \\
\frac{1}{n} \cdot\sum_{i\in\angles{n}} D(Y_i) = 
\frac{1}{n} \cdot\parentheses{\sum_{i\in\angles{n}} m + Y_i\cdot \frac{M-m}{g}} = \\ 
m + \parentheses{\frac{1}{n} \cdot\sum_{i\in\angles{n}} Y_i}\cdot \frac{M-m}{g} = 
D\parentheses{\frac{1}{n} \cdot\sum_{i\in\angles{n}} T(C(\nabla_i))}\ ,
\end{multline*}
}

\noindent where we used (i) $C(\nabla_i)=Z_i$, (ii) % is the message sent by the worker, 
$T(C(\nabla_i))=T_{b,g}[Z_i]=Y_i$% is the vector after translation by the PS
, and  (iii) $D(Y_i)=m + Y_{i}\cdot \frac{M-m}{g}$.

The above result shows that \cref{alg:thc_advanced} is homomorphic for many choices 
%of pre- and post-processing procedures and 
of lookup tables. 
%In the following subsections, we describe and motivate our design for these building blocks. Finally, 
As we later demonstrate, very small lookup tables (e.g., for $b=4$ we can usually use $g\in\set{16,\ldots,51}$) are sufficient to obtain accurate quantization.

%\minlan{Mention that a small table size is good enough mostly}

\begin{algorithm}[t]
\small
\caption{Non-uniform THC}
\label{code:alg2}
%\begin{multicols}{2}
\begin{algorithmic}[1]%\vspace*{-8mm}
  \Statex\hspace*{-5mm} \textit{\textbf{Input:}} bit budget $b$, granularity $g$,  and their lookup-table $T_{b,g}$ \vspace{-2mm}
  \Statex\hspace*{-5mm}\hrulefill\vspace{-4mm}
  \Statex 
  \Statex\hspace*{-5mm} \qquad\textit{Main stage} \Comment{Preliminary stage same as in Algorithm \ref{alg:thc_initial} } \vspace{-2mm}
  \Statex\hspace*{-5mm}\hrulefill
  % \hspace*{-4mm}\textbf{Worker $i$:}
  % \State $x_i'= \text{Preprocess}(x_i)$\label{line:alg2:preprocess}
  % \State Compute $m_i=\min \set{x_i'}$ and $M_i = \max \set{x_i'}$\label{line:alg2:miMi}
  % \State Send $(m_i,M_i)$ to the PS\label{line:alg2:pre_send}
  % \Statex 
  % \hspace*{-4mm}\textbf{PS:}
  % \State Compute $m=\min_i \set{m_i}$ and $M=\max_i \set{M_i}$\label{line:alg2:comp_mM}
  % \State Send $(m,M)$ to workers\label{line:alg2:mM}
  \Statex 
  \hspace*{-4mm}\textbf{Worker $i$:}  
  \State Compute $Q = \mathit{CalcQuantizationValues}(m,M,T_{b,g})$\label{line:alg2:comp_q}
  \State Compute $X_i = \mathit{SQ}(\nabla_i,Q)$\label{line:alg2:comp_sq}
  \State Compute $Y_i = (X_i - m) \cdot \frac{g}{M-m}$ \Comment{$Y_i\in \angles{g+1}^d$}\label{line:alg2:comp_Y}
  \State Compute $Z_i = T_{b,g}^{-1}[Y_i]$\Comment{ $Z_i\in \angles{2^{b}}^d$ }\label{line:alg2:comp_Z} 
  \State Send $Z_i$ to PS\label{line:alg2:send_Z}
  \Statex 
  \hspace*{-4mm}\textbf{PS:}
  \State Compute $Y = \sum_{i\in\angles{n}} T_{b,g}[Z_i]$\label{line:alg2:compsumZi}\Comment{$Y\in \angles{g\cdot n+1}^d$}
  \State Send $Y$ to workers\label{line:alg2:sendSumXi}
  \Statex 
  \hspace*{-4mm}\textbf{Worker $i$:}  
  \State Compute $Y_{\mathit{avg}} = \frac{1}{n} \cdot Y$\label{line:alg2:compsumYavg}\Comment{$Y_{\mathit{avg}}\in [0,g]^d$}
  \State Estimate $\widehat\nabla_{\mathit{avg}} = m + Y_{\mathit{avg}} \cdot \frac{M-m}{g}$\label{line:alg2:comp_avg}
  %\State $\widehat x_{\mathit{avg}} = \text{Postprocess}(\widehat x_{\mathit{avg}}')$\label{line:alg2:postprocess}
\end{algorithmic}
\label{alg:thc_advanced}
\end{algorithm}

\vspace*{-2mm}
\section{Optimizing THC}\label{sec:choiceofparams}
%\minlan{give the framework from sec 4 here in the preamble, and tell us where 5.1-5.3 fits in the framework}
We next describe optimizations that improve THC's bandwidth-accuracy tradeoff (\S\ref{sec:RHT}-\S\ref{sec:lookup}) and speed (\S\ref{ref:truncation}).
%Heretofore, we describe the general THC framework, that allows a wide range of pre-processing procedures and lookup tables. Here, we \mbox{describe and motivate our specific choices.}
%\minlan{What're your specific choices based on? what are your optimization objectives as opposed to other choices?}

\vspace*{-2mm}
\subsection{Pre- and Post-processing Using the Randomized Hadamard Transform}\label{sec:RHT}
\crupdate{For pre-processing, we utilize the Randomized Hadamard Transform (RHT) that improves quantization accuracy by reducing the expected range and transforming coordinates to approach a normal distribution.} The RHT of a vector $x\in\mathbb R^d$ is defined as $\frac{1}{\sqrt d}\cdot H\cdot D\cdot x$, where $H$ is the Hadamard matrix~\cite{hedayat1978hadamard} and $D$ is a diagonal matrix with i.i.d. Radamacher variables (taking $\pm 1$ with equal probabilities) on its diagonal.
An important property RHT is that the special recursive structure of $H$ allows a fast GPU-friendly $O(d\log d)$ time implementation, significantly faster than general matrix multiplication. The post-processing is then the inverse transform, i.e., $RHT^{-1}(x)=\frac{1}{\sqrt d}\cdot D\cdot H\cdot x$, \mbox{which has identical complexity to RHT.}

RHT has two key benefits: First,  RHT reduces the range of coordinate values, improving accuracy.  More concretely, a key quantity that determines the error of a quantization scheme is its range (the difference between the largest and smallest quantization values), $M-m$ (see~\cref{sec:usq} for more details).
Intuitively, when the range is large, one is forced to quantize to values that are further away from the encoded quantity, thus increasing the error.
As proven by~\cite{ailon2006approximate}, and using $M',m'$ to denote the maximal and minimal value after the RHT transform,
$\mathbb E[M'-m']=O\parentheses{(M-m)\cdot \sqrt{{\log d}\ /\ {d}}}$. This decrease in the expected range significantly improves the quantization accuracy.

To further decrease the range, and thus the quantization error, we leverage known results about the distribution of the transformed coordinates to derive a threshold $t_p$\crupdate{, for an appropriate $p\in (0,1)$,} such that approximately a $p$ fraction of the transformed coordinates are expected to fall outside $[-t_p,t_p]$.  Namely, it is known that each coordinate in the RHT of a vector $x\in\mathbb R^d$ follows a distribution that approaches (for a large enough $d$) the normal distribution $\mathcal N(0,1/\norm{x}^2)$~\cite{vargaftik2021drive}. In particular, this means that the probability of a coordinate landing outside the range diminishes exponentially in $t_p$, giving an opportunity to significantly reduce the range at the expense of a small bias and
allowing us to pick an appropriate $p$, as we describe below in \S~\ref{ref:truncation}. Our algorithm then optimizes the quantization values to minimize the error of the coordinates in $[-t_p,t_p]$, and truncates the rest by rounding those larger than $t_p$ to $t_p$ and those smaller than $-t_p$ to $-t_p$.

To address this small bias, we compensate for it using a technique called error-feedback (EF)~\cite{ErrorFeedback}, which includes sending the vector $x=\nabla + e$ that adds the previous error $e$ to the current gradient $\nabla$, and later updating $e$ to account for the quantization error.
It is known that when the bias is not too large, EF guarantees the convergence of the training~\cite{li2021distributed}. \crupdate{The effect of rotation and error feedback is evaluated in detail in Appendix \ref{app:optimizations}, along with comparison to uniform THC}. 

%To pick a single $t_p$ value that ensures that at most a $p$ fraction of the coordinates fall outside the range with high probability, we can to consider the norms of the workers' gradients. 

%Because of this, if we use the RHT, we can construct an offline table for this specific distribution that will perform well regardless of the initial data (and its distribution) at the workers.

%Intuitively, the distribution has tails with an exponentially diminishing probability, and by restricting the range of coordinates on which we are unbiased, we can significantly reduce the error.

%Namely, for a parameter $p\in(0,1)$, each worker $i$ ignores the extreme (i.e., in absolute value) $p$-fraction of the coordinates in pre-processed gradient when determining $m_i$ and $M_i$.
%After getting the global $m$ and $M$, the worker then clamps (i.e., truncates) the vector to $[m,M]$ (i.e., all coordinates smaller than $m$ increase to $m$ those larger than $M$ decrease to $M$).
%To reduce the support of this distribution, and as only an exponentially diminishing fraction of the coordinates fall far from $0$, we bound the support of the transformed coordinates. 
%
The second key benefit of RHT is that, since the distribution of the coordinates after applying RHT is known, we can compute \mbox{the optimal lookup table $T$ \emph{offline}, as we explain next.}

\vspace*{-5mm}
\subsection{Constructing the Optimal Lookup Table}\label{sec:lookup}
  
%As a further optimization, we change the distribution slightly to bound the support of the transformed coordinates.
%As before, this reduces the error,

%As a result, the error for the values in $[m,M]$ decreases, but we introduce a bias for coordinates outside this range. 
%For example, if the vector $(-100,-1,0,1,100,\allowbreak 111)$ and $p=0.5$, we get $m=-1, M=1$, and thus the sender clamps the vector to $(-1,-1,0,1,1,1)$ before quantizing it. 
\crupdate{As explained, after applying RHT and truncating, our goal is to design a lookup table that minimizes the quantization error of the resulting vector. This vector has two types of coordinates: (1) \emph{the truncated coordinates} have no error (beyond the truncation) since, by design, there are always quantization values at $\set{-t_p,t_p}$; (2) the \emph{non-truncated coordinates} have a distribution that approaches the truncated normal distribution. Intuitively, this allows us to design the lookup table $T$ to minimize the quantization error of a truncated normal random variable and thereby minimize the NMSE. 

Formally, the optimal lookup table needs to minimize the error in quantizing a truncated normal random variable $(A\sim\mathcal N(0,1) \mid A\in[-t_p,t_p])$, 
%quantizing values in the range $[-t_p, t_p]$ 
where $t_p = \Phi^{-1}(1-p/2)$ and $\Phi$ is the CDF of the normal distribution.} As we later show, we can use this pre-computed table for a normal random variable \emph{with any variance} by scaling the quantization values.
%The expected range (i.e., $M'-m'$) further reduces to $O\parentheses{(M-m)/ \sqrt{\ {d}}}$, thus decreasing the quantization error.  

%\minlan{Start with a summary of what table you want to construct and what's the goal: define optimal}
%As mentioned, after RHT, the vectors' values have a distribution that approaches the normal distribution. 
%We thus seek to find a lookup table that minimizes the variance for quantizing a normal $\mathcal N(0,1)$ random variable $A$, truncated to $[-t_p,t_p]$.
% However, each worker has a different norm. To ensure that at most a $p$-fraction of the coordinates is truncated at each worker, we 
% we use the range $[-t_p,t_p]$. 
%Also, since the data distribution in distributed training is i.i.d. (e.g., by data reshuffling),
% the norms of the gradients are likely to be similar.
%Because of this concentration, we choose to optimize the quantization values for the worker with the maximal norm. 
Formally, denoting by $P(a,z)$ the probability of sending the index $z\in\angles{2^b}$ given a value $a\in[-t_p,t_p]$, the optimization problem aims to find the table $T$ and probabilities $P$.
Let us further denote by $\phi$ the pdf of the normal distribution. 
Then, the following optimization problem gives the optimal lookup table as~$T_{b,g,p}[z]=T[z]$:

%$$ T_{b,g,p}\ceil{z} = \min\set{T_{b,g,p}[a] \mid T_{b,g,p}[a] \ge z}$$
%$$ T_{b,g,p}\floor{z} = \max\set{T_{b,g,p}[a] \mid T_{b,g,p}[a] \le z}$$

\resizebox{.99095\columnwidth}{!}{
\centering
%\noindent
\hspace*{-7.0mm}
$
\begin{array}{l}
\displaystyle{\minimize_{P,T}}  \displaystyle \int_{-t_p}^{t_p} \sum_{z\in \angles{2^{b}}} P(a,z) \cdot \parentheses{a-T[z]}^2 \cdot \phi(a)\cdot  da\\\medskip
\text{subject to}\\\medskip
{\small (\textit{\textcolor{gray}{Unbiasedness}})} \displaystyle\ \  \sum_{z \in \angles{2^{b}}} P(a,z) \cdot T[z] = a  \quad\hfill\forall\, a \in [-t_p, t_p]\\\medskip
{\small (\textit{\textcolor{gray}{Probability}})}\,\ \quad \displaystyle \sum_{z\in \angles{2^{b}}}P(a,z)=1 \hfill\forall\, a \in [-t_p, t_p]\\\medskip
\quad\quad\quad\quad\quad\quad\,\,\, P(a,z)\ge0 \hfill \forall\, a \in [-t_p, t_p],\, z\in \angles{2^{b}}\\\medskip
{\small (\textit{\textcolor{gray}{Granularity}})}\quad\,\,\, T[z]\in \set{\frac{2t_p}{g}\cdot i  - t_p\mid i \in \angles{g+1}}\hfill \quad\forall\, z\in \angles{2^{b}}
\end{array}
$
}

%The problem minimizes the expected squared error for quantizing a normal $\mathcal N(0,1)$ random variable $A$, truncated to $[-t_p,t_p]$. Given $A=a$, $P(a,z)$ denote the probability of sending the table index $z$ whose corresponding table value is $T[z]$.
%\mitz{This is convincing me that $T[z]$ should be called table value and not bucket.}

As we elaborate in Appendix~\ref{app:lookup}, we optimally solve the above problem. \crupdate{To that end, we wrote a specialized ILP solver that leverages various properties of the above optimization problem to reduce the search space and speed up the computation of $T$.} Recall that for any $b,g,p$, we compute the optimal $T_{b,g,p}$ table only once offline  and thus the solver's runtime does {not affect THC's performance.}

%\begin{algorithm}[t]
%\small
%\caption{Pre- and post-processing at worker $i$}
%\label{code:alg1}
%%\begin{multicols}{2}
%\begin{algorithmic}[1]%\vspace*{-8mm}
%  %\Statex\hspace*{-5mm} \textit{\textbf{Input:}} bit budget $b$\vspace{-1mm}
%  \Statex\hspace*{-5mm}\hrulefill\vspace{-1mm}
%  \Statex\hspace*{-5mm} \qquad\textit{Preprocessing $(x)$}\vspace{-2mm}
%  \Statex\hspace*{-5mm}\hrulefill
%%  \State $\text{If EF-enabled:\quad} x+ e$ 
%  \State $R = \texttt{RHT}(x)$\Comment{ $R= HDx$ }  
%  \State\Return $\texttt{clamp}(R, \min=m, \max=M)$
%  \Statex\hspace*{-5mm}\hrulefill
%  \Statex\hspace*{-5mm} \qquad\textit{Postprocessing $(x)$}\vspace{-2mm}
%  \setcounter{ALG@line}{0}
%  \Statex\hspace*{-5mm}\hrulefill
%%  \State $\text{If EF-enabled:\quad} e = x_i - Postprocessing(X_i)$
%  \State\Return $\texttt{RHT}^{-1}(x)$\Comment{ $DH x$ }
%\end{algorithmic}
%\label{alg:prepost}
%\end{algorithm}

    %\item 
%\end{itemize}
\vspace*{-1mm}
\subsection{Accelerating the Preliminary Stage}\label{ref:truncation}
Heretofore, we discussed a preliminary stage in which the workers exchange information that depends on the transformed vectors to determine the quantization range. A natural implementation would, therefore, include transforming the vectors using RHT and then exchanging the information required for setting $M$ and $m$.
%truncation values (i.e., $t_p$) of the different workers. 

Instead, we leverage special properties of RHT, namely that (i) it preserves the \emph{norm} of the transformed vector and (ii) that there is a tight connection between the \emph{maximal norm} and the values $M,m$ we are seeking. Accordingly, each client $i$ first computes the norm of its vectors $x_i$ and then parallelizes the following operations: performing the RHT and obtaining the maximum norm among the workers' gradients using the PS. Then it can set $M = (t_p/\sqrt{d}) \cdot \ell, m=-M$ where $ \ell = \max \norm{x_i}^2$ and proceed to the quantization.

\newcommand{\learningColor}{brown}
\newcommand{\prelimColor}{cyan}
\newcommand{\preprocColor}{magenta}
\newcommand{\postprocColor}{purple}
\newcommand{\mainTHCColor}{blue}

\vspace*{-2mm}
\subsection{Putting It All Together}
We are now ready to describe our complete THC training process, whose pseudocode is given~\cref{alg:general_docoFL}, color coding the different steps.
In the first \textcolor{\learningColor}{learning} step, each worker computes its local gradient and adds the error-feedback. Next, the \textcolor{\prelimColor}{preliminary stage}, in which the clients exchange their norms, is parallelized with the RHT part of the
\textcolor{\preprocColor}{pre-processing}; later, the transformed vector is normalized and clamped.
Next, during \textcolor{\mainTHCColor}{main stage}, the workers and PS follow our \crupdate{Non-uniform THC algorithm (see Algorithm~\ref{alg:thc_advanced})}  to obtain an estimate of the average of the pre-processed vectors. Then, each worker \textcolor{\postprocColor}{post-processes} the estimate using the inverse transform.
Finally, in the second \textcolor{\learningColor}{learning} \mbox{step, the workers update the error-feedback and model. }

%We get that the (truncated) range reduces to $O\parentheses{\norm{x}\sqrt{\frac{1}{d}}}$.

\algrenewcommand\algorithmicprocedure{\textbf{in parallel}}

\setlength{\textfloatsep}{10pt}
\setlength{\intextsep}{10pt}

{\small
\begin{algorithm}[t]
\caption{Training with THC}\label{alg:general_docoFL}
% \DontPrintSemicolon
% \KwInput{Learning rate $\eta$;
% Global/local learning rates $\eta_g, \eta_{\ell}$;
\begin{algorithmic}[1]
    \State {Input:} Support parameter $p$ and its threshold $t_p$. Bit budget $b$, granularity $g$, and lookup-table $T_{b,g,p}$. 
    \For{$r=0,1,\ldots$}
    %\For{$t\in\angles{\infty}$}
        \For{worker $i\in \angles{n}$ in parallel}
            \State \textcolor{\learningColor}{Compute local gradient, $\nabla_i$}
            %\State If EF-enabled, 
            \State \textcolor{\learningColor}{$x_i = \nabla_i + e^r_i$} \Comment{Error-feedback for round $r$}
            \Procedure {}{}
            \State \textcolor{\prelimColor}{(i) \textbf{Send} $\norm{x_i}$ to PS}
            \State \hspace{3.56806mm}\textcolor{\prelimColor}{\textbf{Receive} $\ell=\max_i{\norm{x_i}}$ from PS}
            \State \textcolor{\preprocColor}{(ii) $R_i = \texttt{RHT}(x_i)$}\Comment{ $R= HD^rx_i$ }  
            \EndProcedure
              \State\textcolor{\preprocColor}{ $M= (t_p/\sqrt d) \cdot \ell~~;~~$$m= -M$}
              \State\textcolor{\preprocColor}{ $x_i' = \texttt{clamp}(R_i, \min=m, \max=M)$ }\Comment{Truncation}
              \State\textcolor{\mainTHCColor}{ $Q = \mathit{CalcQuantizationValues}(m,M,T_{b,g,p})$}
              \State\textcolor{\mainTHCColor}{ $X_i = \mathit{SQ}(x_i',Q)$}\Comment{Stochastic Quantization}
              \State\textcolor{\mainTHCColor}{ $Y_i = (X_i - m) \cdot \frac{g}{M-m}$ }\Comment{$Y_i\in \angles{g+1}^d$}
              \State\textcolor{\mainTHCColor}{ $Z_i = T_{b,g, p}^{-1}[Y_i]$}\Comment{ $Z_i\in \angles{2^{b}}^d$ }
              \State\textcolor{\mainTHCColor}{ \textbf{Send} $Z_i$ to PS}
              \State\textcolor{\mainTHCColor}{ \textbf{Receive} $Y = \sum_{i\in\angles{n}} Y_i = \sum_{i\in\angles{n}} T_{b,g,p}[Z_i]$ from PS} 
              \State\textcolor{\mainTHCColor}{ Compute $Y_{\mathit{avg}} = \frac{1}{n} \cdot Y$}\Comment{$Y_{\mathit{avg}}\in [0,g]^d$}
              \State\textcolor{\mainTHCColor}{ Estimate $\widehat x_{\mathit{avg}}' = m + Y_{\mathit{avg}} \cdot \frac{M-m}{g}$}
              \State\textcolor{\postprocColor}{ Compute $\widehat\nabla_{\mathit{avg}} = \texttt{RHT}^{-1}(\widehat x_{\mathit{avg}}')$ }\Comment{Global update}
              \State\textcolor{\learningColor}{ $e^{r+1}_i = x_i - \texttt{RHT}^{-1}(X_i)$}               \Comment{Quantization error}
              \State\textcolor{\learningColor}{ Update model using $\widehat\nabla_{\mathit{avg}}$}
        \EndFor
    \EndFor
    % \State {\bfseries Output:} $x_s\sim\text{Uniform}(x_1,\ldots, x_T)$
\end{algorithmic}
\end{algorithm}
}
%-------------------------------------------------------------------------------

%-------------------------------------------------------------------------------

\section{THC with Other System Opportunities}\label{sec:others}
% \section{System Support for THC}\label{sec:others}

% In this section, we discuss how \sysname leverages in-network aggregation and handles packet loss and stragglers.
In this section, we discuss how \sysname leverages the opportunities of in-network aggregation and explore potential optimizations to address issues such as packet loss and stragglers.

{\bf Aggregation at Programmable Switches}
In our THC framework, we can offload PS completely to programmable switches for further hardware acceleration.
Our THC design simplifies the PS by removing the compression and decompression operations. Since THC already compresses floating-point gradients to integer table indices, it fits programmable switches well. We do not need additional conversions from floating points to integers at workers as used in previous~work~\cite{SwitchML, ATP, unlockswitch}.

% \lam{this paragraph below is not actually related to switch or hardware limitation according to minghao, this is more related to the scalability issues (number of workers), maybe we should drop below?}
% \lam{seems like this overflow here is completely different from switch overflow issue}
% \mitz{Below is still unclear.  Are you saying the user has to pick the right number of bits to avoid overflow.  State what happens when there is an overflow.}
% Overflow is another problem of concern in previous switch solutions~\cite{SwitchML, ATP}. In \sysname, we make the bit budget per table value a configurable parameter. Users can choose the number of bits based on the number of workers they intend to deploy to avoid integer overflow, considering hardware limitations. Note that THC doesn't impose any constraint on the number of bits to use for table values and indices. For the simplicity of design, here we use eight bits per table value and four bits per table index in our system prototype.

{\bf Packet Loss and Stragglers}
% \sysname is a compression algorithm that can tolerate data loss, allowing it to be directly applied to previous techniques that tolerate to data loss, like packet loss~\cite{wang2020domain} or straggler problems~\cite{usingtrio, esa} that can usually be seen during the training in data center. 
%\lam{need straggler problem citation that is not particular in programmable switch}%, which are commonly encountered during training in data centers.
% \sysname is a compression algorithm that can tolerate data loss, allowing it to be directly applied to existing research that tolerate to data loss, offers another path to solve the packet loss~\cite{wang2020domain} or straggler problems~\cite{usingtrio, esa} by accepting lossy training. 
\sysname is a compression algorithm that tolerates data loss, allowing it to ignore the tail outliers that can negatively impact performance during training introduced by packet loss~\cite{wang2020domain} or straggler problems~\cite{usingtrio, esa}.

Packet losses and stragglers may occur between workers and the PS. For each gradient, if a worker doesn't receive the corresponding aggregation result packet within a specified time threshold, the worker could fill in the missing data with zeros and continue with the received aggregation results.
This practice may lead to some workers updating their models with different information. To mitigate the impact, we can implement a synchronization scheme, where workers coordinate their model parameters after every epoch by choosing to copy the parameters of another worker when encountering severe packet loss. Our simulation results shows that there is no significant impact on THC model accuracy or convergence within the reasonable data center packet loss rate range (less than 1\%~\cite{LossRadar, Pingmesh}) (see Section~\secref{sec:simulation}). This result aligns with the observations in~\cite{wang2020domain}. 
%To further support this, we conducted simulations at loss rates $1\%$ and $0.1\%$ .

To handle packet losses from workers to the PS, we can perform partial aggregation: the PS broadcasts partial aggregation results once it hears from the majority (e.g., 90\%) of workers. 
%Partial aggregation also has the potential of hiding some packet losses on the path from workers to the switch, as partial results might reach the worker experiencing packet loss before a timeout event is triggered. 
We evaluate the impact of stragglers and partial aggregation in Section~\secref{sec:simulation} and show that THC with partial aggregation over $90\%$ of workers reach the baseline accuracy.

\section{Implementation}
{\bf \sysname Worker}.
%We integrated two modules into the BytePS system: the {\it compression module}  and the {\it communication module}. 
\crupdate{We develop \sysname prototype of workers atop BytePS's PyTorch extension \cite{BytePS}.} During each iteration, the BytePS worker receives the calculated gradient from the front-end PyTorch and passes it to our {\em compression module}. 
Our compression module runs the \sysname algorithm on GPUs and employs a GPU-friendly implementation of \crupdate{RHT}. It also keeps the error-feedback records to compensate for the biased quantization as mentioned in Section~\ref{sec:choiceofparams}. 
%In our \sysname system prototype, we developed this module atop PyTorch and integrate it directly into the BytePS PyTorch extension.

The communication module is a C++ module developed based on Data Plane Development Kit (DPDK), which provides kernel bypassing so that applications can directly receive data from the NIC using busy polling. The communication module assembles packets based on compression results and communicates with the PS.
% Packets sending and receiving are conducted over the Data Plane Development Kit (DPDK), which is a simple framework for high-performance packet processing in data plane applications. 
% It enables customizable packet assembling and provides kernel bypassing so that applications can directly receive data from the NIC using busy polling. 

% RDMA is an alternative approach that also enables kernel bypassing. 
Another approach is to adopt the RDMA protocol, such as RoCEv2~\cite{rocev2}. However, adopting the RDMA protocol requires additional header parsing functions on the switch side. SwitchML~\cite{SwitchML} has demonstrated that the RoCEv2 protocol can be used with in-network aggregation by carefully parsing the header. For \sysname, we consider this as future work.

%\lam{a bit difficult to understand but that should be fine}

% Our system instruments compression code into BytePS. During each iteration, each BytePS worker first finishes back-propagation on its dataset partition and generates local gradients. The worker then compresses local gradients and packs them into packets and determines the aggregator slot to use for each packet.\minlan{What's an aggregator slot?} It then writes the aggregator slot index and slot round number as packet header fields. After receiving a response packet from the PS, the worker increments the corresponding aggregator slot's round number by one as mentioned previously. The worker finally performs decompression and updates the model with the global update. \minlan{key point is that the worker does additional thing for THC but it's lightweight. It's not very obvious in the above writing}

%\lam{do you think we should move RDMA to discussion? I want to keep this section short and simple}
% However, parsing the RDMA headers ~\lam{RoCE} and supporting RDMA~\lam{RoCE}  congestion control requires non-trivial additional work on the programmable switch side. Therefore, we go with DPDK in this project. \minlan{Can we just say we will work on RDMA in the future and we will get even better perf?}

{\bf \sysname Parameter Servers.}
We implement two versions of parameter servers: the software version written in C++ and the programmable switch version  implemented on the Intel Tofino switch~\cite{tofino}.
% \lam{any fancy thing that used in  switch, recirculation etc}
In the programmable switch version, the PS performs table lookup using  the ``Table'' control block. After the table lookup, the switch sends packets carrying table values through recirculation ports. The "Register" extern then takes care of value aggregation. Please see Appendix~\ref{appendix:switch_resources} for the resource usage of the programmable switch PS.
%we also provide a software PS implementation written in C++. This software PS follows the exact logic of the programmable switch PS. The only difference is that the software PS uses integer arrays as aggregator slots. Figure~\ref{fig:ps_logic} shows the overall workflow on the PS side.
% We also provide a software PS implementation written in C++. This software PS follows the exact logic of the programmable switch PS. %(Pseudocode~\ref{code:ps_logic}).
%The only difference is that the software PS aggregates the gradients in software slots. 

%-------------------------------------------------------------------------------

%-------------------------------------------------------------------------------
\vspace{-1mm}
\section{Evaluation}
\label{sec:eval}
% \minghao{
% New Evaluation Plan:
% \newline Lab testbed: 4 workers with Co-located PS; 4 workers with Tofino2.
% \newline AWS: 8 workers (p3x16 instances) each with 8 GPUs and uses Co-located PS.
% \newline Tasks (throughput): ResNet50, ResNet101, ResNet152, VGG16, VGG19, RoBERTa-base, RoBERTa-large, Bart-large, GPT-2
% \newline Tasks (TTA): VGG16,  RoBERTa-base, GPT-2
% \newline baseline (Co-located PS): 
% \begin{enumerate}
%     \item AWS: BytePS (TCP), Horovod (TCP)
%     \item testbed: Horovod (RDMA), TopK, Terngrad
% \end{enumerate}
% }
\begin{figure*}[htb]
         \centering
         \includegraphics[clip, trim=4.6cm 0.5cm 4.7cm 0.75cm, width=\textwidth]{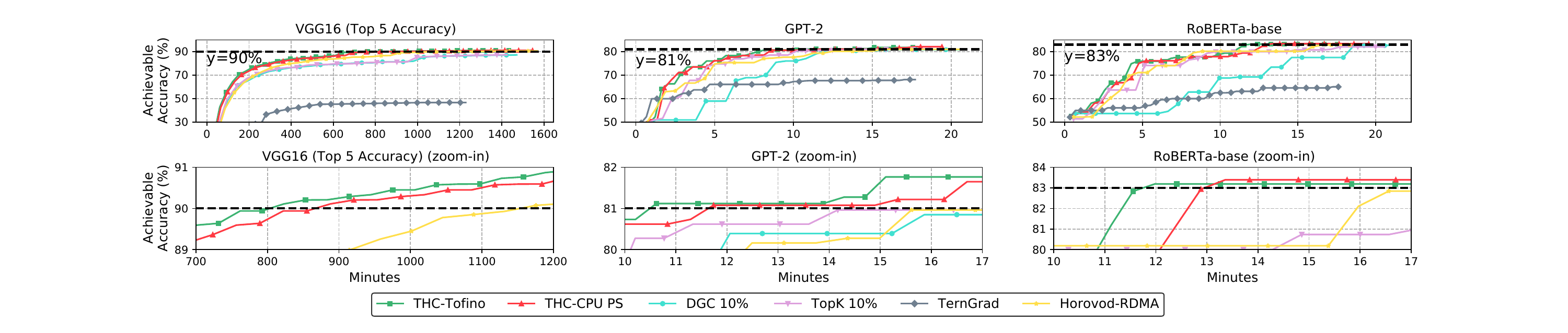}
         \vspace*{-5mm}
         \caption{\crupdate{Time to accuracy (TTA) comparison over one image processing task (i.e., VGG16) and two NLP tasks (i.e., GPT-2 and RoBERTa-base). The second row zooms in on the competitive results.}}
         \vspace*{-4mm}
         \label{fig:3tta}
    \label{fig:tta}
\end{figure*}
We evaluate our THC prototype by training popular computer vision and language models on a local four-worker testbed and AWS EC2. Experiments show that employing THC gives shorter time-to-accuracy (\crupdate{ $1.40\times$ to $1.47\times$} speedup) and higher throughput (\crupdate{$25\%$ to $54\%$} increase) on our local testbed with a programmable switch, 
compared to state-of-the-art training frameworks. THC also outperforms \crupdate{DGC-10\%}, TopK-10\%, and TernGrad in time-to-accuracy as it has little impact on accuracy and eliminates the overhead at the PS.

% \Comment{check if below is right at "large"}
%on a testbed with four workers and one PS and by running multi-client simulations.
\textbf{Testbed Setup.} Our local testbed has four GPU machines \crupdate{as workers}, each with one NVIDIA A100 GPU
% , two AMD EPYC 7313 16 cores CPU at 3.0 GHz, and 512 GB memory. 
and one NVIDIA MCX516A-CCAT ConnectX-5 100G dual-port NIC connecting to a Tofino2 switch.
%For simulation setup, please see Section~\secref{sec:simulation}.
For large-scale experiments, we use eight AWS EC2 p3.16xlarge instances, each with eight NVIDIA V100 GPUs and 25 Gbps network bandwidth. 

% \minghao{should also clearly state the exact configs for each figure (e.g., batch size, etc.)}

% \mitz{Don't think you need a parenthetical describing BytePS below, described back in Section 7.}\minghao{right right}
\textbf{Systems for Comparison.} 
% \mitz{Above:  always nice to say a word or two about why this configurations, if there's room}
% \Comment{I would move all Espresso discussion to a footnote, MM}
%On our local testbed, 
We run three versions of THC: (1) with a programmable switch as the PS (labeled as {\em THC-Tofino}), (2) with the software PS process running on a single stand-alone CPU machine \crupdate{(connected to the Tofino2 with a ConnectX-5 100G dual-port NIC)} ({\em THC-CPU PS}), and (3) with colocated PSes for each worker ({\em THC-colocated}), which \crupdate{build on BytePS's PS architecture and uses BytePS's RDMA module for a fair comparison.} We compare THC with two state-of-the-art systems: (1) Horovod which uses RDMA for communication ({\em Horovod-RDMA}) and (2) BytePS with colocated PSes for each worker ({\em BytePS}). 
We also compare THC against \crupdate{three} popular compression algorithms:
\crupdate{DGC\cite{DGC} and TopK\cite{TopK} are sparsification algorithms that only communicate the top $k$\% of coordinates by magnitude (here we set $k=10$ and refer to them as {\em DGC 10\%} and {\em TopK 10\%})}; and TernGrad~\cite{TernGrad}, a quantization algorithm that converts each coordinate into a value $x\in \{ -1, 0,1\}$ (we refer to it as {\em TernGrad}). TernGrad represents a stream of quantization algorithms~\cite{bernstein2018signsgd, QSGD} with small differences in design. \crupdate{{\em BytePS}, {\em DGC 10\%}, {\em TopK 10\%}, and {\em TernGrad} all use BytePS's colocated PSes and RDMA module.}
We also tried Espresso\cite{ByteComp} but faced convergence issues.\footnote{We followed the instructions in Espresso's git repository and installed all desired versions of software. Unfortunately, we couldn't get models to converge (training loss became "nan" within three iterations and the accuracy stalled around 0.1\%). We contacted the authors but they do not have time to fix it. See \url{https://github.com/zhuangwang93/Espresso/issues/3}.}
%signSGD (which converts to a sign \{-1,,1\}) and QSGD. We run both TopK 10\% and TernGrad with our Software PS and DPDK communication backend. 

% and didn't receive help from the authors. What we encountered also matches an open issue (\url{https://github.com/zhuangwang93/Espresso/issues/3}) unanswered at the time of submission.}
% \minghao{TODO: give an extensive description of the settings we are comparing against, e.g., make it clear that THC has switch and sw variants }

On AWS EC2, we deploy THC with software PS built on top of BytePS \cite{BytePS} servers. We compare THC against BytePS with colocated PS ({\em BytePS}) and Horovod \cite{horovod}. All systems use the TCP protocol to communicate. Unless noted otherwise, we use the following THC configuration: granularity 30, $p$-fraction 1/32, and 16 quantization levels. This configuration avoids overflow for up to eight workers, saturates the worker to PS bandwidth of our system prototype, and consistently achieves high accuracy across various models. 

\textbf{Workloads.} We evaluate THC with network-intensive \crupdate{\cite{ATP, SwitchML}} computer vision models (VGG16 and VGG19 \cite{vgg}) and language models (RoBERTa-base, RoBERTa-large \cite{roberta}, Bart-large\cite{bart}, BERT-base\cite{bert}, and OpenAI GPT-2 \cite{GPT2}). We train the vision models with the ImageNet1K dataset \cite{imagenet} and train the language models with the GLUE (General Language Understanding Evaluation benchmark) SST2 (The Stanford Sentiment Treebank) task \cite{sst2}. Unless noted otherwise, we set the per-GPU batch size as 32. We include results for computation-intensive models that don't benefit much from accelerated inter-machine communication in Appendix \ref{appendix:c}.

{\bf Metrics.} We first measure time-to-accuracy (TTA) as the training time needed to reach a target validation accuracy. We set the target accuracy based on the convergence of our no-compression baseline.
%, of training ResNet50, VGG16, and RoBERTa-base with different frameworks and compression algorithms (THC, TopK 10\%, and TernGrad). 
We then present the training throughput (images per second or tokens per second referred to as {\em samples per second}) for all training tasks. We also show the breakdown of computation and communication time to highlight the factors that contribute to THC's improvements. 

% \newline
% \begin{enumerate}
%     \item full rotation \minghao{my understanding is that there's no huge accuracy difference between full and partial rotation. probably just stick with partial rotation and discuss full rotation in section 5.4?}
%     \item partial rotation: 10 should be good
%     \item error feedback: likely worthwhile to always enable ef 
%     \item bit budget: 8 should be good for ~20 workers; can definitely choose a higher number if we are having fewer workers.
%     \item norm-normalization
% \end{enumerate}
% \begin{enumerate}
%     \item full rotation
%     \item partial rotation: 10 should be good
%     \item error feedback: likely worthwhile to always enable ef up to have ~20 workers. With many workers, might not be necessary to enable. the cost is really just the inverse rotation
%     \item quantization levels: 8 should be good for ~20 workers; can definitely choose a higher number if we are having fewer workers. This is the main thing and affects whether we need rotation
%     \item norm-normalization: this need further results on the norm communication time vs. rotation time
%     \item just ef and no rotation and check the time-to-accuracy
% \end{enumerate}

\vspace*{-3mm}
\subsection{End-to-End Training Performance}
\vspace*{-1mm}
% \minghao{would be nice to quote SwitchML numbers}
% To analyze THC's effectiveness in reducing the overall training time of distributed training jobs, we record the TTA when training ResNet50 and VGG16 with ImageNet1K and when training RoBERTa-base with SST2 at 100Gbps. 
%  shows the top5 validation accuracy for training VGG16, ResNet50, and RoBERTa-base. 

% - VGG: 
% better than Horovod. RDMA is faster than dpdk but we still gain

% THC software vs switch:

% compare with all the others...

% - different models... generally good. 
% model difference

% - different compression

{\bf Time-to-accuracy.} 
\label{sec:tta}
%Due to space constraints, we choose three representative models, VGG16, GPT-2, and RoBERTa-base, to report the TTA results.
Figure~\ref{fig:tta} shows that for GPT-2, THC-Tofino reaches the 81\% target accuracy $1.47\times$ faster than the Horovod-RDMA baseline, and THC-Software PS reaches the target accuracy $1.33\times$ faster. 
For RoBERTa-base, THC-Tofino achieves the 83\% target accuracy $1.43\times$ faster than the Horovod-RDMA baseline; and THC-Tofino also achieves a $1.40\times$ speedup to reach the $90\%$ target accuracy for VGG16.
Note that even though our system prototype uses DPDK, which has similiar performance with RDMA, we still notably outperform Horovod-RDMA.
THC-software PS reduces network communication with minimal impact on model convergence, thanks to the optimizations explained in Section~\ref{sec:choiceofparams}.
Using the programmable switch (THC-Tofino) further accelerates the training by reducing the volume of transmitted data through in-network aggregation.

\begin{figure}[t!]
    \centering
    \includegraphics[clip, trim=0.35cm 0.5cm 0.38cm 0.4cm, width=0.49\textwidth]{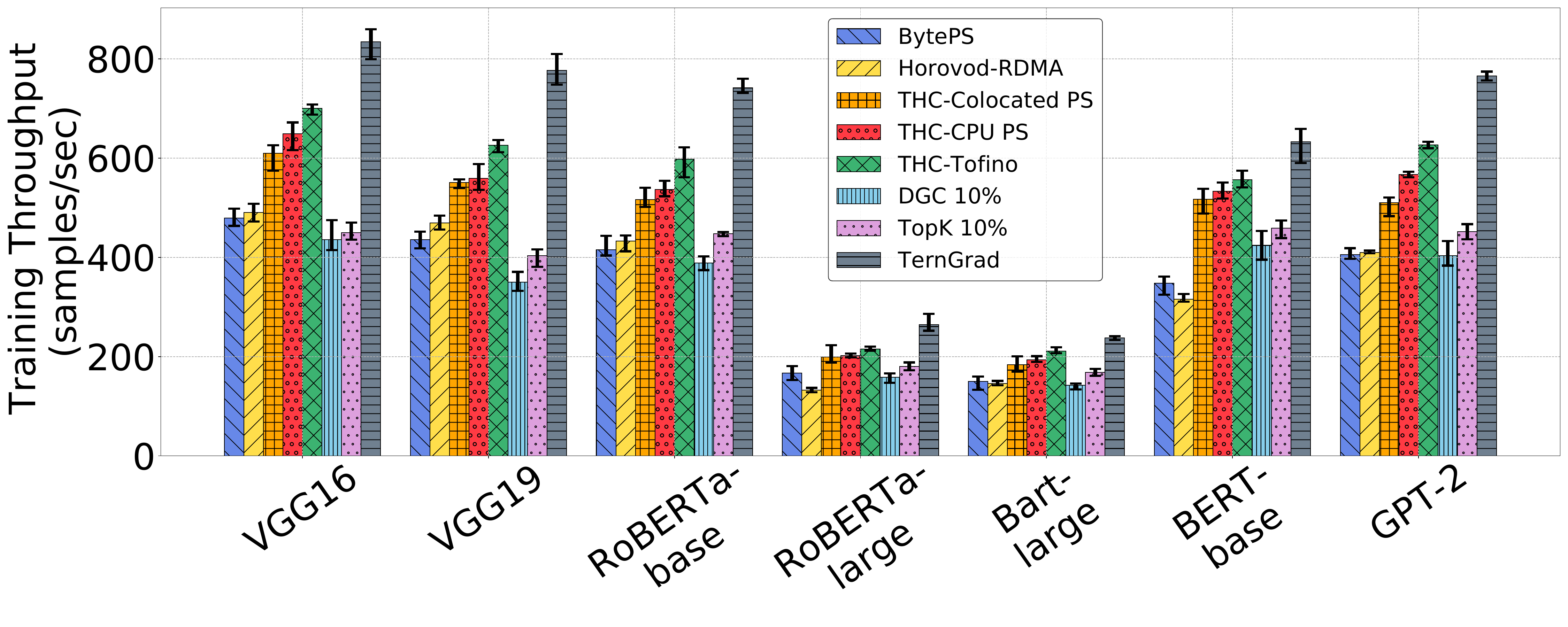} 
    \vspace*{-6mm}
      \caption{\crupdate{Training throughput with 100Gbps links over different network-intensive architectures. }}
      \label{fig:training_throughput}
\end{figure}

% \minghao{I am here. fix the section below}
% \Comment{Reiterate the target accuracy, and say where TopK stalls.}
Although training with \crupdate{DGC 10\% and TopK 10\% also} approach the target accuracy, they suffer from the high PS compression overhead that leads to longer training epoch time and TTA.
TernGrad stalls at low accuracy for all three models despite its high training throughput (Figure~\ref{fig:training_throughput}). This is because TernGrad loses information during its compression and thus cannot improve end-to-end TTA with more training epochs.

{\bf Training throughput.}
To examine the synchronization round time reduction we achieve with THC, we measure the throughput in Figure~\ref{fig:training_throughput}.
THC-Tofino provides higher throughput than all alternatives (except TernGrad). For example, THC-Tofino achieves 54\% improvement over Horovod-RDMA for GPT-2. \crupdate{THC-colocated has 11\% to 37\%} higher throughput than TopK because THC eliminates the PS-side compression operations.
TernGrad provides the highest throughput because it uses fewer bits per coordinate and has shorter PS time and compression overhead. However, TernGrad does not improve TTA as Figure~\ref{fig:tta} shows due to its low accuracy (Section~\ref{sec:network-compute-tradeoff}).

\begin{figure}
%     \centering
%     \begin{minipage}[b]{0.50\textwidth}
%       \centering 
%       \includegraphics[width=\textwidth]{figures/eval_example/throughput_all_with_compression.pdf} 
%       \caption{\small Training throughput
% (with 100G bandwidth between each worker and the PS). }
%       \label{fig:training_throughput} 
%     \end{minipage}\hfill
    \begin{minipage}[b]{0.235\textwidth}
      \centering 
      \includegraphics[width=\textwidth]{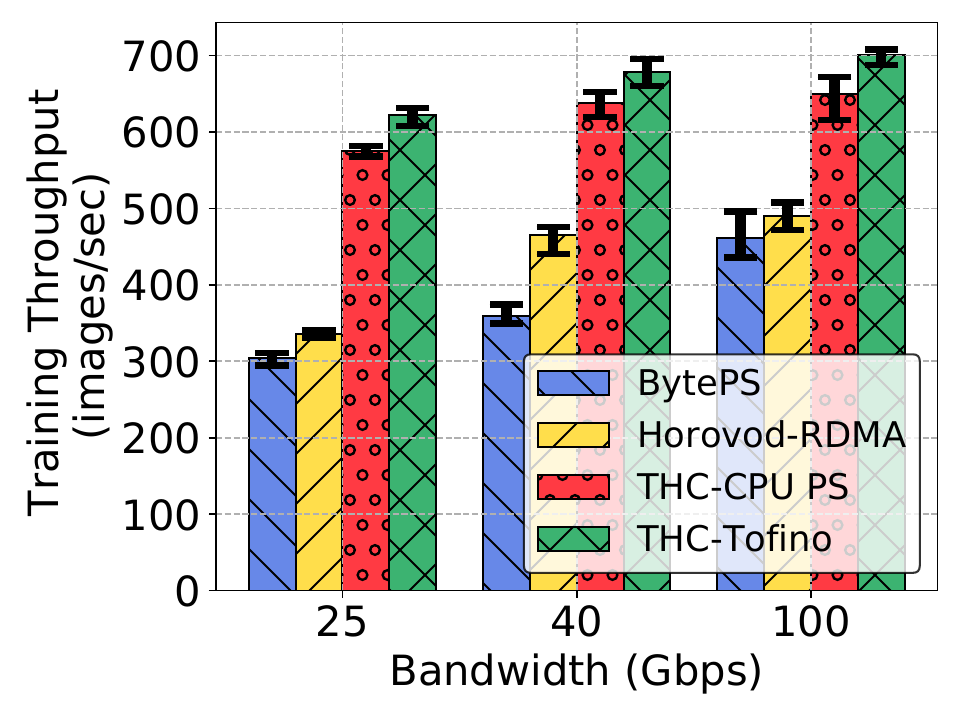} 
      %\vspace*{-2mm}
      \caption{\small Training throughput for different bandwidths.}
      \label{fig:bandwidth_diff_throughput} 
    \end{minipage}
    \begin{minipage}[b]{0.235\textwidth}
      \centering 
      \includegraphics[width=\textwidth]{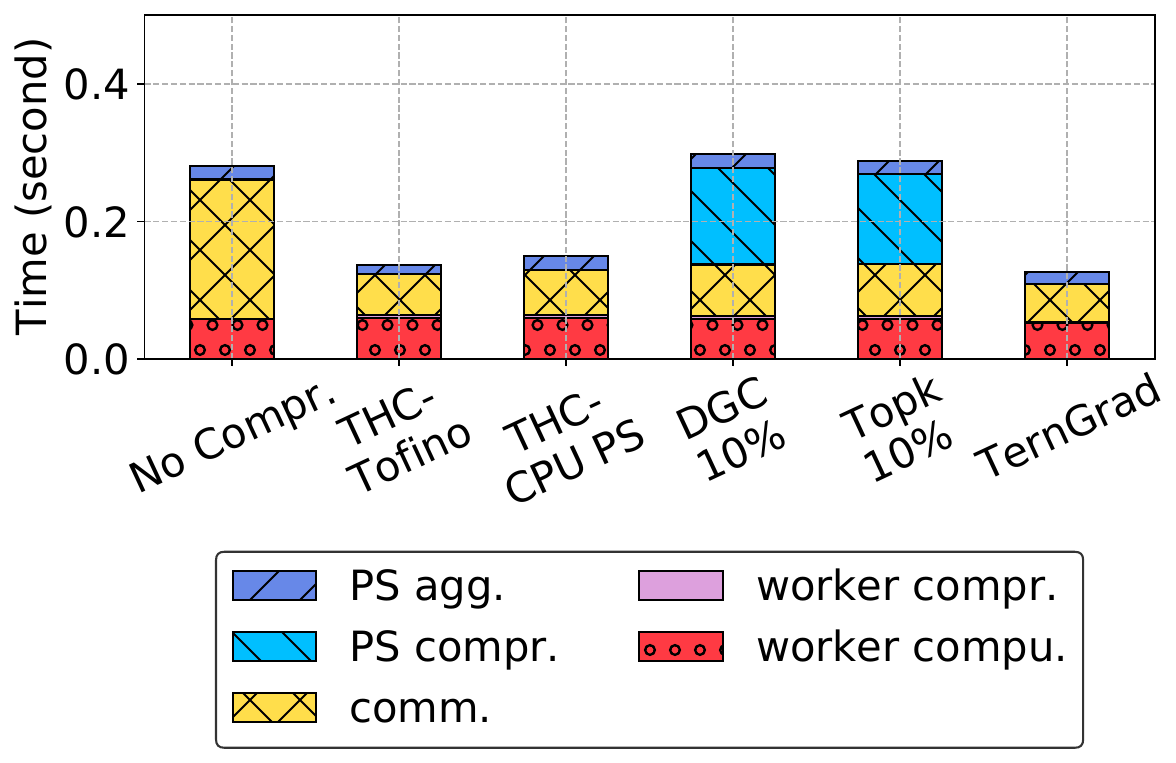} 
      %\vspace*{-2mm}
      \caption{\small \crupdate{ Average training round time breakdown.}}
      \label{fig:vgg16_breakdown} 
    \end{minipage}
        % \vspace{-5mm}
\end{figure}
{\bf Effectiveness with different bandwidth.} 
We train the VGG16 architecture under different network bandwidth settings (25, 40, and 100Gbps) in Figure~\ref{fig:bandwidth_diff_throughput}. THC-Tofino achieves 
$1.85\times$, $1.45\times$, and $1.43\times$ training throughput over Horovod-RDMA at 25Gbps, 40Gbps, and 100Gbps respectively.
When the bandwidth decreases from 40Gbps to 25Gbps, the throughput of Horovod-RDMA drops significantly because it faces more network bottlenecks. Meanwhile, the performance of THC-Tofino and THC-CPU PS downgrades gracefully as the bandwidth decreases, leading to higher training speedups at low bandwidths. We expect THC to offer more benefits under slower networks (e.g., 10Gbps, 1Gbps) which we might encounter in federated learning settings.

%Compression algorithms usually offer the most benefit in low bandwidth settings. To compare the speedups THC achieves with different bandwidths,

% 

% \textbf{Figure 3: vary the bandwidth}
% Goal: Vary the bandwidth (25, 40, 100 Gbps) to show our solution might gain more when bandwidth is limited

% Metric: throughput

% x: bandwidth: 25, 40, 100G

% y: throughput
% curves: 
% \begin{enumerate}
%     \item Tofino DPDK with on-GPU THC
%     \item SoftwarePS DPDK with on-GPU THC
%     \item N-to-N TCP w/o compression
%     \item RDMA w/o compression

% \end{enumerate}

\subsection{Breakdown of Network and Compute Time}
\label{sec:network-compute-tradeoff}
In Figure~\ref{fig:vgg16_breakdown}, we break down the time per iteration into the time spent at the PS, workers, and the communication, when training VGG16 at 100Gbps with THC-Tofino, THC-CPU PS, TopK 10\%, and TernGrad. At the PS, we measure the compression time and aggregation time. At the workers, we measure the training time and compression time.

% - THC software: Compared no compression, we reduce communication time by XX, and PS time by XX. worker increase by XX. distributed workers, key bottlenck at PS/bw...operations are on GPU
% - THC switch version has further tradeoff: .... 
% - Compare with THC software with TopK . TopK is also on GPU at the worker but on CPU at the server. therefore, although we have similar comm time, the PS time difference is huge

% - Compare with TernGrad: TernGrad on GPU and has limited operations at the PS. it gives both shorter comm and on PS time. but doesn't improve TTA

% \minlan{Start from positive point}
\crupdate{THC-CPU PS reduces the gradient communication time (comm.) to $32.5\%$ of that of the no-compression baseline}. As a tradeoff, we introduce compress and decompress operations on the GPU on the worker side. However, these operations only increase the overall worker time by 9.5\%. The result demonstrates that in distributed training settings where workers periodically synchronize a large amount of data, saving bandwidth by slightly increasing the worker GPU computation time is worthwhile. THC-Tofino achieves more savings by further reducing communication time through in-network aggregation and offloading the PS to the switches. 

For \crupdate{DGC 10\% and} TopK 10\%, they have to run expensive sorting operations on the PS \crupdate{(DGC 10\% additionally requires local gradient accumulation)}, introducing a significant overhead at the PS side. Therefore, although TopK10\% gives similar communication time as THC-CPU PS, the overall round time is $46.5\%$ higher than that of THC-CPU PS.

TernGrad uses only two bits per coordinate and requires simple summation at the PS. Thus, it has a short communication and PS time. However, TernGrad produces high NMSE and correspondingly can produce poor TTA results.

\vspace{-3mm}
\subsection{THC Performance on AWS EC2}
\begin{figure}[t]
    \centering
    \includegraphics[clip, trim=0.35cm 0.5cm 0.38cm 0.4cm, width=0.35\textwidth]{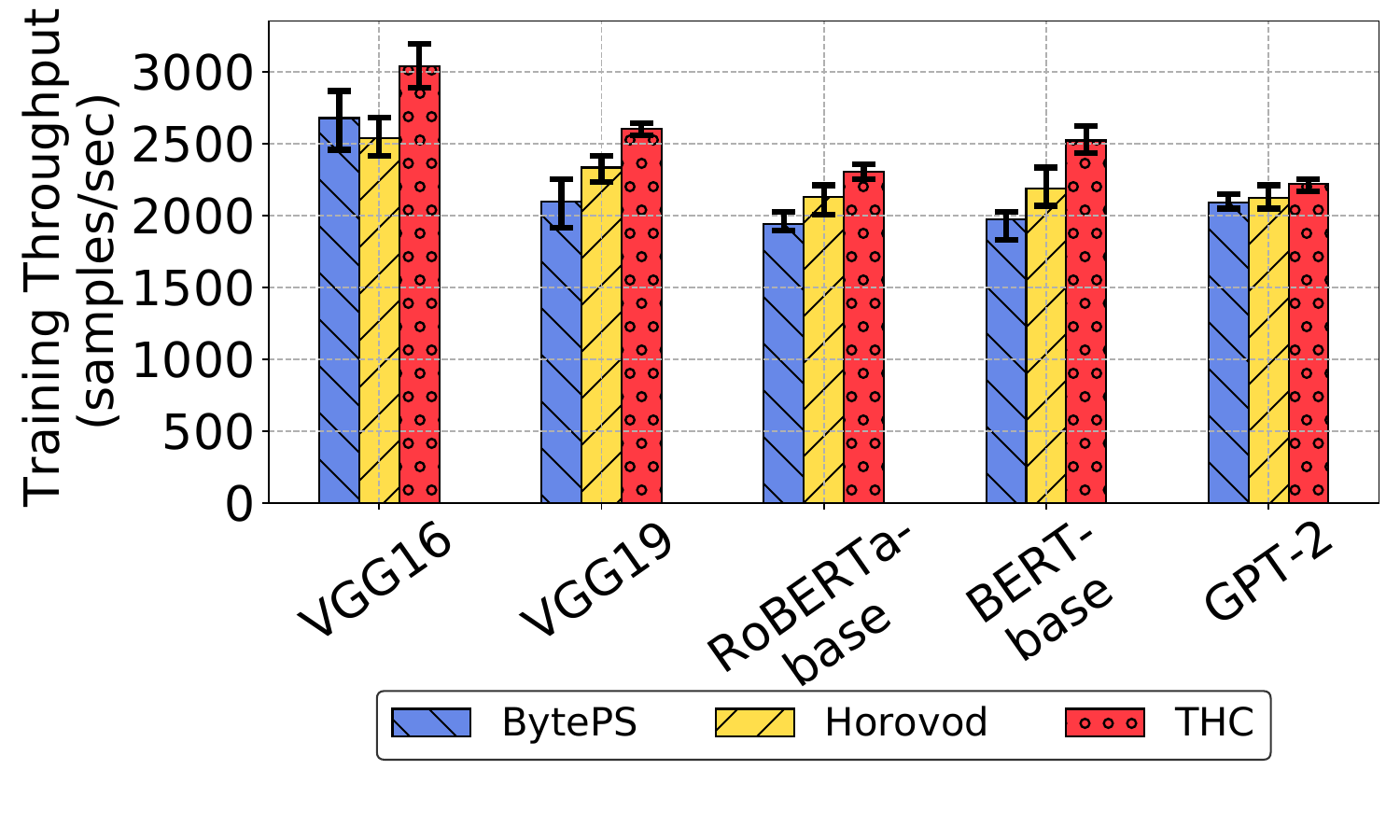} 
    \vspace*{-6mm}
      \caption{Training throughput across eight AWS EC2 p3.16xlarge instances.}
      \label{fig:training_throughput_AWS}
\end{figure}

\vspace{-0mm}
We measured throughput on AWS EC2 instances equipped with workers containing multiple GPUs at a larger scale (Figure~\ref{fig:training_throughput_AWS}). Since AWS instances have 8 GPUs per worker, we have a higher intra-machine communication overhead compared to our local testbed setting. This means that inter-machine communication overhead, which THC optimizes for, takes a smaller portion of training time. Despite this, THC consistently outperformed all the baseline models, resulting in throughput improvements of \crupdate{1.05$\times$ to 1.16$\times$}.\footnote{Note that Bart-large and RoBERTa-large are not displayed in Figure~\ref{fig:training_throughput_AWS} due to the V100 GPU's memory limitations on the EC2 environment. We used a smaller batch size and reported it in Appendix \ref{appendix:c} separately.}

% \minghao{TODO: explain that we move large models result with smaller batch size to appendix}

\vspace{1mm}
\subsection{THC Simulation Results}
\label{sec:simulation}
\vspace{0mm}
We run simulations to understand THC under different system configurations. \crupdate{As the number of workers increase, the performance of THC increases due to the UHC property. We show that the error of THC scales well from 4 to 64 workers, in contrast to biased compression algorithms like TopK whose error can increase by $9.9\times$ over the same margin}. Furthermore, under the synchronization and partial aggregation schemes, THC shows less than a $0.5\%$ drop in training accuracy in the presence of packet loss and straggling workers. 

{\bf Simulation Environment.} We simulate THC on an academic cluster with 4 A100/V100 GPUs per node. \crupdate{Multiple worker training is modeled by storing multiple passes of the backpropagation before performing an update step. This allows us to compress and decompress the aggregated gradient with THC's algorithm (and others) before updating the model, reproducing the communication steps in actual systems.} 

For the scalability experiments, we train BERT and RoBERTa \cite{roberta} on SST2 \cite{sst2} with batch size 8. The configuration uses granularity $36$, $p$-fraction 1/32, and bit budget 4. We choose language tasks for the scalability results because they are more sensitive to small compression errors in the gradient. The other simulations train ResNet50 \cite{resnet} models on the CIFAR100~\cite{CIFAR100} dataset with a batch size 128, workers 10, granularity 20, $p$-fraction 1/512, and bit budget 4. 

\begin{figure}[t]
    \centering
    \begin{minipage}[t]{0.49\linewidth}
      \centering 
        \includegraphics[width=\textwidth]{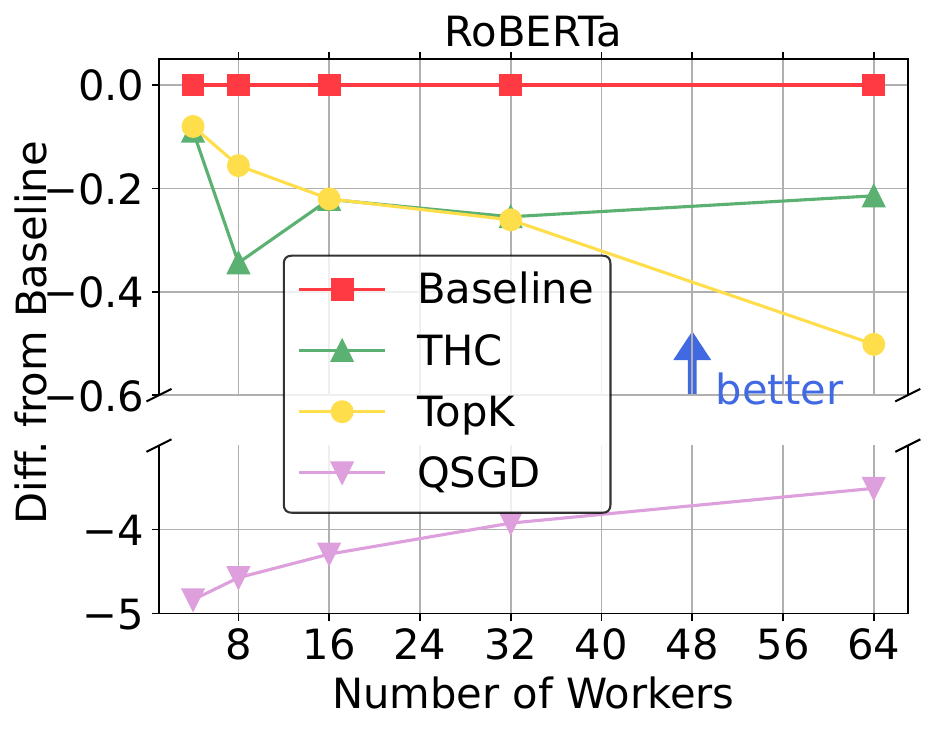}
    \end{minipage}
    \begin{minipage}[t]{0.49\linewidth}
      \centering 
      \includegraphics[width=\linewidth]{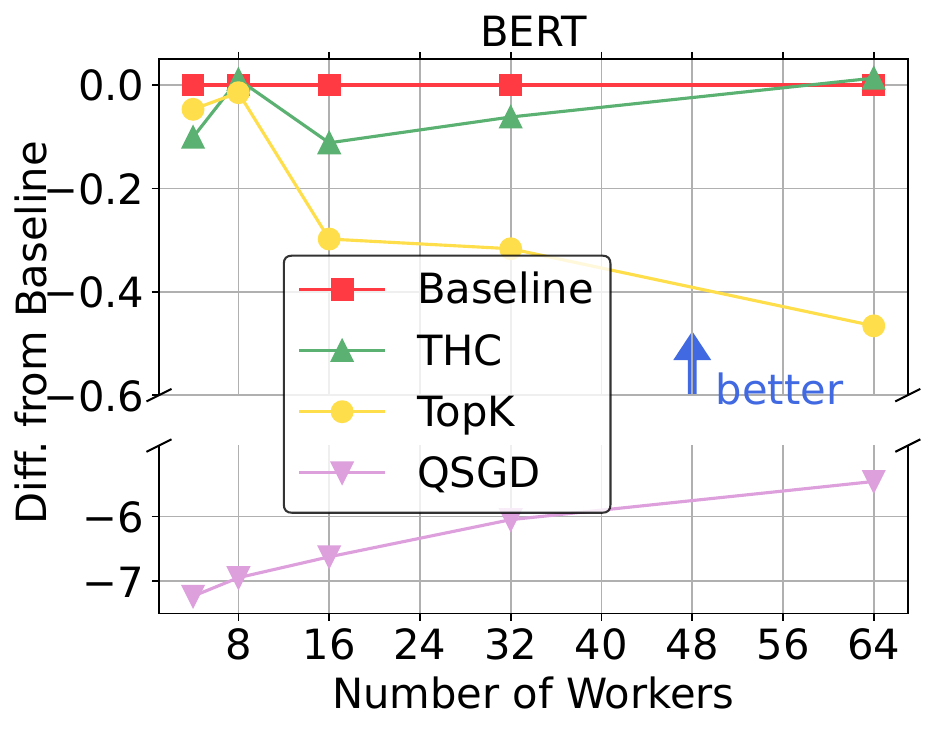}    
    \end{minipage}
    \vspace*{-1mm}
    \caption{{\small Scalability of THC. \crupdate{Displaying the difference in training accuracy from baseline after two epochs (i.e. if the final training accuracy for 4 worker RoBERTa baseline is 95\%, then the accuracy for 4 worker RoBERTa QSGD would be 90\%).}} }
    \vspace*{-0mm}
    % \kevin{choose one overflow frequency, mention why we beat the baseline} \lam{hi kevin, since we have multiple baseline, can we make sure all lines use different color?}
    % \kevin{yeah, i'll regenerate}
    \label{fig:scalability}  
\end{figure}

{\bf Scalability.} We fine-tune a model for two epochs for each compression algorithm with 4, 8, 16, 32, and 64 workers and then compute the percentage difference from the uncompressed baseline accuracy. We track the difference in accuracy rather than the absolute accuracy because machine learning effects can alter the baseline accuracy as the batch size (number of workers) changes. We compare THC with bit budget 4 against baseline (no compression), TopK \cite{TopK} and Quantized Stochastic Gradient Descent (QSGD) \cite{QSGD}. QSGD is chosen because we want to compare compression algorithms that have the same compression ratio: QSGD is analogous to an unbiased version of TernGrad/SignSGD with a tunable compression ratio. 
%Note that since we utilize the version of THC that sends 4 bits upstream (to the PS) and 8 bits downstream (from the PS) per coordinate, 
We choose the k-value and number of intervals of TopK and QSGD respectively to match the overall compression ratio of THC with bit budget 4.

\crupdate{Figure \ref{fig:scalability} shows that for BERT, the accuracy of THC actually improves as the number of workers increases, with the error decreasing from $-0.1\%$ to $0$ (no difference from baseline) from 4 to 64 workers. Although there seems to be some outlier at 8 workers that is likely due to randomness in the training process, these trends match our predictions in \cref{THC} and can be attributed to the increased accuracy of the estimate of the average gradient. In comparison, the error of TopK quickly inflates by $9.9\times$ from $-0.047\%$ to $-0.46\%$ in the same region because bias in the compression dominates and causes larger compression errors. The data for RoBERTa show a similar trend, with THC becoming the most accurate at 32 workers and beyond.}

\crupdate{Such results are promising for THC because actual system performance depends on both throughput and compression accuracy.} As we showed in~\secref{sec:tta}, THC has a higher throughput in training than most compression algorithms. \crupdate{At 16 workers and beyond, THC also shows the highest accuracy, implying that THC will have better time-to-accuracy results for a system with many workers. The advantage over other compression algorithms is more apparent in larger scale systems because biased algorithms such as TopK lose accuracy at~scale. }
%while THC will only become better as the system scales.

However, we cannot increase the number of workers to an arbitrary large size without incurring costs to accuracy. From each worker, the largest value per coordinate aggregated at the switch is equal to the granularity, so the maximum aggregated result is $m=g \times (\#\ of\ workers)$ and the number of bits needed to send this value downstream is $\ceil{\log_2 m}$. If we keep the number of bits sent downstream constant, we must decrease the granularity for a larger number of workers to prevent overflow, which increases the error. This can be seen from our results in Figure \ref{fig:nmse} in Appendix \ref{appendix:c}. One advantage however is that as the granularity decreases, we can also decrease the bit budget for THC, sending fewer bits per coordinate upstream. On the other hand, if we keep the granularity constant, then we must send more bits per coordinate downstream, decreasing the throughput. In these experiments, we kept the granularity constant and instead adjusted the compression ratio of the downstream data (including for TopK and QSGD for fair comparison). It is likely that the optimal strategy is to employ a combination of the options depending on the specifics of the system.

% However, the effect of increasing the total batch size has inherent diminishing returns, since using larger total batch sizes leads to worse generalizability in final test accuracy.

\begin{figure}[t]
    \centering
    %\vspace*{-2mm}
    \begin{minipage}[t]{0.49\linewidth}
      \centering 
      \includegraphics[width=\linewidth]{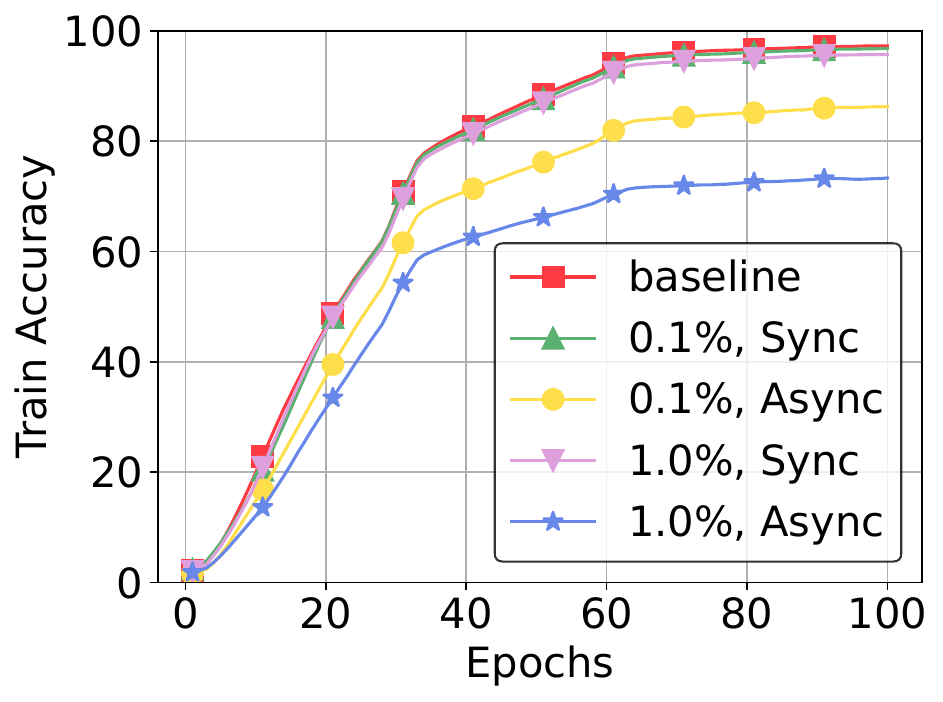} 
    \end{minipage}
    \begin{minipage}[t]{0.49\linewidth}
      \centering 
      \includegraphics[width=\linewidth]{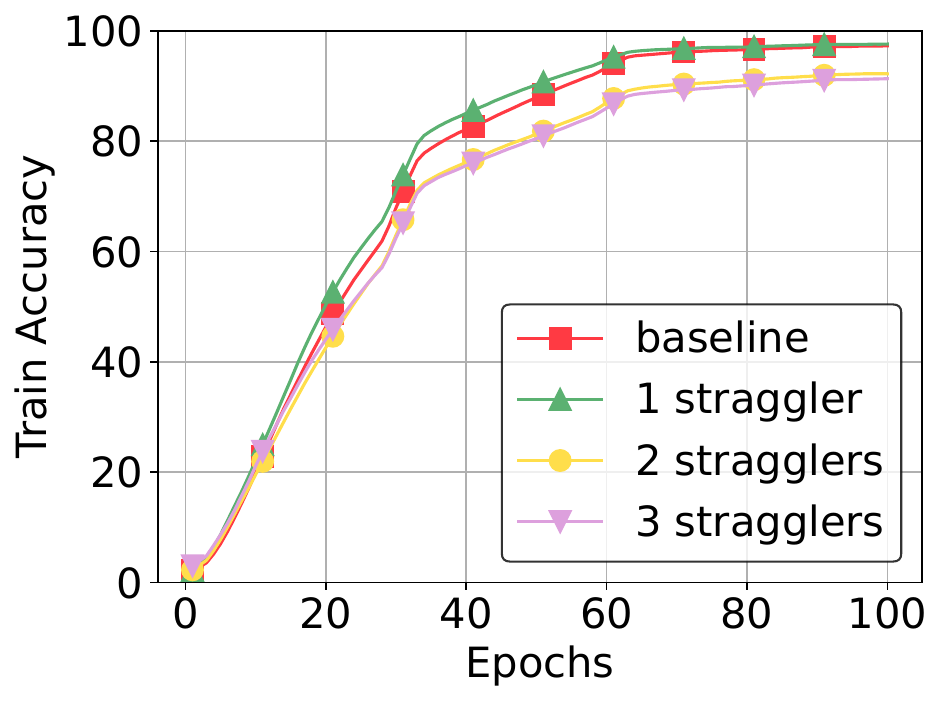}    
    \end{minipage}
    \vspace*{-2mm}    
    \caption{Resiliency to Gradient Losses. Displaying attainable accuracy with packet drops and stragglers.}%{\normalsize left: packet loss rates with synch/no synch, right: stragglers}
    \vspace*{-1mm}
    %\vspace*{-2mm}
    \label{fig:resiliency} 
\end{figure}

% \begin{figure}[h]
%       \centering 
%         % \includegraphics[width=\linewidth]{figures/simulation/packet_loss.pdf} 
%       \includegraphics[width=\linewidth]{figures/eval_example/new_kevin/packets.pdf}
%       \caption{Packet Loss Resiliency}%\ran{We cannot show such a figure, accuracy has to increase with \#workers.}\ran{Add a grid}}
%       \label{fig:packet} 
% \end{figure}

% \begin{figure}[h]
%       \centering 
%         % \includegraphics[width=\linewidth]{figures/simulation/stragglers.pdf} 
%       \includegraphics[width=\linewidth]{figures/eval_example/new_kevin/straggler.pdf}
%       \caption{Straggler Resiliency}%\ran{We cannot show such a figure, accuracy has to increase with \#workers.  Avoid having non-integer ticks. }\ran{Add a grid}}
%       \label{fig:straggler} 
% \end{figure}

{\bf Packet Loss.} 
%Though we mentioned that our network has a negligible loss rate in Section 6, we show that 
%THC can retain accuracy even in lossy networks 
%by artificially simulating dropped packets both upstream and downstream as 0s in the gradient. 
%To combat packet loss, we propose a synchronization scheme where workers coordinate their model parameters after every epoch by choosing to copy the parameters of one worker. 
We simulate packet losses in both directions between workers and the PS.
Figure \ref{fig:resiliency} shows that even with a packet loss rate of 1\%, which greatly decreases the final training accuracy with no synchronization, our proposed synchronization scheme reduces the training accuracy drop from 24\% to 1.5\%. With 0.1\% packet loss, synchronization reduces the accuracy discrepancy from 11\% to 0.5\%, which is nearly indistinguishable from baseline. The corresponding test accuracies
%to Figure \ref{fig:resiliency} 
are shown in Figure \ref{fig:resiliency_test} of Appendix \ref{appendix:c} and demonstrate~similar results. 

%Since we simulate packet losses in both directions, these results also show that packet losses moving from the PS to the worker are more problematic, since dropped packets from the PS can result in workers having slightly different models. Over time, the error can accumulate and make the aggregated gradient nonsensical, but this is solved by our synchronization scheme.
% which allows THC to be accurate even under high rates of loss.

{\bf Stragglers.}
To mitigate stragglers, we model a scheme where the PS only waits for the top n\% of the workers for aggregation. Figure \ref{fig:resiliency} shows the effect of randomly choosing a 1/2/3 stragglers during each round and dropping their gradients. With 10 workers, this corresponds to waiting for 90\%/80\%/70\% of the workers. Waiting for the top 90\% reaches the baseline accuracy, whereas 80\% and 70\% show only a 5-6\% decrease in final training accuracy. 

\section{Related Work and Discussions}
%\minghao{change this section into Discussion and add limitations of THC into this section?}
\vspace{2mm}
% \minghao{cite HiPress and follow--up works}

{\bf Systems Support for Gradient Compression.}
\crupdate{Gradient compression systems have been developed to train large models that are increasingly bottlenecked by communication, given the slower growth of bandwidth compared to GPU capability.} Previous systems (e.g., HiPress \cite{HiPress} and Espresso \cite{ByteComp}) focus on 
compression awareness and finding compression strategies and work division (e.g., compression on GPU or CPU). These systems maximize the overlap of efficient communication and compression to hide existing overhead. THC, on the other hand, mitigates compression overhead by reducing the number of compress/decompress operations. \sysname is complementary with these works.

%at the PS
 % We are complementary with previous works that focus on designing efficient communication and compression plan that considering existing overhead.
% It complement with previous works that maximize the overlap of efficient communication and compression to hide existing overhead.
{\bf In-network Aggregation for ML.}
SwitchML \cite{SwitchML} and ATP~\cite{ATP} have demonstrated the benefits of aggregating gradients within networks, but they do not support compression as the data is restricted to the format that can be directly aggregated. OmniReduce~\cite{omnireduce} and ASK~\cite{ask} propose using a key-value data structure for in-network aggregation. However, these approaches require compressing the gradient data into the key-value format and make assumptions about the data itself. In \sysname, we support aggregation directly on compressed data, making it orthogonal to previous works. This leads to efficient in-network aggregation with compression.

{\bf Supporting Other AllReduces.}
An important future research direction is incorporating homomorphic compression in other types of all-reduce like ring-based or tree-based. Currently, compression schemes fail to improve the performance of these types~\cite{MLSYS2022_773862fc}.
For example, in the widely-deployed ring all-reduce that requires $O(n^2)$ aggregation operations, existing schemes would need an excessive number of decompression and re-compression operations, leading to poor accuracy and slowdown compared to an uncompressed baseline.
THC makes the first step towards making compression algorithms ring-based or tree-based all-reduce friendly. For example, we may run the reduce operation directly on gradients compressed with Uniform THC using the same number of bits required for the PS aggregation (e.g., 8). However, this method is not compatible with our various optimizations, such as sending just $b$ (e.g., 4) bits or using the lookup table, and is thus sub-optimal. 

\crupdate{{\bf Colocated with Other Training Paradigms.}
While THC primarily focuses on data parallelism, it can seamlessly integrate with state-of-the-art training paradigms that use hybrid approaches combining tensor, pipeline, and data parallelism~\cite{megatron-lm, megatron-lm-in-scale}. Since all these optimizations are clearly separated in different dimensions, \sysname can be applied to the line of data parallelism without additional adaption. Moreover, Megatron-LM~\cite{megatron-lm-in-scale} reports that data parallelism remains the dominating factor in training throughput; thus, gradient exchanges will continue to make a significant contribution to computation and communication costs. We believe that increasing the degree of data parallelism is still a better choice, emphasizing \sysname's crucial role in training optimization.}

\crupdate{{\bf Compatibility with Security.}
An extensively studied application of homomorphism is Homomorphic Encryption (HE)\cite{Homomorphic_Encryption} in the security field. Although we consider HE as orthogonal to our work, it might be feasible to combine THC with other security practices. For example, applying differential privacy \cite{differential_privacy} techniques first and then compressing the tensors with THC can be practicable.
}

% \lam{move this to discussion/limitation, add one more point that in-network aggregation is more suitable to PS architecture, and add the limitation of number of workers in the discussion.}
% \kevin{I talk somewhat about the limitation of number of workers in the end of the scalability section. Feel free to delete/move that.}

% \minghao{
% {\bf Other Parallelisms.} Data Parallelism is often deployed with Model Parallelism and Pipeline Parallelism. Gradient compression still accelerates the gradient synchronization process in Data Parallelism. 
% }
%-------------------------------------------------------------------------------

%-------------------------------------------------------------------------------
\section{Conclusion}
\vspace{2mm}
\label{sec:conclusion_and_future}
THC is a novel framework that formally defines homomorphic compression. As homomorphic compression supports direct aggregation of compressed data, it also allows an elegant combination of gradient compression and in-network aggregation. To demonstrate THC's generalizability, we build a distributed DNN training system prototype that employs both the THC algorithm and in-network aggregation to accelerate gradient synchronization. Testbed experiments with four GPU workers, one programmable switch, and 100Gbps network show that our system prototype achieves up to 1.47$\times$ TTA improvement when we enable both gradient compression and in-network aggregation.

\section{Acknowledgment}
\crupdate{We thank the NSDI reviewers and our shepherd, Qun Huang, for their invaluable feedback. This work was supported in part by ACE, one of the seven centers in JUMP 2.0, a Semiconductor Research Corporation (SRC) program sponsored by DARPA. Ran Ben Basat was supported by the Meta Network for AI faculty award. Michael Mitzenmacher was supported in part by NSF grants CCF-2101140, CNS-2107078, and DMS-2023528.
We thank Vyas Sekar for proposing the term `Homomorphic Compression'.}
%-------------------------------------------------------------------------------
\bibliographystyle{plain}
\bibliography{references}

\begin{thebibliography}{10}

\bibitem{Homomorphic_Encryption}
Abbas Acar, Hidayet Aksu, A.~Selcuk Uluagac, and Mauro Conti.
\newblock A survey on homomorphic encryption schemes: Theory and
  implementation.
\newblock {\em ACM Comput. Surv.}, 51(4), jul 2018.

\bibitem{MLSYS2022_773862fc}
Saurabh Agarwal, Hongyi Wang, Shivaram Venkataraman, and Dimitris
  Papailiopoulos.
\newblock On the utility of gradient compression in distributed training
  systems.
\newblock In D.~Marculescu, Y.~Chi, and C.~Wu, editors, {\em Proceedings of
  Machine Learning and Systems}, volume~4, pages 652--672, 2022.

\bibitem{ailon2006approximate}
Nir Ailon and Bernard Chazelle.
\newblock Approximate nearest neighbors and the fast johnson-lindenstrauss
  transform.
\newblock In {\em Proceedings of the thirty-eighth annual ACM symposium on
  Theory of computing}, pages 557--563, 2006.

\bibitem{QSGD}
Dan Alistarh, Demjan Grubic, Jerry~Z. Li, Ryota Tomioka, and Milan Vojnovic.
\newblock Qsgd: Communication-efficient sgd via gradient quantization and
  encoding.
\newblock In {\em Proceedings of the 31st International Conference on Neural
  Information Processing Systems}, NIPS'17, page 1707–1718, Red Hook, NY,
  USA, 2017. Curran Associates Inc.

\bibitem{rocev2}
InfiniBand~Trade Association.
\newblock {InfiniBand Trade Association. RoCE v2 Specification. }.
\newblock \url{https://cw.infinibandta.org/document/dl/7781}, 2014.

\bibitem{HiPress}
Youhui Bai, Cheng Li, Quan Zhou, Jun Yi, Ping Gong, Feng Yan, Ruichuan Chen,
  and Yinlong Xu.
\newblock Gradient compression supercharged high-performance data parallel dnn
  training.
\newblock In {\em Proceedings of the ACM SIGOPS 28th Symposium on Operating
  Systems Principles}, SOSP '21, page 359–375, New York, NY, USA, 2021.
  Association for Computing Machinery.

\bibitem{DBLP:conf/icalp/BasatMV21}
Ran~Ben Basat, Michael Mitzenmacher, and Shay Vargaftik.
\newblock How to send a real number using a single bit (and some shared
  randomness).
\newblock In Nikhil Bansal, Emanuela Merelli, and James Worrell, editors, {\em
  48th International Colloquium on Automata, Languages, and Programming,
  {ICALP} 2021, July 12-16, 2021, Glasgow, Scotland (Virtual Conference)},
  volume 198 of {\em LIPIcs}, pages 25:1--25:20. Schloss Dagstuhl -
  Leibniz-Zentrum f{\"{u}}r Informatik, 2021.

\bibitem{basat2022quick}
Ran~Ben Basat, Shay Vargaftik, Amit Portnoy, Gil Einziger, Yaniv Ben-Itzhak,
  and Michael Mitzenmacher.
\newblock {QUIC-FL: Quick Unbiased Compression for Federated Learning}.
\newblock {\em arXiv preprint arXiv:2205.13341}, 2022.

\bibitem{ben2024optimal}
Ran Ben-Basat, Yaniv Ben-Itzhak, Michael Mitzenmacher, and Shay Vargaftik.
\newblock {Optimal and Near-Optimal Adaptive Vector Quantization}.
\newblock {\em arXiv preprint arXiv:2402.03158}, 2024.

\bibitem{bernstein2018signsgd}
Jeremy Bernstein, Yu-Xiang Wang, Kamyar Azizzadenesheli, and Animashree
  Anandkumar.
\newblock signsgd: Compressed optimisation for non-convex problems.
\newblock In {\em International Conference on Machine Learning}, pages
  560--569. PMLR, 2018.

\bibitem{gpt3}
Tom Brown, Benjamin Mann, Nick Ryder, Melanie Subbiah, Jared~D Kaplan, Prafulla
  Dhariwal, Arvind Neelakantan, Pranav Shyam, Girish Sastry, Amanda Askell,
  et~al.
\newblock Language models are few-shot learners.
\newblock {\em Advances in neural information processing systems},
  33:1877--1901, 2020.

\bibitem{BytePSrepo}
ByteDance.
\newblock {BytePS Environment Variables}.
\newblock \url{https://github.com/bytedance/byteps/blob/master/docs/env.md},
  2021.

\bibitem{LaMDA}
Aaron~Daniel Cohen, Adam Roberts, Alejandra Molina, Alena Butryna, Alicia Jin,
  Apoorv Kulshreshtha, Ben Hutchinson, Ben Zevenbergen, Blaise~Hilary
  Aguera-Arcas, Chung ching Chang, Claire Cui, Cosmo Du, Daniel De~Freitas
  Adiwardana, Dehao Chen, Dmitry~(Dima) Lepikhin, Ed~H. Chi, Erin Hoffman-John,
  Heng-Tze Cheng, Hongrae Lee, Igor Krivokon, James Qin, Jamie Hall, Joe
  Fenton, Johnny Soraker, Kathy Meier-Hellstern, Kristen Olson, Lora~Mois
  Aroyo, Maarten~Paul Bosma, Marc~Joseph Pickett, Marcelo~Amorim Menegali,
  Marian Croak, Mark Díaz, Matthew Lamm, Maxim Krikun, Meredith~Ringel Morris,
  Noam Shazeer, Quoc~V. Le, Rachel Bernstein, Ravi Rajakumar, Ray Kurzweil,
  Romal Thoppilan, Steven Zheng, Taylor Bos, Toju Duke, Tulsee Doshi,
  Vincent~Y. Zhao, Vinodkumar Prabhakaran, Will Rusch, YaGuang Li, Yanping
  Huang, Yanqi Zhou, Yuanzhong Xu, and Zhifeng Chen.
\newblock Lamda: Language models for dialog applications.
\newblock In {\em arXiv}. 2022.

\bibitem{bert}
Jacob Devlin, Ming-Wei Chang, Kenton Lee, and Kristina Toutanova.
\newblock Bert: Pre-training of deep bidirectional transformers for language
  understanding.
\newblock {\em arXiv preprint arXiv:1810.04805}, 2018.

\bibitem{dorfman2023docofl}
Ron Dorfman, Shay Vargaftik, Yaniv Ben-Itzhak, and Kfir~Yehuda Levy.
\newblock Docofl: Downlink compression for cross-device federated learning.
\newblock 2023.

\bibitem{differential_privacy}
Cynthia Dwork, Aaron Roth, et~al.
\newblock The algorithmic foundations of differential privacy.
\newblock {\em Foundations and Trends{\textregistered} in Theoretical Computer
  Science}, 9(3--4):211--407, 2014.

\bibitem{omnireduce}
Jiawei Fei, Chen-Yu Ho, Atal~N. Sahu, Marco Canini, and Amedeo Sapio.
\newblock Efficient sparse collective communication and its application to
  accelerate distributed deep learning.
\newblock In {\em Proceedings of the 2021 ACM SIGCOMM 2021 Conference}, ACM
  SIGCOMM '21, page 676–691, New York, NY, USA, 2021. Association for
  Computing Machinery.

\bibitem{brainwave}
Jeremy Fowers, Kalin Ovtcharov, Michael Papamichael, Todd Massengill, Ming Liu,
  Daniel Lo, Shlomi Alkalay, Michael Haselman, Logan Adams, Mahdi Ghandi,
  Stephen Heil, Prerak Patel, Adam Sapek, Gabriel Weisz, Lisa Woods, Sitaram
  Lanka, Steven~K. Reinhardt, Adrian~M. Caulfield, Eric~S. Chung, and Doug
  Burger.
\newblock A configurable cloud-scale dnn processor for real-time ai.
\newblock In {\em 2018 ACM/IEEE 45th Annual International Symposium on Computer
  Architecture (ISCA)}, pages 1--14, 2018.

\bibitem{gholami2018integrated}
Amir Gholami, Ariful Azad, Peter Jin, Kurt Keutzer, and Aydin Buluc.
\newblock Integrated model, batch, and domain parallelism in training neural
  networks.
\newblock In {\em Proceedings of the 30th on Symposium on Parallelism in
  Algorithms and Architectures}, pages 77--86, 2018.

\bibitem{gruntkowska2022ef21}
Kaja Gruntkowska, Alexander Tyurin, and Peter Richt{\'a}rik.
\newblock {EF21-P and Friends: Improved Theoretical Communication Complexity
  for Distributed Optimization with Bidirectional Compression}.
\newblock {\em arXiv preprint arXiv:2209.15218}, 2022.

\bibitem{Pingmesh}
Chuanxiong Guo, Lihua Yuan, Dong Xiang, Yingnong Dang, Ray Huang, Dave Maltz,
  Zhaoyi Liu, Vin Wang, Bin Pang, Hua Chen, Zhi-Wei Lin, and Varugis Kurien.
\newblock Pingmesh: A large-scale system for data center network latency
  measurement and analysis.
\newblock {\em SIGCOMM Comput. Commun. Rev.}, 45(4):139–152, aug 2015.

\bibitem{esa}
Wang Hao, Qin Yuxuan, Lao ChonLam, Le~Yanfang, Wu~Wenfei, and Chen Kai.
\newblock Preemptive switch memory usage to accelerate training jobs with
  shared in-network aggregation.
\newblock In {\em 2023 IEEE 30th International Conference on Network Protocols
  (ICNP)}, pages 1--11, 2022.

\bibitem{resnet}
K.~{He}, X.~{Zhang}, S.~{Ren}, and J.~{Sun}.
\newblock Deep residual learning for image recognition.
\newblock In {\em 2016 IEEE Conference on Computer Vision and Pattern
  Recognition (CVPR)}, pages 770--778, 2016.

\bibitem{ask}
Yongchao He, Wenfei Wu, Yanfang Le, Ming Liu, and ChonLam Lao.
\newblock A generic service to provide in-network aggregation for key-value
  streams.
\newblock In {\em Proceedings of the 28th ACM International Conference on
  Architectural Support for Programming Languages and Operating Systems, Volume
  2}, ASPLOS 2023, page 33–47, New York, NY, USA, 2023. Association for
  Computing Machinery.

\bibitem{hedayat1978hadamard}
A~Hedayat and Walter~Dennis Wallis.
\newblock Hadamard matrices and their applications.
\newblock {\em The Annals of Statistics}, pages 1184--1238, 1978.

\bibitem{tofino}
Intel.
\newblock {Barefoot Tofino}.
\newblock \url{https://www.barefootnetworks.com/technology/#tofino}.

\bibitem{microsoftsparecpu}
Myeongjae Jeon, Shivaram Venkataraman, Amar Phanishayee, Junjie Qian, Wencong
  Xiao, and Fan Yang.
\newblock Analysis of {Large-Scale} {Multi-Tenant} {GPU} clusters for {DNN}
  training workloads.
\newblock In {\em 2019 USENIX Annual Technical Conference (USENIX ATC 19)},
  pages 947--960, Renton, WA, July 2019. USENIX Association.

\bibitem{BytePS}
Yimin Jiang, Yibo Zhu, Chang Lan, Bairen Yi, Yong Cui, and Chuanxiong Guo.
\newblock A unified architecture for accelerating distributed {DNN} training in
  heterogeneous {GPU/CPU} clusters.
\newblock In {\em 14th USENIX Symposium on Operating Systems Design and
  Implementation (OSDI 20)}, pages 463--479. USENIX Association, November 2020.

\bibitem{ErrorFeedback}
Sai~Praneeth Karimireddy, Quentin Rebjock, Sebastian Stich, and Martin Jaggi.
\newblock Error feedback fixes signsgd and other gradient compression schemes.
\newblock In {\em International Conference on Machine Learning}, pages
  3252--3261. PMLR, 2019.

\bibitem{sipml}
Mehrdad Khani, Manya Ghobadi, Mohammad Alizadeh, Ziyi Zhu, Madeleine Glick,
  Keren Bergman, Amin Vahdat, Benjamin Klenk, and Eiman Ebrahimi.
\newblock Sip-ml: High-bandwidth optical network interconnects for machine
  learning training.
\newblock In {\em Proceedings of the 2021 ACM SIGCOMM 2021 Conference}, ACM
  SIGCOMM '21, page 657–675, New York, NY, USA, 2021. Association for
  Computing Machinery.

\bibitem{konevcny2018randomized}
Jakub Kone{\v{c}}n{\`y} and Peter Richt{\'a}rik.
\newblock Randomized distributed mean estimation: Accuracy vs. communication.
\newblock {\em Frontiers in Applied Mathematics and Statistics}, 4:62, 2018.

\bibitem{CIFAR100}
Alex Krizhevsky, Geoffrey Hinton, et~al.
\newblock Learning multiple layers of features from tiny images.
\newblock 2009.

\bibitem{ATP}
ChonLam Lao, Yanfang Le, Kshiteej Mahajan, Yixi Chen, Wenfei Wu, Aditya Akella,
  and Michael Swift.
\newblock {ATP}: In-network aggregation for multi-tenant learning.
\newblock In {\em 18th USENIX Symposium on Networked Systems Design and
  Implementation (NSDI 21)}, pages 741--761. USENIX Association, April 2021.

\bibitem{bart}
Mike Lewis, Yinhan Liu, Naman Goyal, Marjan Ghazvininejad, Abdelrahman Mohamed,
  Omer Levy, Ves Stoyanov, and Luke Zettlemoyer.
\newblock Bart: Denoising sequence-to-sequence pre-training for natural
  language generation, translation, and comprehension.
\newblock {\em arXiv preprint arXiv:1910.13461}, 2019.

\bibitem{li2020pytorch}
Shen Li, Yanli Zhao, Rohan Varma, Omkar Salpekar, Pieter Noordhuis, Teng Li,
  Adam Paszke, Jeff Smith, Brian Vaughan, Pritam Damania, et~al.
\newblock Pytorch distributed: Experiences on accelerating data parallel
  training.
\newblock {\em arXiv preprint arXiv:2006.15704}, 2020.

\bibitem{li2021distributed}
Xiaoyun Li, Belhal Karimi, and Ping Li.
\newblock On distributed adaptive optimization with gradient compression.
\newblock In {\em International Conference on Learning Representations}, 2021.

\bibitem{LossRadar}
Yuliang Li, Rui Miao, Changhoon Kim, and Minlan Yu.
\newblock Lossradar: Fast detection of lost packets in data center networks.
\newblock In {\em Proceedings of the 12th International on Conference on
  Emerging Networking EXperiments and Technologies}, CoNEXT '16, page
  481–495, New York, NY, USA, 2016. Association for Computing Machinery.

\bibitem{DGC}
Yujun Lin, Song Han, Huizi Mao, Yu~Wang, and Bill Dally.
\newblock Deep gradient compression: Reducing the communication bandwidth for
  distributed training.
\newblock In {\em International Conference on Learning Representations}, 2018.

\bibitem{janus}
Juncai Liu, Jessie~Hui Wang, and Yimin Jiang.
\newblock Janus: A unified distributed training framework for sparse
  mixture-of-experts models.
\newblock In {\em Proceedings of the ACM SIGCOMM 2023 Conference}, ACM SIGCOMM
  '23, page 486–498, New York, NY, USA, 2023. Association for Computing
  Machinery.

\bibitem{roberta}
Yinhan Liu, Myle Ott, Naman Goyal, Jingfei Du, Mandar Joshi, Danqi Chen, Omer
  Levy, Mike Lewis, Luke Zettlemoyer, and Veselin Stoyanov.
\newblock Roberta: {A} robustly optimized {BERT} pretraining approach.
\newblock {\em CoRR}, abs/1907.11692, 2019.

\bibitem{kashin2}
Yurii Lyubarskii and Roman Vershynin.
\newblock {Uncertainty Principles and Vector Quantization}.
\newblock {\em IEEE Transactions on Information Theory}, 56(7):3491--3501,
  2010.

\bibitem{htfpga}
Bradley McDanel, Sai~Qian Zhang, H.~T. Kung, and Xin Dong.
\newblock Full-stack optimization for accelerating cnns using powers-of-two
  weights with fpga validation.
\newblock In {\em Proceedings of the ACM International Conference on
  Supercomputing}, ICS '19, page 449–460, New York, NY, USA, 2019.
  Association for Computing Machinery.

\bibitem{megatron-lm-in-scale}
Deepak Narayanan, Mohammad Shoeybi, Jared Casper, Patrick LeGresley, Mostofa
  Patwary, Vijay Korthikanti, Dmitri Vainbrand, Prethvi Kashinkunti, Julie
  Bernauer, Bryan Catanzaro, Amar Phanishayee, and Matei Zaharia.
\newblock Efficient large-scale language model training on gpu clusters using
  megatron-lm.
\newblock In {\em Proceedings of the International Conference for High
  Performance Computing, Networking, Storage and Analysis}, SC '21, New York,
  NY, USA, 2021. Association for Computing Machinery.

\bibitem{dlrm}
Maxim Naumov, Dheevatsa Mudigere, Hao-Jun~Michael Shi, Jianyu Huang, Narayanan
  Sundaraman, Jongsoo Park, Xiaodong Wang, Udit Gupta, Carole-Jean Wu,
  Alisson~G Azzolini, et~al.
\newblock Deep learning recommendation model for personalization and
  recommendation systems.
\newblock {\em arXiv preprint arXiv:1906.00091}, 2019.

\bibitem{NVIDIASHARP}
NVIDIA.
\newblock {NVIDIA Scalable Hierarchical Aggregation and Reduction Protocol
  (SHARP). }.
\newblock \url{https://docs.nvidia.com/networking/display/SHARPv200}, 2020.

\bibitem{bytescheduler}
Yanghua Peng, Yibo Zhu, Yangrui Chen, Yixin Bao, Bairen Yi, Chang Lan, Chuan
  Wu, and Chuanxiong Guo.
\newblock A generic communication scheduler for distributed dnn training
  acceleration.
\newblock In {\em Proceedings of the 27th ACM Symposium on Operating Systems
  Principles}, SOSP '19, page 16–29, New York, NY, USA, 2019. Association for
  Computing Machinery.

\bibitem{finn-resnet}
Lucian Petrica, Tobias Alonso, Mairin Kroes, Nicholas Fraser, Sorin Cotofana,
  and Michaela Blott.
\newblock Memory-efficient dataflow inference for deep cnns on fpga.
\newblock In {\em 2020 International Conference on Field-Programmable
  Technology (ICFPT)}, pages 48--55, 2020.

\bibitem{philippenko2020bidirectional}
Constantin Philippenko and Aymeric Dieuleveut.
\newblock Bidirectional compression in heterogeneous settings for distributed
  or federated learning with partial participation: tight convergence
  guarantees.
\newblock {\em arXiv preprint arXiv:2006.14591}, 2020.

\bibitem{whenshouldhotos19}
Dan R.~K. Ports and Jacob Nelson.
\newblock When should the network be the computer?
\newblock In {\em Proceedings of the Workshop on Hot Topics in Operating
  Systems}, HotOS '19, page 209–215, New York, NY, USA, 2019. Association for
  Computing Machinery.

\bibitem{GPT2}
Alec Radford, Jeffrey Wu, Rewon Child, David Luan, Dario Amodei, Ilya
  Sutskever, et~al.
\newblock Language models are unsupervised multitask learners.
\newblock {\em OpenAI blog}, 1(8):9, 2019.

\bibitem{cassini}
Sudarsanan Rajasekaran, Manya Ghobadi, and Aditya Akella.
\newblock Cassini: Network-aware job scheduling in machine learning clusters.
\newblock {\em arXiv preprint arXiv:2308.00852}, 2023.

\bibitem{pmlr-v162-rajbhandari22a}
Samyam Rajbhandari, Conglong Li, Zhewei Yao, Minjia Zhang, Reza~Yazdani
  Aminabadi, Ammar~Ahmad Awan, Jeff Rasley, and Yuxiong He.
\newblock {D}eep{S}peed-{M}o{E}: Advancing mixture-of-experts inference and
  training to power next-generation {AI} scale.
\newblock In Kamalika Chaudhuri, Stefanie Jegelka, Le~Song, Csaba Szepesvari,
  Gang Niu, and Sivan Sabato, editors, {\em Proceedings of the 39th
  International Conference on Machine Learning}, volume 162 of {\em Proceedings
  of Machine Learning Research}, pages 18332--18346. PMLR, 17--23 Jul 2022.

\bibitem{nuqsgd}
Ali Ramezani-Kebrya, Fartash Faghri, Ilya Markov, Vitalii Aksenov, Dan
  Alistarh, and Daniel~M Roy.
\newblock Nuqsgd: Provably communication-efficient data-parallel sgd via
  nonuniform quantization.
\newblock {\em J. Mach. Learn. Res.}, 22:114--1, 2021.

\bibitem{yolo_2016_CVPR}
Joseph Redmon, Santosh Divvala, Ross Girshick, and Ali Farhadi.
\newblock You only look once: Unified, real-time object detection.
\newblock In {\em Proceedings of the IEEE Conference on Computer Vision and
  Pattern Recognition (CVPR)}, June 2016.

\bibitem{imagenet}
Olga Russakovsky, Jia Deng, Hao Su, Jonathan Krause, Sanjeev Satheesh, Sean Ma,
  Zhiheng Huang, Andrej Karpathy, Aditya Khosla, Michael Bernstein,
  Alexander~C. Berg, and Li~Fei-Fei.
\newblock {ImageNet Large Scale Visual Recognition Challenge}.
\newblock {\em International Journal of Computer Vision (IJCV)},
  115(3):211--252, 2015.

\bibitem{kashin1}
Mher Safaryan, Egor Shulgin, and Peter Richt{\'a}rik.
\newblock {Uncertainty Principle for Communication Compression in Distributed
  and Federated Learning and the Search for an Optimal Compressor}.
\newblock {\em arXiv preprint arXiv:2002.08958}, 2020.

\bibitem{SwitchML}
Amedeo Sapio, Marco Canini, Chen-Yu Ho, Jacob Nelson, Panos Kalnis, Changhoon
  Kim, Arvind Krishnamurthy, Masoud Moshref, Dan Ports, and Peter Richtarik.
\newblock Scaling distributed machine learning with {In-Network} aggregation.
\newblock In {\em 18th USENIX Symposium on Networked Systems Design and
  Implementation (NSDI 21)}, pages 785--808. USENIX Association, April 2021.

\bibitem{horovod}
Alexander Sergeev and Mike Del~Balso.
\newblock Horovod: fast and easy distributed deep learning in tensorflow.
\newblock {\em arXiv preprint arXiv:1802.05799}, 2018.

\bibitem{ParamScaling}
Jaime Sevilla, Lennart Heim, Anson Ho, Tamay Besiroglu, Marius Hobbhahn, and
  Pablo Villalobos.
\newblock Compute trends across three eras of machine learning.
\newblock {\em arXiv preprint arXiv:2202.05924}, 2022.

\bibitem{moe}
Noam Shazeer, Azalia Mirhoseini, Krzysztof Maziarz, Andy Davis, Quoc Le,
  Geoffrey Hinton, and Jeff Dean.
\newblock Outrageously large neural networks: The sparsely-gated
  mixture-of-experts layer.
\newblock {\em arXiv preprint arXiv:1701.06538}, 2017.

\bibitem{megatron-lm}
Mohammad Shoeybi, Mostofa Patwary, Raul Puri, Patrick LeGresley, Jared Casper,
  and Bryan Catanzaro.
\newblock Megatron-lm: Training multi-billion parameter language models using
  model parallelism.
\newblock {\em arXiv preprint arXiv:1909.08053}, 2019.

\bibitem{vgg}
Karen Simonyan and Andrew Zisserman.
\newblock Very deep convolutional networks for large-scale image recognition,
  2014.

\bibitem{sst2}
Richard Socher, Alex Perelygin, Jean Wu, Jason Chuang, Christopher~D Manning,
  Andrew~Y Ng, and Christopher Potts.
\newblock Recursive deep models for semantic compositionality over a sentiment
  treebank.
\newblock In {\em Proceedings of the 2013 conference on empirical methods in
  natural language processing}, pages 1631--1642, 2013.

\bibitem{TopK}
Sebastian~U. Stich, Jean-Baptiste Cordonnier, and Martin Jaggi.
\newblock Sparsified sgd with memory.
\newblock In {\em Proceedings of the 32nd International Conference on Neural
  Information Processing Systems}, NIPS'18, page 4452–4463, Red Hook, NY,
  USA, 2018. Curran Associates Inc.

\bibitem{network-bottleneck1}
Peng Sun, Wansen Feng, Ruobing Han, Shengen Yan, and Yonggang Wen.
\newblock Optimizing network performance for distributed dnn training on gpu
  clusters: Imagenet/alexnet training in 1.5 minutes, 2019.

\bibitem{hadamardSQ}
Ananda~Theertha Suresh, X~Yu Felix, Sanjiv Kumar, and H~Brendan McMahan.
\newblock {Distributed Mean Estimation With Limited Communication}.
\newblock In {\em International Conference on Machine Learning}, pages
  3329--3337. PMLR, 2017.

\bibitem{eden}
Shay Vargaftik, Ran~Ben Basat, Amit Portnoy, Gal Mendelson, Yaniv~Ben Itzhak,
  and Michael Mitzenmacher.
\newblock Eden: Communication-efficient and robust distributed mean estimation
  for federated learning.
\newblock In {\em International Conference on Machine Learning}, pages
  21984--22014. PMLR, 2022.

\bibitem{vargaftik2021drive}
Shay Vargaftik, Ran Ben-Basat, Amit Portnoy, Gal Mendelson, Yaniv Ben-Itzhak,
  and Michael Mitzenmacher.
\newblock Drive: One-bit distributed mean estimation.
\newblock {\em Advances in Neural Information Processing Systems}, 34:362--377,
  2021.

\bibitem{wang2020domain}
Hao Wang, Jingrong Chen, Xinchen Wan, Han Tian, Jiacheng Xia, Gaoxiong Zeng,
  Weiyan Wang, Kai Chen, Wei Bai, and Junchen Jiang.
\newblock Domain-specific communication optimization for distributed dnn
  training.
\newblock {\em arXiv preprint arXiv:2008.08445}, 2020.

\bibitem{wang2016database}
Wei Wang, Meihui Zhang, Gang Chen, H.~V. Jagadish, Beng~Chin Ooi, and Kian-Lee
  Tan.
\newblock Database meets deep learning: Challenges and opportunities.
\newblock {\em SIGMOD Rec.}, 45(2):17–22, September 2016.

\bibitem{TopoOpt}
Weiyang Wang, Moein Khazraee, Zhizhen Zhong, Manya Ghobadi, Zhihao Jia,
  Dheevatsa Mudigere, Ying Zhang, and Anthony Kewitsch.
\newblock {TopoOpt}: Co-optimizing network topology and parallelization
  strategy for distributed training jobs.
\newblock In {\em 20th USENIX Symposium on Networked Systems Design and
  Implementation (NSDI 23)}, pages 739--767, Boston, MA, April 2023. USENIX
  Association.

\bibitem{wang2018machine}
Zheng Wang and Michael O’Boyle.
\newblock Machine learning in compiler optimization.
\newblock {\em Proceedings of the IEEE}, 106(11):1879--1901, 2018.

\bibitem{ByteComp}
Zhuang Wang, Haibin Lin, Yibo Zhu, and T.~S.~Eugene Ng.
\newblock Hi-speed dnn training with espresso: Unleashing the full potential of
  gradient compression with near-optimal usage strategies.
\newblock In {\em Proceedings of the Eighteenth European Conference on Computer
  Systems}, EuroSys '23, page 867–882, New York, NY, USA, 2023. Association
  for Computing Machinery.

\bibitem{TernGrad}
Wei Wen, Cong Xu, Feng Yan, Chunpeng Wu, Yandan Wang, Yiran Chen, and Hai Li.
\newblock Terngrad: Ternary gradients to reduce communication in distributed
  deep learning.
\newblock In {\em Proceedings of the 31st International Conference on Neural
  Information Processing Systems}, NIPS'17, page 1508–1518, Red Hook, NY,
  USA, 2017. Curran Associates Inc.

\bibitem{weng2022mlaas}
Qizhen Weng, Wencong Xiao, Yinghao Yu, Wei Wang, Cheng Wang, Jian He, Yong Li,
  Liping Zhang, Wei Lin, and Yu~Ding.
\newblock Mlaas in the wild: Workload analysis and scheduling in large-scale
  heterogeneous gpu clusters.
\newblock In {\em 19th USENIX Symposium on Networked Systems Design and
  Implementation (NSDI 22)}, pages 945--960. USENIX Association, 2022.

\bibitem{wu2016google}
Yonghui Wu, Mike Schuster, Zhifeng Chen, Quoc~V Le, Mohammad Norouzi, Wolfgang
  Macherey, Maxim Krikun, Yuan Cao, Qin Gao, Klaus Macherey, et~al.
\newblock Google's neural machine translation system: Bridging the gap between
  human and machine translation.
\newblock {\em arXiv preprint arXiv:1609.08144}, 2016.

\bibitem{vitis_ai}
Xilinx.
\newblock {Vitis AI}.
\newblock
  \url{https://www.xilinx.com/products/design-tools/vitis/vitis-ai.html}, 2023.

\bibitem{usingtrio}
Mingran Yang, Alex Baban, Valery Kugel, Jeff Libby, Scott Mackie, Swamy
  Sadashivaiah~Renu Kananda, Chang-Hong Wu, and Manya Ghobadi.
\newblock Using trio: Juniper networks' programmable chipset - for emerging
  in-network applications.
\newblock In {\em Proceedings of the ACM SIGCOMM 2022 Conference}, ACM SIGCOMM
  '22, page 633–648, New York, NY, USA, 2022. Association for Computing
  Machinery.

\bibitem{unlockswitch}
Yifan Yuan, Omar Alama, Jiawei Fei, Jacob Nelson, Dan R.~K. Ports, Amedeo
  Sapio, Marco Canini, and Nam~Sung Kim.
\newblock Unlocking the power of inline {Floating-Point} operations on
  programmable switches.
\newblock In {\em 19th USENIX Symposium on Networked Systems Design and
  Implementation (NSDI 22)}, pages 683--700, Renton, WA, April 2022. USENIX
  Association.

\bibitem{network-bottleneck2}
Xiang Zhou, Ryohei Urata, and Hong Liu.
\newblock Beyond 1 tb/s intra-data center interconnect technology: Im-dd or
  coherent?
\newblock {\em Journal of Lightwave Technology}, 38(2):475--484, 2020.

\end{thebibliography}

\appendix 

\section{Uniform THC Preliminaries}\label{app:compressionPreliminaries}

Given a bandwidth budget of $b$ bits per coordinate (e.g., per gradient entry), we define a compression scheme using a pair of compression and decompression operators.

\begin{definition}[compression operator]
A compression operator $C:\mathbb{R}^d \to \{0,1\}^{b\cdot d}$ takes a real-valued $d$-dimensional vector and outputs a $(b\cdot d)$-bits compressed representation. 
\end {definition}
\begin{definition}[decompression operator]
A decompression operator $D:\{0,1\}^{b\cdot d}\to \mathbb{R}^d$ takes a $(b\cdot d)$-bits compressed representation and outputs a real-valued $d$-dimensional estimate of the input vector.
\end {definition}

More generally, a compression scheme may require sending some additional information. We, therefore, allow $b\cdot d + O(1)$ bits in practice.

%or may use slightly more than exactly $bd$ bits (such as $bd+O(1)$), or may even use a variable number of bits depending on the input. Our definitions apply to our work but can be trivially generalized.

For a vector $x\in\mathbb R^d$, we denote its estimate by $\widehat{x}=D(C(x))$. The goal of a compression scheme is then, given a bandwidth budget $b$, to minimize some error metric, e.g., the expected squared error, $\mathbb{E} [\norm{x-\widehat{x}}^2]$. 
Before describing THC, for ease of presentation, we present a simplified (uniform) version of the THC framework and later generalize it.

\subsection{Uniform Homomorphic Compression}
In distributed deep learning, at each training round, the mean of the workers' gradients forms the update of the model's parameters for the next round. 
Without compression, we could add all the workers' gradients and divide the results by the number of workers. 

To reduce the bandwidth with a minimal impact on accuracy, the Distributed Mean Estimation (DME) problem has been extensively studied~\cite{hadamardSQ,kashin1,kashin2,vargaftik2021drive,eden,QSGD,konevcny2018randomized}.
Namely, in DME, workers compress their gradients before sending them for aggregation.
Most DME works only consider compression in this direction, while messages from the parameter server to workers remain uncompressed. To achieve bidirectional compression, several works further suggest that the server, after decompressing and aggregating the gradients, will re-compress the result before sending it back (e.g.,~\cite{gruntkowska2022ef21,philippenko2020bidirectional}), introducing additional delay and error. Avoiding these problems motivates the following definition; for convenience, we henceforth \mbox{denote $\angles i=\set{0,\ldots,i-1}$ for any $i\in\mathbb N$.}

%Indeed, these actions occur in most gradient compression schemes (e.g.,~\cite{hadamardSQ,kashin1,kashin2,eden,QSGD,nuqsgd}). For bi-directional gradient compression, there exists only a handful of works (e.g.,~\cite{gruntkowska2022ef21,philippenko2020bidirectional}), and in these works, they may compress the average \ran{it's not they, its the server, average of what, etc.}, introducing further complexity and delay as well as additional compression error. Avoiding these problems motivates the following definition.  
% {\bf MM:  It's not clear what you're saying here:  I think what you mean to say is However, if we compress the gradients to reduce bandwidth, it is not generally the case that we can sum the compressed forms;  rather, it appears we have to decompress each compressed gradient before summing them.  Avoiding this problem motivates the following definition.}

% While the compression and decompression operators are defined on single vectors, to reason about HC compression, we must consider sets of vectors (e.g., a gradient for each worker). We start by introducing the following key definition.

\begin{definition}[Uniform Homomorphic Compression]\label{def:homomorphic_compression}

We say that a compression scheme $(C,D)$ is \emph{uniform homomorphic} \mbox{if for any $n,d\in\mathbb N$ and  $x_0,x_1,\ldots,x_{n-1} \in \mathbb R^d$,  it satisfies} $$\frac{1}{n} \cdot\sum_{i\in\angles{n}} D(C(x_i)) = D\parentheses{\frac{1}{n} \cdot\sum_{i\in\angles{n}} C(x_i)}.$$ That is, a Uniform Homomorphic Compression (UHC) scheme allows averaging the compressed representations and applying a single decompression invocation.
\end {definition}

% Our observation is that while gradient compression is a highly studied topic, most schemes do not consider \emph{bidirectional} compression (e.g.,~\cite{hadamardSQ,kashin1,kashin2,eden,QSGD,nuqsgd}).
% {\bf MM:  Note to check that bidirectional has been sufficiently explained/defined earlier.  My take is that this paragraph seems out of context with the definitions;  we should reorder to explain this as an advantage of UHC up front.}
% While some theoretical results have been derived for such compression methods (e.g.,~\cite{gruntkowska2022ef21,philippenko2020bidirectional}), these suffer from two main drawbacks. First, they require compute-intensive decompression, aggregation, and re-compression at the parameter server. Second, the re-compression introduces an additional error that further slows down the convergence.
That is, by using UHC, the parameter server can sum up the compressed gradients and send $\sum_iC(x_i)$ back (in compressed form) without increasing the delay and error.

\subsection{Uniform Stochastic Quantization}\label{sec:usq}
Two desired properties of gradient quantization schemes in a distributed setting are \emph{unbiasedness} (i.e., $\mathbb{E} [\widehat{x}] = x$) and \emph{independence} (i.e., each worker makes the random choice of their quantization values independently). These features are especially useful in distributed deep learning as the errors of the different workers then cancel out on average rather than add up, leading to a better estimation of the mean. 
% That is, some workers will round the value up, some round down, and the mean would give a good estimate of the mean of input coordinates.
% {\bf MM:  The previous sentence assumes rounding, which we haven't discussed in this section and isn't a requirement.  Rephrase/revise?}
% Notice that this argument does not hold for compressing the update sent from the server to the workers, as a worker received only a single message. Thus, there is no error cancellation. Homomorphic compression elegantly addresses this issue as it does not introduce any additional error at the server.
% {\bf MM:  I feel like this paragraph, or at least the last few sentences, belong in the previous section when talking about bidirectional compression.}

One of the most fundamental compression techniques that offer both properties is \emph{uniform stochastic quantization} (USQ). 
Intuitively, given a vector, $x\in \mathbb R^d$, and denoting its minimum by $m$ and maximum by $M$, using USQ to quantize each coordinate $x[j]$ to a single bit means rounding it to $m$ with probability $(x[j]-m)/(M-m)$ and to $M$ with probability $(M-x[j])/(M-m)$.
That is, the sender encodes the coordinate $x[j]$ using one bit $C(x)[j]\in\set{0,1}$ and the receiver estimates it as $\widehat x = D(C(x)) = m+(M-m)\cdot C(x)$.
Note that $m$ and $M$ need to be sent to the receiver as well. However, considering that $d$ is large, this overhead is negligible.
% {\bf MM:  Seems important to point out here that $m$ and $M$ may need to be sent as well!}

This idea generalizes to any number of bits per coordinate $b$ by partitioning the \emph{range} $[m,M]$ into $2^b-1$ \emph{uniform} (i.e., equal-length) intervals where each entry is rounded to one of its nearest endpoints $c_0, c_1$ with probabilities $(x[j]-c_0)/(c_1-c_0)$ and $(c_1-x[j])/(c_1-c_0)$ to make the estimate unbiased. That is, when the message for coordinate $x[j]$, $C(x)[j]\in\angles{2^{b}}$, is sent to the receiver that estimates the coordinate as $m+C(x)[j]\cdot (M-m)/(2^b-1)$. 
%{\bf MM: Might be worth clarifying which 2 values $C(x)[j]$ takes on and with which probabilities...}

%which uses a lookup table $T:\set{0,\ldots,2^{b}-1}\to[m,M]$ to estimate the input vector as $\widehat x = D(C(x))=\set{T(C(x)[j])\mid j\in\set{1,\ldots,d}}$. 

%\minlan{Can we talk about 4.3 first before discussing the drawbacks of existing works? }
Despite its popularity and simplicity, in our context, USQ has two main drawbacks. First, USQ is not homomorphic; this is because each worker {\em has its own} minimum and maximum values, and accordingly, the $b$-bit messages describing the same coordinate by different workers are not amenable to aggregation without decompressing the messages. 
Second, USQ's error highly depends on the input vector's distribution, e.g., the difference between the minimum and maximum. For example, for $b=1$, if the input vector is $(1,-1,0,0,\ldots,0)$, all the zero-valued coordinates will be rounded with an error of $1$, and the vector's estimate will greatly differ from the input.

Several recent works propose pre-processing each gradient and post-processing the average's estimate. The idea behind this approach is to change the vector's distribution prior to quantization to avoid bad cases. 
%In many cases, a post-processing step applies the inverse transform to allow the receiver to estimate the input vector rather than the transformed one.
For example, \cite{hadamardSQ} proposes to preprocess by applying the randomized Hadamard transform to ensure that the range is small with high probability and postprocess using the inverse transform. As another example, Kashin's representation~\cite{kashin1,kashin2} allows projecting the vector into a higher-dimensional {space with similar magnitude coefficients. }

\section{Optimally Solving the Lookup Table Optimization Problem}\label{app:lookup}
As described in~\cref{sec:lookup}, the solution of the following optimization problem yields the optimal lookup table $T_{b,g,p}=T$.
Observe that the problem depends on the parameters $b,g,p$, where $b$ is the number of bits workers send for each quantized coordinate, $g$ is the granularity (the range of values that the table can take), and $p$ is the expected fraction of transformed and scaled coordinates that are not taken into account when determining the truncation range $[-t_p,t_p]$.

\medskip
\resizebox{.995\columnwidth}{!}{
\centering
%\noindent
\hspace*{-7.0mm}
$
\begin{array}{l}
\displaystyle{\minimize_{P,T}}  \displaystyle \int_{-t_p}^{t_p} \sum_{z\in \angles{2^{b}}} P(a,z) \cdot \parentheses{a-T(z)}^2 \cdot \phi(a)\cdot  da\\\medskip
\text{subject to}\\\medskip
{\small (\textit{\textcolor{gray}{Unbiasedness}})} \displaystyle\ \  \sum_{z \in \angles{2^{b}}} P(a,z) \cdot T(z) = a  \quad\hfill\forall\, a \in [-t_p, t_p]\\\medskip
{\small (\textit{\textcolor{gray}{Probability}})}\,\ \quad \displaystyle \sum_{z\in \angles{2^{b}}}P(a,z)=1 \hfill\forall\, a \in [-t_p, t_p]\\\medskip
\quad\quad\quad\quad\quad\quad\,\,\, P(a,z)\ge0 \hfill \forall\, a \in [-t_p, t_p],\, z\in \angles{2^{b}}\\\medskip
{\small (\textit{\textcolor{gray}{Granularity}})}\quad\,\,\, T(z)\in \set{\frac{2t_p}{g}\cdot i  - t_p\mid i \in \angles{g+1}}\hfill \quad\forall\, z\in \angles{2^{b}}
\end{array}
$
}
\medskip

Recall that, without loss of generality, we may assume that $0=T(0) < T(1)<\ldots< T(2^b-1)=g$, which significantly narrows down the search range from $\parentheses{g+1}^{2^b}$ (which is the number of options to choose a table value for each table index).
Namely, this observation means that the number of possible options for $T$ is $SaB(g-2^b-1,2^b-1)$ ($SaB$ stands for the \emph{stars-and-bars}), where $SaB(n,k)={n+k-1 \choose k-1}$ is the number of options for throwing $n$ identical balls into $k$ distinct bins.
This is because we can think of $2^b-1$ bins representing the values $\set{T(i+1)-T(i)\mid i\in\angles{2^b-1}}$; this way, `throwing a ball' into the $i$'th bin corresponds to increasing the difference by $1$, and we have $g-2^b-1$ balls as all bins must be non-empty to enforce the strict monotonicity.
The number of options is therefore $SaB(g-2^b-1,2^b-1)={g-3\choose 2^b-2}\ll (g+1)^{2^b}$. For example, if $b=4, g=51$, we reduce the number of options from $52^{15}\approx 5.5\cdot 10^{25}$ to ${48\choose 14}\approx4.8\cdot 10^{11}$. We note that these $b,g$ values are the largest ones that we have found to be of interest as they yield a solution whose accuracy is on par with an uncompressed baseline.

To further reduce the number of options, if $g$ is odd, we leverage the symmetry of the normal distribution (i.e., that $\phi(a)=\phi(-a)$ for any $a\in\mathbb R$). In particular, together with the fact that the number of table indices is even, this implies that a table index $T(z)$ exists in the optimal table if and only if there exists $z'$ such that $T(z')=g-T(z)$.
In particular, this can be manifested as the following additional constraint:

\newcommand{\brackets}[1]{\left[#1\right]}

\medskip
\resizebox{.995\columnwidth}{!}{
\centering
%\noindent
\hspace*{-7.0mm}
$
\begin{array}{l}
{\small (\textit{\textcolor{gray}{symmetry}})}\quad\,\,\, T(z)=T(z+2^{b-1})  - \frac{g+1}{2}\hfill \quad\forall\, z\in \angles{2^{b-1}}
\end{array}
$
}
\smallskip

Notice that this further reduces the number of options to $SaB\parentheses{\frac{g+1}{2}-2^{b-1}-1,2^{b-1}-1}$. Using the $b=4, g=51$ example, we reduced the number of options to just $100947$.

This allows us to efficiently solve the problem optimally by computing the target integral for every possible value of $T$.
For each value, we use the fact that stochastic quantization (picking one of the two closest quantization values with probabilities that make it unbiased) is optimal given the quantization values~\cite{DBLP:conf/icalp/BasatMV21}. In particular, this means that we can easily derive the optimal $P$ values and thus compute the integral.
For example, if $b=2, g=4$ and $T(0)=0, T(1)=1, T(2)=3, T(3)=4$, we can compute the integral as:

\bigskip
\resizebox{.995\columnwidth}{!}{
\hspace*{-7.0mm}
$
\begin{array}{l}
\displaystyle \int_{-t_p}^{-t_p/2}\parentheses{ \frac{a-(-t_p)}{t_p/2} \cdot \parentheses{a-(-t_p/2)}^2 + \frac{-t_p/2-a}{t_p/2} \cdot \parentheses{a-(-t_p)}^2} \cdot \phi(a)\cdot  da    \\\\
+\displaystyle \int_{-t_p/2}^{t_p/2}\parentheses{ \frac{a-(-t_p/2)}{t_p} \cdot \parentheses{a-t_p/2}^2 + \frac{t_p/2-a}{t_p} \cdot \parentheses{a-(-t_p/2)}^2} \cdot \phi(a)\cdot  da    \\\\
+\displaystyle \int_{t_p/2}^{t_p}\parentheses{ \frac{a-t_p/2}{t_p/2} \cdot \parentheses{a-t_p}^2 + \frac{t_p-a}{t_p/2} \cdot \parentheses{a-t_p/2}^2} \cdot \phi(a)\cdot  da    
\end{array}
$
}
\medskip

\textbf{Enumerating over the options:} The last ingredient in our solver is a simple method to iterate over all the stars-and-bars options. Namely, consider wanting to throw $n$ balls into $k$ bins and let $B[i]$ denote the number of balls in the $i$'th bin. Then, the enumeration process is given by~\cref{alg:sab}.
\begin{algorithm}[t]
\small
\caption{Stars-and-Bars $(n,k)$ Enumeration}
\label{code:alg1appendix}
%\begin{multicols}{2}
\begin{algorithmic}[1]%\vspace*{-8mm}
%  \State $\text{If EF-enabled:\quad} x+ e$ 
    \State Initialize $B[0]=n, \quad B[1]=B[2]=\ldots,B[k-1]=0$
    \State \texttt{Yield}$(B)$ \Comment{This is the $0$'th option}
    \For {$j=1,2,\ldots,SaB(n,k)-1$}
        \State $a = \min\set{i \mid B[i] > 0}$
        \State $B[a+1] = B[a+1] + 1$
        \State $S = B[a]-1$
        \State $B[a]=0$
        \State $B[0] = S$
        \State \texttt{Yield}$(B)$ \Comment{$B$ is the $j$'th option}
    \EndFor
\end{algorithmic}
\label{alg:sab}
\end{algorithm}

The resulting solver is quite efficient: we ran it once for each of over $4000$ different $(b,g,p)$ combinations and computed all the optimal tables within mere minutes.

\crupdate{
\section{Additional Parameter Server Details}
\subsection{Pseudocode of Parameter Server (PS)}}
\label{sec:switch_aggregation}

\makeatletter
\newenvironment{pseudops}[1][htb]{%
    \renewcommand{\ALG@name}{Pseudocode}% Update algorithm name
   \begin{algorithm}[#1]%
  }{\end{algorithm}}
\makeatother

\setlength{\textfloatsep}{9pt}
{\small
\begin{pseudops}
\small
\renewcommand\thealgorithm{1}
\caption{\sysname PS processing logic}
\label{code:ps_logic}

\begin{algorithmic}[1]
\Statex\hspace*{-5mm} \textit{\textbf{Input:}} Gradient packet $pkt$ from worker. Workers generate {$pkt$.round\_num} and insert the {$pkt$.num\_worker} along with gradient data for each packet before sending.\vspace{-1.5mm}
  \Statex\hspace*{-5mm}\hrulefill\vspace{-0.5mm}

  % \Statex\hspace{-1mm}\hrulefill
    \If{$pkt$.roundnum < expected\_roundnum[$pkt$.agtr\_idx]} \label{line:precodestraggler}
        \State Notify straggler \label{line:precodestraggler2}
    \Else
        \If{\resizebox{.7581347909095\columnwidth}{!}{$pkt$.roundnum $=$ expected\_roundnum[$pkt$.agtr\_idx]}}
            \State recv\_count[$pkt$.agtr\_idx] += 1 \label{line:precode_reuseslot}
        \Else
            \State recv\_count[$pkt$.agtr\_idx] = 1 \label{line:precode_newslot}
            \State expected\_roundnum[$pkt$.agtr\_idx] = $pkt$.roundnum
        \EndIf
        \State Table indices lookup
        \State Aggregate table values 
        \If{recv\_count[$pkt$.agtr\_idx] = $pkt$.num\_worker} \label{line:precode_check_workers}
            \State Multicast back aggregation result \label{line:precode_send_back_start}
        \Else
            \State Drop $pkt$ 
        \EndIf \label{line:precode_send_back_end}
    \EndIf
\end{algorithmic}
    % \vspace{-5mm}
\end{pseudops}
}
The PS progressing logic is demonstrated as shown in Pseudocode~\ref{code:ps_logic}. 
%
%When the switch receives this packet, 
When workers' compressed gradient packets arrive, the PS will first check whether the \codesm{pkt.roundnum} is less than the \codesm{expected\_roundnum} it stores. If so, then this packet is carrying obsolete data, and the PS will discard this packet and notify the sender that it is likely straggling (Line \ref{line:precodestraggler}-\ref{line:precodestraggler2}).
%(in our current prototype, the straggling worker will try to catch up by simply skipping the current iteration and starting the backpropagation for the next iteration right away. More complex straggler handling such as having all workers periodically synchronize their models can be implemented.) 
Otherwise, the PS regards it as a normal case and updates the corresponding \codesm{recv\_count} counter.
%either receiving the first packet for a unused slot 
(Line \ref{line:precode_reuseslot}-\ref{line:precode_newslot}) %(the round number is greater than the stored one) 
%or a following packet for a slot (Line \ref{line:precode_newslot}). %(the round number is the same as the stored one)
After the PS finishes the table lookup and aggregation process for each packet, it will check if the aggregation is complete by comparing its \codesm{recv\_count} counter and the \codesm{pkt.num\_worker} (Line~\ref{line:precode_check_workers}). If the aggregation is complete, it will multicast the aggregated gradient packet \mbox{back, or drop the packet (Line \ref{line:precode_send_back_start}-\ref{line:precode_send_back_end}).   }
 %and the switch will handle this packet accordingly. On the worker's side, each worker increments the round number by one when it receives the data aggregated by a specific aggregator slot.

\crupdate{
\vspace*{-2mm}
\subsection{Switch Resources Usage}
\vspace*{-1mm}
\label{appendix:switch_resources}
The programmable switch version of \sysname Parameter Server has 32 aggregation blocks. Each aggregation block has a copy of the lookup table and can aggregate 32 bits (i.e., four 8-bit table values) in one pass. Overall, the programmable switch PS consumes 39.9 Mb of SRAM and 35 ALUs. \sysname workers send packets of 1024 table indices, so each packet needs $\frac{1024}{(32\times4)}=8$ passes to have all 1024 elements aggregated. Therefore, we recirculate a packet twice through each of the four pipelines (eight passes in total) and consume up to two recirculation ports per pipeline. When we have $N$ workers, the first $N-1$ packets are dropped after aggregation while the $N$-th packet is recirculated through all four pipelines once again to collect the aggregation results.
}

\section{Evaluation Figures}
\label{appendix:c} 
\subsection{Computational-intensive Model Training with THC}
\begin{figure}

    \begin{minipage}[b]{0.235\textwidth}
      \centering 
      \includegraphics[width=\textwidth]{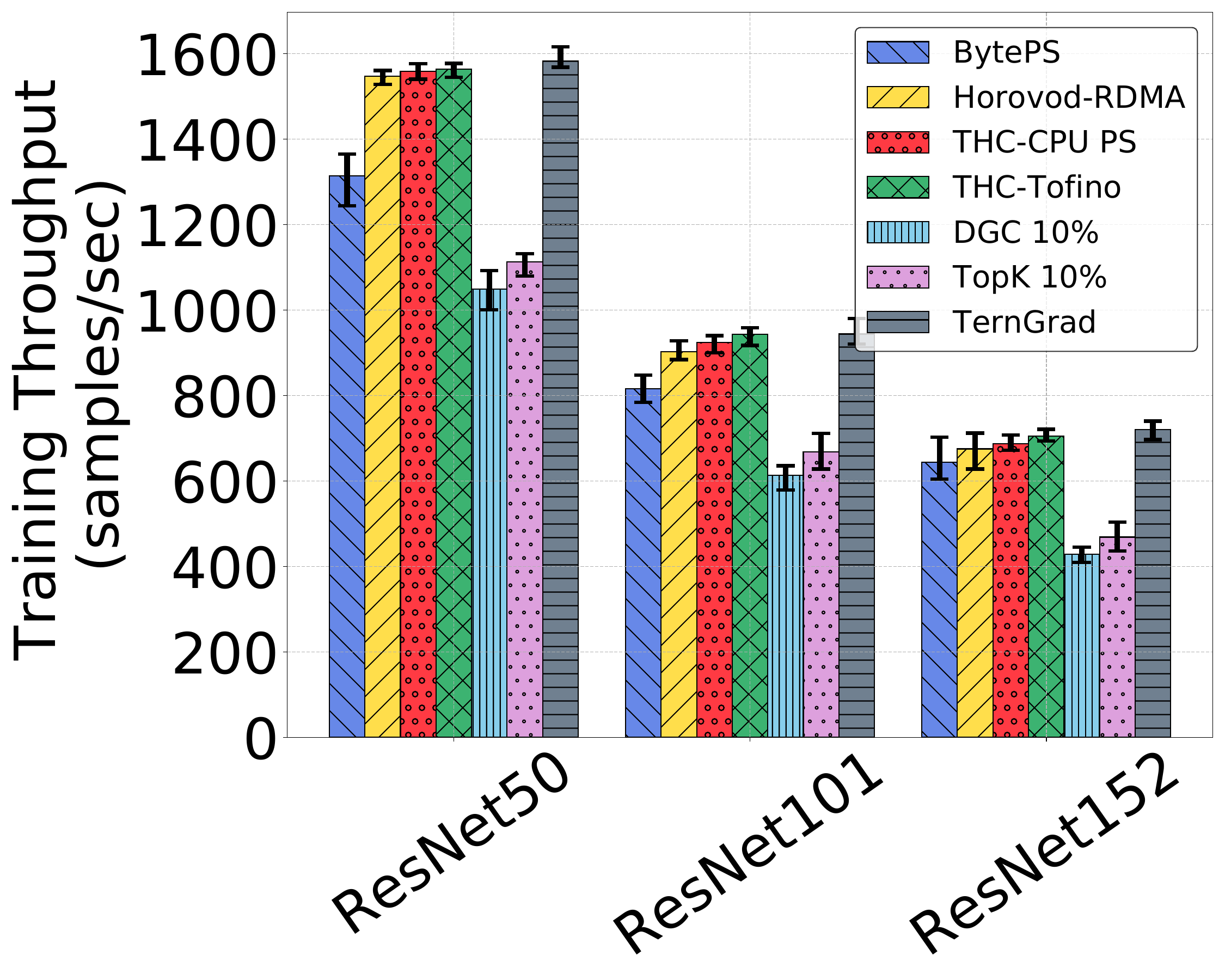} 
      %\vspace*{-2mm}
     \caption{\crupdate{Throughput of training ResNet models on the local testbed.}}
    \label{fig:throughput_resnet} 
    \end{minipage}
          \centering %here
    \begin{minipage}[b]{0.235\textwidth}
      \centering 
      \includegraphics[width=\textwidth]{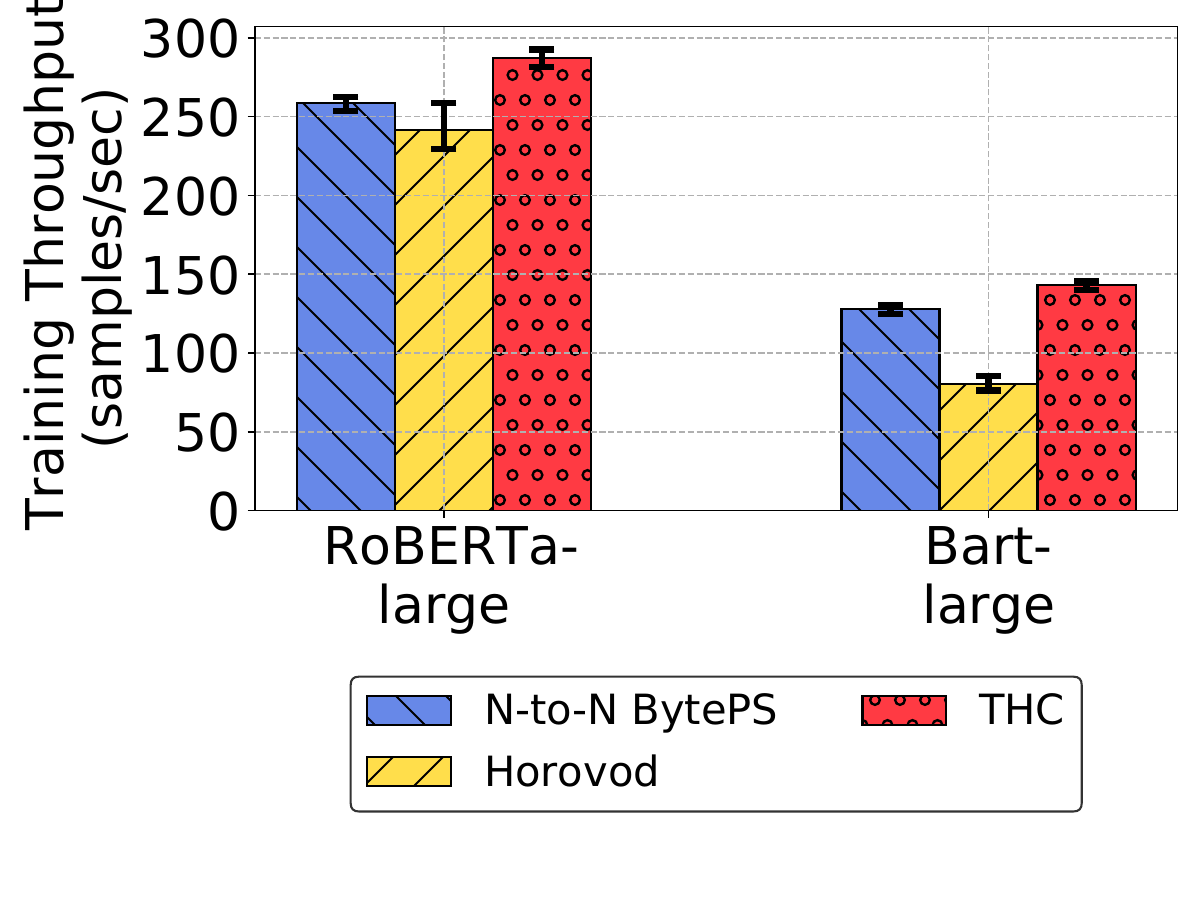} 
      %\vspace*{-2mm}
      \caption{\small Throughput of training RoBERTa-large and Bart-large on AWS EC2.}
      \label{fig:AWS_large_model} 
    \end{minipage}
        % \vspace{-5mm}
\end{figure}

Figure \ref{fig:throughput_resnet} demonstrates the throughput results of training ResNet\cite{resnet} models on our local testbed. Due to their computational-intensive nature, ResNet models don't experience much network bottleneck and hence don't benefit from accelerating inter-machine communication. Even with the most aggressive TernGrad compression, we are only able to improve the training throughput by up to $4.5\%$ of that of Horovod-RDMA. So, computational-intensive models are poor candidates for gradient compression and should be trained with full-precision gradients unless the network bandwidth is low.

\subsection{AWS EC2 Large Language Models Training Results}
Figure \ref{fig:AWS_large_model} show the throughput results of training RoBERTa-large and Bart-large on AWS EC2. We achieve a $1.11\times$ throughput improvement for RoBERTa-large and a $1.12\times$ throughput improvement for Bart-large.

% \Comment{Explain why using lognormal}
% \Comment{Says 10 workes in the 1st paragraph, but then we have a figure for differing number of workers?  Please check?}
% \kevin{fixed}

\subsection{Optimizations of THC}
\label{app:optimizations}
\crupdate{To see how THC performs with the different optimizations mentioned in Section \ref{THC}, we train both uniform and non-uniform THC with different optimizations. We use 4 workers on SST2 \cite{sst2} with RoBERTa. To measure the performance, we enable all optimizations on THC, and then we run uniform THC (UTHC) with and without rotation and error feedback. We keep all optimizations for THC because the algorithm assumes rotation and error feedback to be enabled a prior, so it would not make sense to disable them for the test.}

\begin{figure}[ht!]
    \centering
    \begin{minipage}[t]{0.49\linewidth}
      \centering 
      \includegraphics[width=\linewidth]{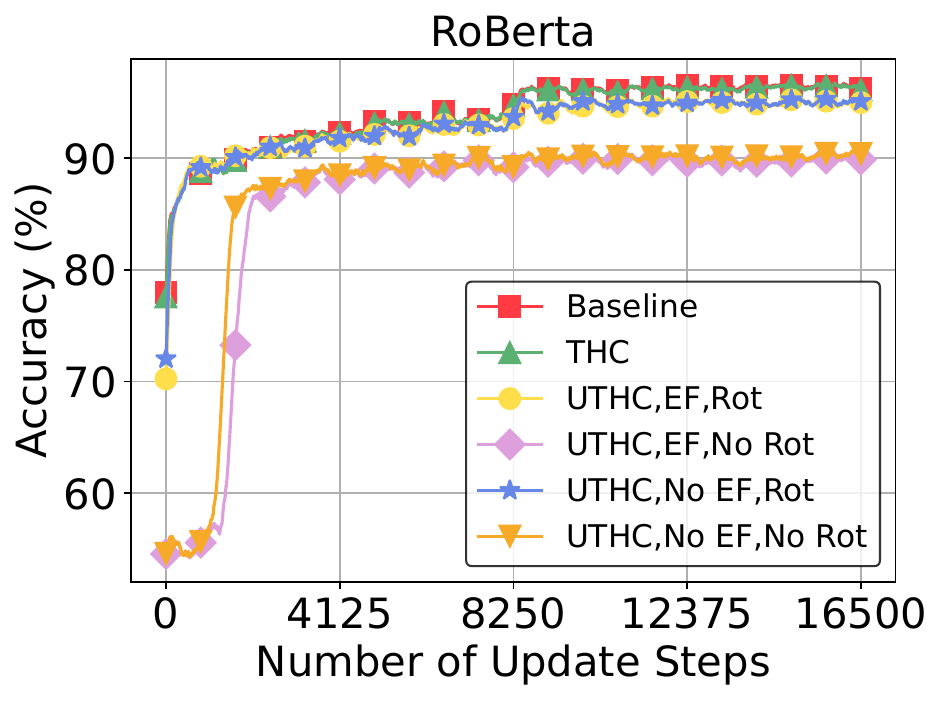} 
    \end{minipage}
    \begin{minipage}[t]{0.49\linewidth}
      \centering 
      \includegraphics[width=\linewidth]{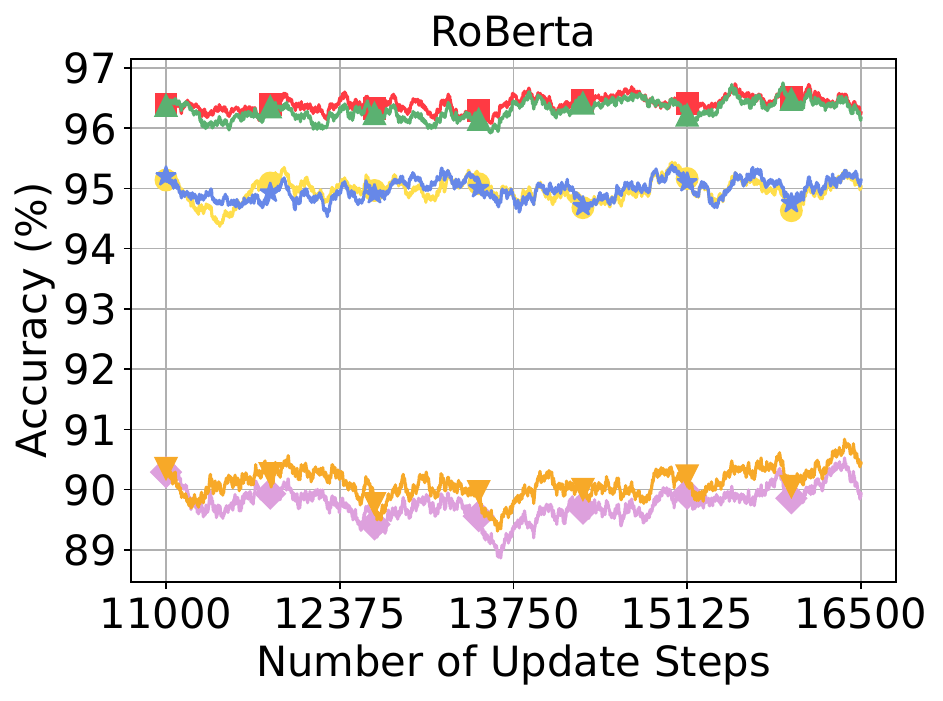}    
    \end{minipage}
    \caption{Accuracy of THC with Optimizations. The figure on the right is zoomed in on the last epoch, and accuracies are calculated on a sliding window of 336 batches.}
    \label{fig:improvements} 
\end{figure}

\crupdate{From Figure \ref{fig:improvements}, we see that THC performs the best overall as expected, nearly reaching baseline accuracy. UTHC with and without error feedback seems to perform similarly, although error feedback seems to increase the performance slightly. The largest difference is disabling rotation, which drops the final accuracy by about $5\%$. This is expected, since removing rotation introduces a large bias into the algorithm, resulting in a large error.}
\subsection{NMSE for Different Granularities}

\begin{figure}[!ht]
    \centering
    \begin{minipage}[t]{0.8\linewidth}
      \centering 
      \includegraphics[width=\textwidth]{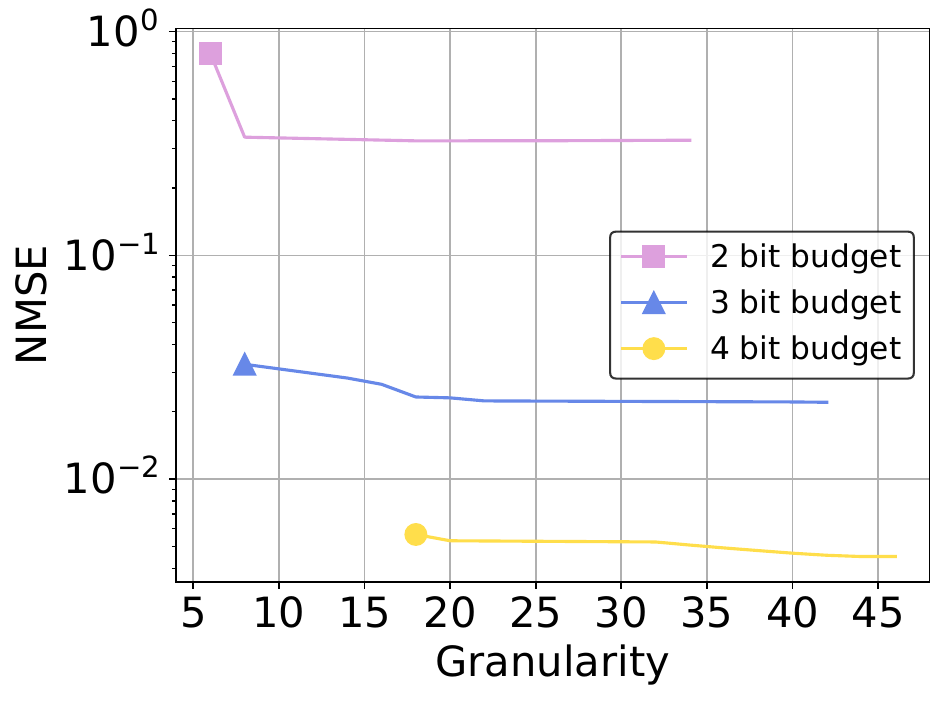} 
    \end{minipage}
    \caption{NMSE under different granularities and number of workers (plotted on log scale)}
    \label{fig:nmse} 
\end{figure}

We repeatedly compute the NMSE of the compression under different granularities. A gradient is first drawn from a lognormal distribution (which well approximate gradients in neural networks) and then copied multiple times to match the number of simulated workers.
We run THC on the copies of the gradient and compute the NMSE. Repeating 100 times, we report the average NMSE as we increase the granularity. 
Figure \ref{fig:nmse} varies the granularity of THC, while maintaining 10 workers and used a $p$-fraction of 1/1024. Three curves are plotted with bit budget 2/3/4. 

In Figure \ref{fig:nmse}, we first note that the largest discrepancy in NMSE is between different bit budgets where the error decreases by almost an order of magnitude between bit budgets of 2 to 3 to 4. The bit budget corresponds directly to the compression ratio of the algorithm, so it is expected that the accuracy should increase as we use more bits for compression. Furthermore, the NMSE of THC also decreases as the granularity increases since larger granularity values allows for more fine-grained choices of quantization values, though this effect is more difficult to see. 

\subsection{Test Accuracies for Resiliency Results}

\begin{figure}[ht!]
    \centering
    \begin{minipage}[t]{0.49\linewidth}
      \centering 
      \includegraphics[width=\linewidth]{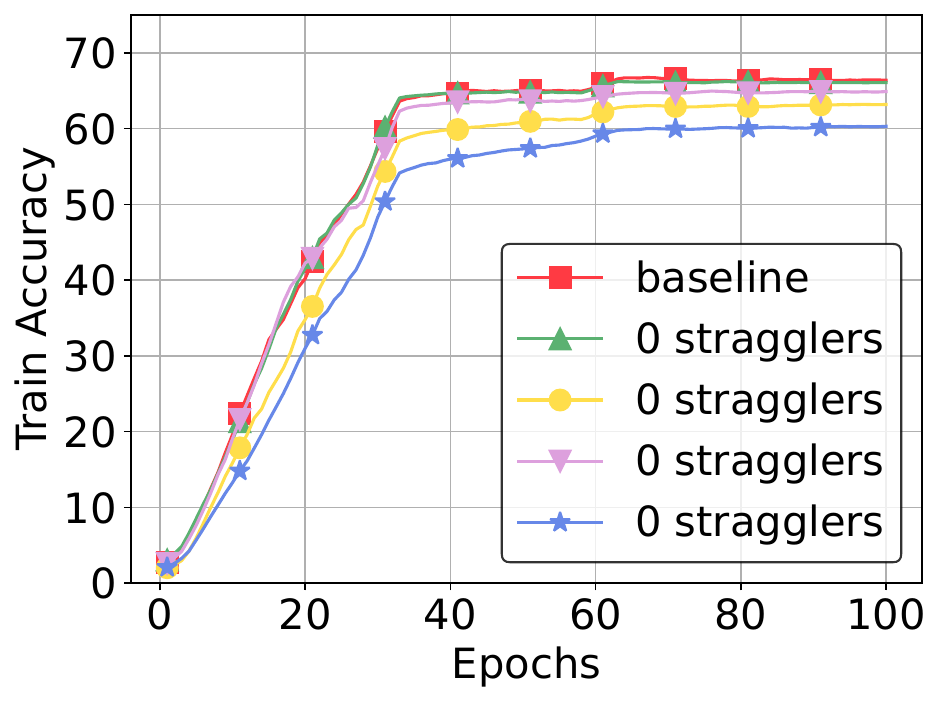} 
    \end{minipage}
    \begin{minipage}[t]{0.49\linewidth}
      \centering 
      \includegraphics[width=\linewidth]{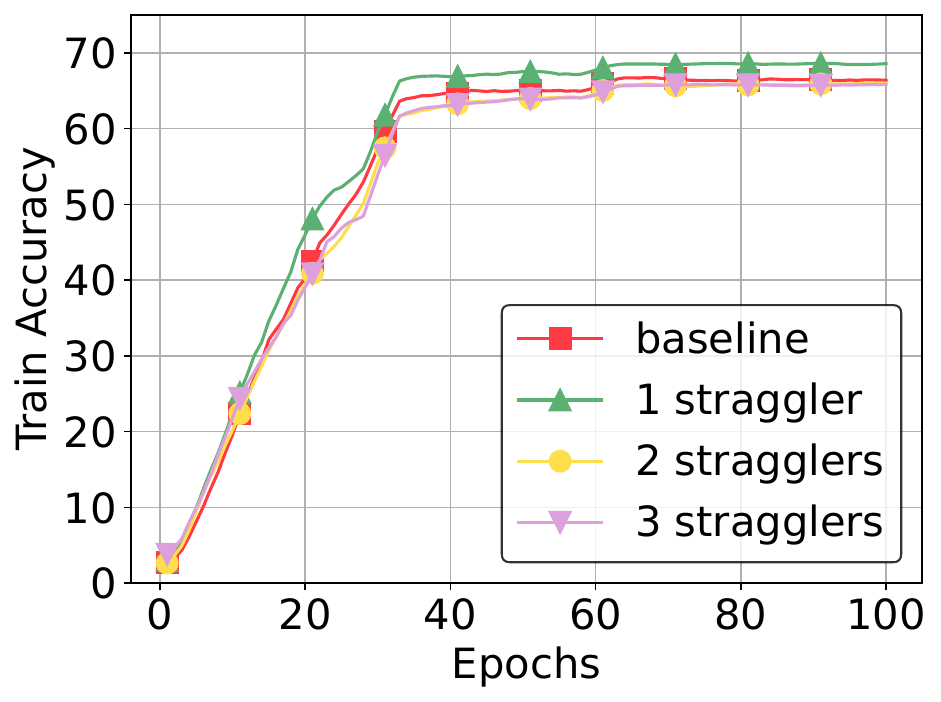}    
    \end{minipage}
    \caption{Test Accuracy with Gradient Losses}
    \label{fig:resiliency_test} 
\end{figure}

Similar to the training accuracy results, the test accuracy shows that (1) synchronization is beneficial when the system is lossy and (2) waiting for the top 90\% of workers does not affect the final accuracy. Figure~\ref{fig:resiliency_test} shows that under 1\%/0.1\% loss, the discrepancy from baseline drops from 6\%/3.2\% to 1.5\%/0.4\%. For 80\%/70\% stragglers, the error from baseline is roughly 0.5\%.

%%%%%%%%%%%%%%%%%%%%%%%%%%%%%%%%%%%%%%%%%%%%%%%%%%%%%%%%%%%%%%%%%%%%%%%%%%%%%%%%
\end{document}